\documentclass[preprint, 12pt, authoryear]{elsarticle}

\usepackage{amssymb}

\usepackage{mathtools}
\usepackage{here}
\usepackage{comment}
\usepackage{subcaption}
\usepackage{adjustbox}

\usepackage{booktabs}
\usepackage{multirow}
\usepackage{tabularx}
\newcolumntype{Y}{>{\centering\arraybackslash}X}

\usepackage[linesnumbered, ruled, vlined]{algorithm2e}

\usepackage[dvipsnames]{xcolor}
\usepackage{tikz}
\usetikzlibrary{positioning, arrows.meta, patterns, automata}
\tikzstyle{black_node}=[circle, draw, very thick, minimum height=8mm, minimum width=8mm]
\tikzstyle{black_rec_node}=[draw, very thick, minimum height=8mm, minimum width=8mm]

\usepackage{url}

\usepackage{thmtools}

\newtheorem{note}{Note}
\newtheorem{definition}{Definition}
\newtheorem{lemma}{Lemma}
\newtheorem{theorem}{Theorem}

\definecolor{revcolor}{rgb}{0,0,0}
\newcommand{\rev}[1]{\textcolor{revcolor}{#1}}
\definecolor{revbcolor}{rgb}{0,0,0}
\newcommand{\revb}[1]{\textcolor{revbcolor}{#1}}
\definecolor{revbremcolor}{rgb}{0,0,0}

\makeatletter
\newcommand\footnoteref[1]{\protected@xdef\@thefnmark{\ref{#1}}\@footnotemark}
\makeatother

\newcommand{\facts}{\mathcal{F}}
\newcommand{\actions}{\mathcal{A}}
\newcommand{\states}{\mathcal{S}}

\newcommand{\initialCondition}{\mathcal{I}}
\newcommand{\goalCondition}{\gamma}
\newcommand{\initialState}{s_\initialCondition}
\newcommand{\initialPolicy}{\pi_\initialCondition}

\newcommand{\pre}{\mathtt{pre}}
\newcommand{\effs}{\mathtt{effs}}
\newcommand{\eff}{\mathtt{eff}}
\newcommand{\add}{\mathtt{add}}
\newcommand{\del}{\mathtt{del}}

\newcommand{\succs}{\mathtt{succs}}
\renewcommand{\succ}{\mathtt{succ}}

\newcommand{\head}{\mathtt{begin}}
\newcommand{\tail}{\mathtt{end}}
\newcommand{\connects}{\mathtt{connects}}

\newcommand{\true}{\mathrm{True}}
\newcommand{\false}{\mathrm{False}}

\newcommand{\domain}{\mathtt{dom}}

\newcommand{\reach}{\mathtt{reach}}
\newcommand{\escape}{\mathtt{escape}}
\newcommand{\front}{\escape}
\newcommand{\open}{\front}

\newcommand{\remain}{\mathtt{remain}}

\newcommand{\decompress}{\mathtt{decompress}}
\newcommand{\pruned}{\mathtt{pruned}}
\newcommand{\buggy}{\mathtt{\revb{ambg}}}

\newcommand{\queue}{\mathtt{Queue}}
\newcommand{\done}{\mathtt{Done}}
\newcommand{\successors}{\mathtt{successors}}
\newcommand{\sign}{\mathtt{sign}}
\newcommand{\signStates}{\mathtt{sign2}}

\newcommand{\concretizer}{\mathtt{concretizer}}

\newcommand{\deltaNearest}{\Delta_{\downarrow}}
\newcommand{\hLMCut}{h^{\mathrm{LM}\text{-}\mathrm{Cut}}}
\newcommand{\hFF}{h^{\mathrm{FF}}}

\journal{Artificial Intelligence}

\begin{document}

\begin{frontmatter}



\title{Planning with Uncertainty: Symmetries, Policy Inference, and Solution Compression}

\author[ufrgs]{Frederico Messa}
\ead{frederico.messa@inf.ufrgs.br}

\author[ufrgs]{André Grahl Pereira}
\ead{agpereira@inf.ufrgs.br}

\tnotetext[license]{\textbf{This manuscript version is made available under the CC-BY-NC-ND 4.0 license.} Accepted for publication in \textit{Artificial Intelligence} (doi: \url{https://doi.org/10.1016/j.artint.2026.104574}).}

\affiliation[ufrgs]{organization={Institute of Informatics, Federal University of Rio Grande do Sul},
            country={Brazil}}

\begin{abstract}
    Fully-observable non-deterministic (FOND) planning is at the core of artificial intelligence planning with uncertainty. It models uncertainty through actions with non-deterministic effects.
    \color{revbcolor}
    In this work, we present a collection of techniques that establish explicit best-first policy-space search as a method competitive with the state of the art for solving FOND planning tasks. We study how to define equivalence relations between policies, allowing part of the search space to be pruned. We show it is possible to use group theory techniques to effectively compute canonical symmetries between states. We also present two contributions that go beyond just policy-space search: we present a procedure that infers in polynomial time a solution policy function given just the specification of its domain set, and an integer-programming formulation procedure that, given a solution policy defined over complete states, yields a set of resource-efficient models that are capable of finding a partial-state policy that represents it unambiguously with the fewest partial states possible.
    \color{black}
\end{abstract}



\begin{keyword}
    FOND Planning \sep
    Heuristic Search \sep
    Pruning \sep
    Symmetries \sep
    Compression
\end{keyword}

\end{frontmatter}


\pagebreak

\section{Introduction}
\label{sec:int}

\color{revcolor}

The ability to plan is one of the defining characteristics of intelligence. It allows the judicious selection of actions to achieve a goal, especially in unfamiliar or complex situations, where blindly performing actions may lead to undesired outcomes, such as unrecoverable failures or unnecessarily high costs. Planning is particularly important in environments with uncertainty, where the agent either does not have full knowledge of the world or does not have full control over the future, and must plan in advance for all possible contingencies that might arise.
In this work, we focus on fully-observable non-deterministic planning, a planning paradigm that models the uncertainty through actions with non-deterministic effects.

Fully-observable non-deterministic (FOND) planning is at the core of artificial intelligence planning with uncertainty. It assumes that the agent is capable of correctly enumerating all possible contingencies that might occur during the execution of his actions and has a plan for all of them. A successful plan for a FOND planning task is a policy describing, in advance, what actions the agent should take in each state he might reach so that he eventually reaches a goal state from the initial state. Although modeling uncertainty, concrete probabilities are not a concern in the context of FOND planning, and all possible outcomes of an action must be considered equally to ensure the plan provides flawless guarantees.

FOND planning has been used in a wide range of applications, including \emph{qualitative numerical planning}~\citep{bonet2020qualitative}, \emph{hyperproperty verification}~\citep{beutner2024non}, \emph{multi-agent planning}~\citep{muise2016planning}, \emph{contingent planning}~\citep{albore2009translation}, \emph{resilient planning}~\citep{aineto2023action}, \emph{generalized planning} and \emph{LTL synthesis}~\citep{bonet2017generalized, camacho2018finite}, and \emph{dialogue-agent systems design}~\citep{muise2019planning}.

There are several works in the literature that present techniques for solving FOND planning tasks, from approaches that use classical planning techniques to first generate optimistic policies and iteratively refine them into flawless solutions \citep{kuter2008using, fu2011simple, muise2012improved, pereira2022iterative, muise2024prp} and approaches that incrementally solve the task for bigger and bigger portions of the state space until solving the original problem \citep{hansen2001lao, mattmuller2010pattern}, to approaches that compile FOND into some more general paradigm \citep{cimatti2003weak, kissmann2009solving, ramirez2014directed, geffner2018compact, rodriguez2022fond, yadav2023declarative}. In this work, we focus on another direction for solving FOND planning tasks: policy-space explicit search~\citep{messa2023best}.

\revb{A$^*$ with Non-Determinism (AND$^*$ for short)~\citep{messa2023best} ``generalizes'' A$^*$~\citep{hart1968formal} for FOND planning. It performs a best-first search in a constructive space of policies to find a solution policy for the FOND planning task.}
Starting with only the empty policy, which maps no state to no action, AND$^*$ uses a priority queue to keep track of a range of incomplete policies that are candidates to become a solution. It repeatedly chooses from the queue the most promising candidate policy until it finds one that is a solution. If the chosen policy is not yet a solution, AND$^*$ improves it to generate new candidate policies, which are then inserted in the queue. The search continues until a solution is found or the queue becomes empty.
AND$^*$ is sound and complete, and has optimality guarantees when using an appropriate heuristic function to guide the search. \revb{In this work, we study the structure of FOND policies and policy-space search aiming to develop a collection of techniques to improve the performance of AND$^*$.}

\color{revbcolor}
We study policy and state equivalences in the context of policy-space search. With the knowledge of policy equivalences, we can make AND$^*$ prune part of the search space by expanding only one policy from each equivalence class. We show that state symmetries can be used to broaden the concepts of policy equivalences and that it is possible to improve the computation of state symmetries in the context of FOND planning, using group theory techniques.

We present two contributions whose applications go beyond AND$^*$ and policy-space search. The first is a procedure, which we call \emph{the concretizer}, that, given a restriction specifying which parts of the state space the agent is allowed to reach, constructs---in time polynomial in the size of this portion of the state space---a solution policy that precisely ``matches'' it, if and only if one exists. We equip AND$^*$ with the concretizer to provide it an extra mechanism for finding solutions. This modification makes it capable of performing more aggressive prunings in its standard policy-space search, without becoming incomplete, as it is able to recover pruned solutions in certain circumstances. The concretizer has the potential to foster new planning approaches, since it allows changing the objective of the search.

The second is a compression technique, which we call \emph{the compressor}, that, given a solution policy defined over complete states, finds---through the use of integer programming---a partial-state policy that unambiguously represents the same information using the minimum number of partial states possible (what can take exponentially less space). We test the compression technique, showing that the compressor is time and memory-efficient for the benchmark tasks we use in this work, and that applying the compressor in the solutions returned by AND$^*$, makes them have state-of-the-art compactness in the majority of the benchmark domains. The compressor is not limited to be used by AND$^*$, as it can be used to compress complete-state solutions provided by any planner.

Finally, we also evaluate applying some satisficing search techniques from the literature to further improve the effectiveness of AND$^*$, and compare it with other planners from the literature.

In sum, this work presents and studies a collection of techniques that, in unison, are able to establish explicit policy-space search, initially introduced by \cite{messa2023best}, as a competitive method for solving FOND planning tasks. FOND planning heuristics and other complementary topics introduced by \cite{messa2023best} are not part of this work.

\subsection{Outline}

This work is structured as follows:

\begin{itemize}
    \item We start by presenting a deeper \textbf{background} on FOND planning and policy-space search (Section~\ref{sec:bkg});
    \item Then, we describe the \textbf{experimental settings}: computational environment, benchmarks, resource limits, etc (Section~\ref{sec:ee});
    \item Then, we present the concept of search-space equivalence pruning and evaluate some possible concepts of \textbf{equivalences between policies}. Here we also present \textbf{the concretizer} (Section~\ref{sec:psym});
    \item We then proceed by presenting concepts of \textbf{state symmetries} and we use them to strengthen the policy equivalences (Section~\ref{sec:ssym});
    \item Then, we present the \textbf{solution compressor} and evaluate it. Here we also enable and evaluate some \textbf{satisficing features} (Section~\ref{sec:comp});
    \item Finally, we compare AND$^*$ with \textbf{state-of-the-art} FOND planners (Section~\ref{sec:sota}) and conclude (Section~\ref{sec:c-fw}).
\end{itemize}

\color{black}
A helpful glossary of the main symbols and functions used in this work is provided on the last page of this work (Appendix A).

\section{Background on FOND Planning and Policy-Space Search}
\label{sec:bkg}

A FOND planning task $\Pi$, in STRIPS~\citep{fikes1971strips} formalism, is a 4-tuple $\langle \facts, \initialCondition, \goalCondition, \actions \rangle$, with $\facts$ a finite set of \emph{facts} that can be either true or false, $\initialCondition \subseteq \facts$ the facts that are true in \emph{the initial state}~$\initialState$, $\goalCondition \subseteq \facts$ the facts that we want to be true, i.e.~\emph{the goal}, and $\actions$ a finite set of \emph{actions}.
$\facts[s]$ are the facts that are true in a state $s$, and a state $s_*$ is a \emph{goal state} iff $\facts[s_*] \supseteq \goalCondition$. There are $2^{|\facts|}$ states, one for each truth valuation. We call $\states$ the set of all possible states, and $\states_* \subseteq \states$ the set with all goal states.

Each action $a \in \actions$ is a pair $\langle \pre(a), \effs(a) \rangle$, with $\pre(a) \subseteq \facts$ the precondition of $a$ and $\effs(a) = \{\eff_{1}(a), \eff_{2}(a), \dots \eff_{n \geq 1}(a)\}$ the list of possible effects of applying~$a$. An action $a$ can be applied to a state~$s$ iff $\facts[s] \supseteq \pre(a)$. Each $\eff_{i}(a)$ of $a$ is a pair $\langle \del_{i}(a), \add_{i}(a) \rangle$, with $\del_{i}(a) \subseteq \facts$, $\add_{i}(a) \subseteq \facts$ and $\del_{i}(a) \cap \add_{i}(a) = \emptyset$. The application of $a$ to a given state $s$ non-deterministically results in a state $s' \in \succs(s, a) = \{\succ_{1}(s, a),\break \succ_{2}(s, a), \dots, \succ_{n}(s, a)\}$, with $\facts[\succ_{i}(s, a)] = (\facts[s] \setminus \del_{i}(a)) \cup \add_{i}(a)$.

An action is said to be \emph{fair}~\citep{cimatti2003weak}, if each possible outcome of an action will occur infinitely often if the action is applied infinitely often in a given state. Otherwise, it is said to be \emph{adversarial}. For simplicity, in this work we assume that all actions are fair, as done in some works in the literature~\citep{pereira2022iterative, messa2023best, muise2024prp}. Nevertheless, most of the contributions presented in this work can be adapted to work in adversarial environments as well. The worst-case complexity of FOND planning is EXPTIME-complete in both cases~\citep{littman1997probabilistic,mattmuller2010pattern}.

A \emph{transition} $t$ is a triplet $\langle s, a, s' \rangle$ with $s$ a state, $a$ an action applicable to $s$, and $s' \in \succs(s, a)$. The start of $t$ is $\rev{\head}(t) = s$ and the end of $t$ is $\rev{\tail}(t) = s'$. And the action of $t$ is $\actions[t] = a$.

A \emph{trajectory} $\omega$ is a sequence of transitions $\langle t_1, t_2, \dots, t_n \rangle$, with $\tail(t_i) = \head(t_{i+1})$ for each $i \in \{1, 2, \dots, n - 1\}$. \revb{For every pair of states $s$ and $s'$, if $n \geq 1$, $\connects(\omega, s, s')$ is defined to be $\top$ if $s = \head(t_1)$ and $s' = \tail(t_n)$, otherwise $\bot$. If $n = 0$, then the trajectory is empty and $\connects(\omega, s, s')$ is defined to be $\top$ if $s = s'$, otherwise~$\bot$.}

A \emph{policy} $\pi$ is a partial function mapping non-goal states to actions, with $\pi[s]$ applicable to $s$ for each $s \in \domain(\pi)$, where $\domain(\pi)$ denotes the set of states mapped by $\pi$, called the \emph{domain} of $\pi$. \revb{We denote $ \pi|_{X} = \{s \mapsto \pi[s] : s \in \domain(\pi) \cap X\}$ for $X \subseteq \states$.}

A \emph{$\pi$-trajectory} is a trajectory $\omega = \langle t_1, t_2, \dots, t_n \rangle$, with $\actions[t_i] = \pi[\head(t_i)]$ for each $i \in \{1, 2, \dots, n\}$. \revb{Following the guidance of a policy~$\pi$ under a non-adversarial environment, one has a chance of reaching a state~$s'$ starting from a state~$s$ if and only if there is a $\pi$-trajectory~$\omega$ with $\connects(\omega, s, s') = \top$.}

\revb{The \emph{reachability set} of $\pi$ is $\reach(\pi) = \{s' : \exists s\in\domain(\pi) \text{ and } \exists\pi\text{-trajec-}\break\text{tory}\ \omega\ \text{such that}\ \connects(\omega, s, s') = \top\} = \domain(\pi) \cup \bigcup_{s \in \domain(\pi)}\succs(s, \pi[s])$.} The reachability set lists all the states that one, following the guidance of the policy, might reach starting from \emph{any} of the states mapped by the policy. We call \revb{$\reach(\pi, s) = \{s' : \exists \pi\text{-trajectory}\ \omega\ \text{such that}\ \connects(\omega, s, s') = \top\}$}.

The \emph{\rev{escape}~set} of~$\pi$ is $\rev{\front}(\pi) = \reach(\pi) \setminus \domain(\pi)$. \revb{It contains the \emph{unmapped} states that one may reach starting from a state mapped by the policy.} We call $\escape(\pi, s) = \reach(\pi, s) \cap \front(\pi)$.

A policy~$\pi$ is \emph{closed} iff $\front(\pi) = \emptyset$ and \emph{goal-closed} iff $\front(\pi) \subseteq \states_*$. A policy~$\pi$ is \emph{proper} iff $\escape(\pi, s) \neq \emptyset$ for each $s \in \domain(\pi)$. We note that if~$\pi$ is proper, then each~$\pi' \subseteq \pi$ is proper as well. A policy that is simultaneously goal-closed and proper is called \emph{strong-cyclic}\footnote{\revb{We note that ``Strong-cyclic'' is a terminology that comes from the literature~\citep{cimatti2003weak}. It does not imply the existence of cyclic trajectories, but allows.}}.

A \emph{solution} for $\Pi$ is a strong-cyclic policy $\pi_*$ with $\initialState \in S_* \cup \domain(\pi_*)$. \rev{In other words, a solution is a policy that under a non-adversarial environment safely guides one from $\initialState$ \emph{eventually} to a goal state, by addressing all the possible outcomes of the taken actions, and ensuring that: (1) from every state mapped by the policy it is not possible to reach a state that is not handled by the policy unless it is a goal state (goal-closedness); (2) from every state mapped by the policy there is always a chance to reach a state not handled by the policy (properness + fairness); and (3) the initial state is either a goal state or is mapped by the policy.
One following a solution policy may possibly visit some states an arbitrary number of times before eventually reaching the goal, due to possible cyclic $\pi_*$-trajectories. Yet, while it does not reach the goal, it is ensured to always reach only states mapped by the policy, from where there will be a new chance to reach the goal.}

We define $\remain(\pi) = (\front(\pi) \setminus \states_*) \cup (\{\initialState\} \setminus (\states_* \cup \domain(\pi)))$. The $\remain$~set captures if a policy is goal-closed and deals with the initial state. Specifically, $\remain(\pi) = \emptyset$ iff $\pi$ is goal-closed and $\initialState \in S_* \cup \domain(\pi_*)$. Consequently, a \emph{proper} policy~$\pi$ is a solution if and only if $\remain(\pi) = \emptyset$.

We remind that a simplified glossary of the symbols and functions we most use throughout this work is present on the last page of this work (in the Appendix).
We now exemplify some of the presented concepts.

\begin{figure}[tb]
    \centering
    \begin{tikzpicture}
        \node[black_node]            at ( 0.0,  0.0) (a) {$s_A$};
        \node[black_node]            at (+1.5, -1.5) (b) {$s_B$};
        \node[black_node]            at ( 0.0, -3.0) (c) {$s_C$};
        \node[black_node]            at (-1.5, -1.5) (d) {$s_D$};
        \node[black_node]            at (-3.0, -3.0) (e) {$s_E$};
        \node[black_node, accepting] at (-4.0, -1.5) (f) {$s_F$};

        \draw[->, thick] ( 0.0, 0.9) -- (a) ;

        \draw[->, thick] (a) -- (b) ;
        \draw[->, thick] (a) -- (d) ;
        \draw[->, thick] (b.250) to [bend left=20] (c.20) ;
        \draw[->, thick] (c) to [bend left=20] (b) ;
        \draw[->, thick] (c) -- (d) ;
        \draw[->, thick] (d.202) -- (f) ;
        \draw[->, thick] (d) to [bend right=20] (e) ;
        \draw[->, thick] (e.20) to [bend right=20] (d.250) ;

        \draw[thick] (0.0, -1.0) .. controls (+0.2, -1.0) and (+0.383, -0.924) .. (+0.707, -0.707) ;
        \draw[thick] (0.0, -1.0) .. controls (-0.2, -1.0) and (-0.383, -0.924) .. (-0.707, -0.707) ;

        \draw[thick] (-2.52, -1.62) .. controls (-2.51, -1.81) and (-2.49, -1.85) .. (-2.4, -2.0) ;

        \node[] at (0, -1.15) () {$a$};
        \node[] at (1.12, -2.6) () {$b$};
        \node[] at (-0.52, -2.1) () {$c_L$};
        \node[] at (0.15, -2.1) () {$c_R$};
        \node[] at (-2.68, -1.88) () {$d$};
        \node[] at (-1.88, -2.6) () {$e$};
    \end{tikzpicture}
    \caption{\rev{Example state space to illustrate the background concepts.}}
    \label{fig:state-space-1}
\end{figure}

\paragraph{Example}

Figure~\ref{fig:state-space-1} depicts the state space of a FOND planning task.  It has six states $s_A$, $s_B$, $s_C$, $s_D$, $s_E$, and $s_F$. The initial state is $s_A$, and $s_F$ is the single goal state. There are six actions $a$, $b$, $c_L$, $c_R$, $d$, and $e$. The only non-deterministic actions are $a$ --- which can lead to either $s_B$ or $s_D$ from $s_A$ --- and $d$ --- which can lead to either $s_E$ or $s_F$ from $s_D$. We use it to exemplify some of the presented concepts regarding policies.

The policy $\pi_1 = \{s_D \mapsto d, s_E \mapsto e\}$ is a strong-cyclic policy, meaning that it is goal-closed and proper. It is goal-closed because $\front(\pi_1) = \{s_F\} \subseteq \states_*$. It is proper because for each $s \in \domain(\pi_1) = \{s_D, s_E\}$, $\escape(\pi_1, s) = \reach(\pi_1, s) \cap \front(\pi_1) \neq \emptyset$, given that there is a $\pi_1$-trajectory leading $s_D$ to $s_F$ and another leading $s_E$ to $s_F$. For instance, the $\pi_1$-trajectories $\langle \langle s_D, d, s_F \rangle \rangle$ and $\langle \langle s_E, e, s_D \rangle, \langle s_D, d, s_F \rangle \rangle$ respectively lead $s_D$ and $s_E$ to $s_F$. Although one may cycle an arbitrary number of times between $s_D$ and $s_E$ before reaching a solution if they follow $\pi_1$ actions starting from one of these states, the assumption of \revb{non-adversariality (fairness)} ensures that the ``incorrect'' outcome of applying $e$ will not occur forever and that the one leading them to $s_F$ will eventually occur. Note that $\pi_1$ is not a solution because it does~not cover the initial state (i.e. $s_A \not\in \domain(\pi_1)$). The only solution for this example~task is $\pi_2 = \{s_A \mapsto a, s_B \mapsto b, s_C \mapsto c_L, s_D \mapsto d, s_E \mapsto e\}$. Note that $\pi_2$ is a superset of~$\pi_1$.

The policy $\pi_3 = \{s_A \mapsto a, s_B \mapsto b, s_C \mapsto c_R, s_D \mapsto d, s_E \mapsto e\}$ is not a solution because it is not proper. It is not proper because $\escape(\pi_3, s_B) = \emptyset$ and $\escape(\pi_3, s_C) = \emptyset$. This occurs because $\succs(s_B, \pi_3[s_B]) = \{s_C\}$ and $\succs(s_C, \pi_3[s_C]) = \{s_B\}$. They form what we call a \emph{deadlock}. Unlike the cycle between $s_D$ and $s_E$, the cycle formed by $s_B$ and $s_C$ in $\pi_3$ is inescapable because there is no outcome that leads outside it.

\subsection{Searching for a Solution -- The AND$^*$ Algorithm}
\label{ssec:and-star}

\begin{algorithm}[th]
    \DontPrintSemicolon
    \SetKwIF{If}{ElseIf}{Else}{if}{:}{else if}{else:}{endif}
    \SetKwFor{While}{while}{:}{endw}
    \SetKwFor{ForEach}{for each}{:}{endfch}
    \SetKwProg{Pn}{Method}{:}{}

    $\initialPolicy \coloneqq \emptyset$,
    $\mathrm{Queue} \coloneqq \{\initialPolicy\}$ \\

    \While{$\mathrm{Queue} \neq \emptyset$}
    {
        Remove from $\mathrm{Queue}$ a policy~$\pi$ with least $f(\pi)$ \\
        \If{\upshape $\pi$ is a solution}
        {
            \textbf{return} $\pi$
        }
        \Else
        {
            \ForEach{\upshape $\pi' \in \successors(\pi)$}
            {
                Insert $\pi'$ in $\mathrm{Queue}$
            }
        }
    }

    \Return{$\bot$} \tcp*[l]{unsolvable task}

    \vspace{2.5mm}
    \hrule
    \vspace{2.5mm}

    \SetKwFunction{Successors}{$\successors$}
    \Pn{\Successors{$\pi$}}
    {
        \If{\upshape $\remain(\pi) \neq \emptyset$}
        {
            Select some state $s$ from $\remain(\pi)$ \\
            \Return{\upshape $\{\pi \sqcup \{s \mapsto a\} : a \in \actions \text{ applicable to }s\}$}
        }
        \Else
        {
            \Return{$\emptyset$}
        }
    }

    \caption{AND$^*$}
    \label{alg:and-star}
\end{algorithm}

The A$^*$ with Non-Determinism (AND$^*$ for short) algorithm~\citep{messa2023best} ``generalizes'' the A$^*$ algorithm~\citep{hart1968formal} for FOND planning. It is a best-first heuristic search algorithm that uses a priority queue over policies to explore the \emph{policy space} of~$\Pi$ and search for a solution policy. It starts with the empty policy~$\emptyset$. At each iteration, it removes from the priority queue a policy~$\pi$ with lowest~$f(\pi)$ value and returns it if it is a solution for~$\Pi$. Otherwise, AND$^*$ expands $\pi$ to generate successor\footnote{We note that there are alternative ways of defining the successors of a given policy~$\pi$. For instance, we could generate instead policies~$\pi' \supseteq \pi$ with $\domain(\pi') = \domain(\pi) \sqcup \remain(\pi)$ mapping \emph{at once} every state in $\remain(\pi)$ to some action. That would result in a ``big-step'' successor function. \citeauthor{messa2023best} \citeyearpar{messa2023best} discuss ``big-step'' vs. ``small-step''.} policies~$\pi' = \pi \sqcup \{s \mapsto a\}$ for some state~$s \in \remain(\pi)$ and each action~$a$ applicable to~$s$ (given that there is one such state~$s$, otherwise $\pi$ is simply discarded). Each generated successor policy is inserted into the priority queue. If the priority queue becomes empty, then there is no solution for~$\Pi$. Algorithm~\ref{alg:and-star} shows the pseudocode for AND$^*$.

\revb{AND$^*$ can be seen as a simple A$^*$ search performed in a different search space: the constructive policy space induced by the custom successor function and initial node AND$^*$ defines.} AND$^*$ is sound and complete~\citep{messa2023best}.

\subsubsection{\revb{Optimality and Heuristics}}

\rev{In FOND planning, one might want not just a solution policy but the most compact one. \revb{Alternatively, one might seek the solution with the minimal expected cost to reach the goal, after introducing a distribution of probabilities for the outcomes of the actions.} In other words, there are multiple possible metrics for FOND planning that one might want to optimize. In principle, the $f(\pi)$ values that AND$^*$ uses to guide its search should evaluate how promising each candidate policy~$\pi$ is for becoming a good solution according to the metric we are trying to optimize. Formally, \cite{messa2023best} show that for any given metric~$\varphi$, AND$^*$ always returns a solution~$\pi_*$ with minimal~$\varphi(\pi_*)$ if the $f$ function used is \emph{admissible} and \emph{goal-aware} in the context of the chosen metric\footnote{Note that for some metrics, such as \emph{maximizing} the domain size ($\varphi(\pi_*) = -|\domain(\pi_*)|$), there exists no $f$ function that simultaneously satisfies the properties of admissibility and goal-awareness.}. They define admissibility and goal awareness for the $f$ function as follows. The function $f$ is admissible iff, for any given policy~$\pi$ that can become a solution through the addition of new mappings (i.e., with at least one solution policy $\pi_* \supseteq \pi$), $f(\pi) \leq \varphi(\pi_*)$ for any solution~$\pi_*$~that~$\pi$~can become. Moreover, $f$ is goal-aware iff $f(\pi) = \varphi(\pi)$~whenever~$\pi$~is a solution.}

Sometimes, it might also be useful to define $g$ and $h^*$ values for candidate policies. When aiming to find a solution policy with minimal domain size, we could say that the $g$-value (the ``already paid cost'') of a given policy~$\pi$ is $|\domain(\pi)|$, and that the $h^*$-value (the ``optimal remaining cost'') of~$\pi$ is the minimum $|\domain(\pi')|$ such that $\pi \sqcup \pi'$ is a solution for $\Pi$ (or $\infty$ if no such $\pi'$ exists). If we have a function~$h$ that estimates~$h^*$ in a never-pessimistic way (that is, $h(\pi) \leq h^*(\pi)$ for any policy $\pi$) and that never returns negative values, then a function $f(\pi) = |\domain(\pi)| + h(\pi)$ is admissible and goal-aware for the domain size metric, and AND$^*$ using such a function only returns solution policies with minimal domain size.

\cite{messa2023best} present a somewhat complex function $f$ called $\deltaNearest$ that employs a classical planning heuristic as a subcomponent and exploits key structural characteristics of FOND planning to evaluate policies focusing on the metric of minimal domain size. If the given classical planning heuristic is admissible (in the classical sense of admissibility \citep[Section 1.6]{edelkamp2011heuristic}), $\deltaNearest$ becomes admissible and goals-aware for the domain size metric. In this work, we conduct most of our experiments using $\deltaNearest$ equipped with the admissible classical planning heuristic $\hLMCut$~\citep{helmert2009landmarks} as AND$^*$'s guiding function, because $\hLMCut$ was the best-performing admissible classical planning heuristic in the experiments of \cite{messa2023best}, and $\deltaNearest$ was the best-performing $f$ function among the ones they presented. In the final sections, we switch to $\deltaNearest$ equipped with the satisficing (non-admissible) classical planning heuristic $\hFF$~\citep{hoffmann2001ff} to maximize coverage, as preserving optimality is no longer of interest there. For more details on $\deltaNearest$, including how it incorporates the use of classical planning heuristics to evaluate policies in the context of FOND planning, we refer readers to \cite{messa2023best}.

\section{Benchmarks and Experimental Settings}
\label{sec:ee}

Throughout this work, we present techniques that can be used to improve AND$^*$'s performance. We evaluate them through empirical experiments.

We use the same two benchmarks as \cite{pereira2022iterative} for our experiments. IPC-FOND, with 379 tasks over 12 FOND planning domains from the International Planning Competition (IPC), and NEW-FOND, introduced by \cite{geffner2018compact}, with 211 tasks over five FOND planning domains. The NEW-FOND benchmark was developed to mislead FOND planners. It has tasks with a large number of trajectories that lead to goal states but are not part of any solution.

As in \cite{messa2023best}, we merge the \emph{blocksworld-2} and \emph{blocks-world-new} domains into a single domain called \emph{blocksworld-advanced}, as they are actually equal domains. Moreover, we remove the tasks without solution, namely 25 tasks from the \emph{first-responders} domain. \rev{These tasks are trivial because even the translator\footnote{\rev{We use PRP's \citep{muise2012improved} translator. It is based on the translator of Fast Downward~\citep{helmert2006fast}, and not just parse tasks written in the PDDL format~\citep{mcdermott20001998}, but also makes some static analysis of the tasks. The time taken by the translator is counted for the time limit of the runs.}} we use to parse the tasks is able to rapidly detect their unsolvability.} Table~\ref{tab:benchmarks} shows the resulting 16 benchmark domains.

\begin{table}[tb]
    \centering
    \fontsize{9}{10.5}\selectfont
    \begin{tabular}{|c|lr}
        \toprule

        \multicolumn{1}{c}{} &
        \textbf{Domain} &
        $\#$ \\

        \midrule
        \multirow{11}{*}{\rotatebox[origin=c]{90}{IPC-FOND}}
         & acrobatics & 8 \\
         & beam-walk & 11 \\
         & blocksworld-original & 30 \\
         & blocksworld-advanced & 55 \\
         & chain-of-rooms & 10 \\
         & earth-observation & 40 \\
         & elevators & 15 \\
         & faults & 55 \\
         & first-responders & 75 \\
         & tireworld-triangle & 40 \\
         & zenotravel & 15 \\
        \midrule
        \multirow{5}{*}{\rotatebox[origin=c]{90}{NEW-FOND}}
         & doors & 15 \\
         & islands & 60 \\
         & miner & 51 \\
         & tireworld-spiky & 11 \\
         & tireworld-truck & 74 \\
        \bottomrule
    \end{tabular}
    \caption{\rev{Resulting benchmark domains.}}
    \label{tab:benchmarks}
\end{table}

We run the experiments on an AMD Ryzen 9 3900X, using limits of $8$ GB RAM and $30$ minutes per task. We use greater $g$-value ($|\pi|$) as the AND$^*$ policy-selection tie-breaker, and we always select to map the most recently added state $s \in \remain(\pi)$ when generating the successors of a policy $\pi$. Additionally, we make AND$^*$ not distinguish between two distinct goal states (i.e., from AND$^*$'s perspective, all generated goal states are the same). This can be seen as a trivial application of the concept of state symmetries. The used code and data are publicly available on Zenodo~\citep{messa_frederico_2026_20027382}.

\section{Policy Equivalence Pruning}
\label{sec:psym}

An equivalence between policies
is any reflexive, symmetric, and transitive binary relation between policies. It has no inherent meaning unless applied in some context. In this work, we study policy equivalences in the context of FOND planning and the search performed by AND$^*$.

Let $\sim$ be a given equivalence relation between policies. We intend to modify AND$^*$ so that if it ever expands a given candidate policy $\pi$, then it will not later expand any other candidate policy~$\pi'$ that is considered equivalent to $\pi$ under $\sim$, effectively pruning part of the search space that AND$^*$ has to explore.

In this work, we are particularly interested in equivalence relations that can enable as much pruning of the search space as possible, without compromising the soundness and completeness of AND$^*$. \revb{In this section, we study two basic concepts of equivalence between policies.}

\subsection{AND$^*$ with Equivalence Pruning}

\rev{In order to employ equivalence pruning, we need to store information about the policies that were already expanded. The naive way to do this would be to store each policy in a set, and then, when expanding a new policy, check if any policy in the set is equivalent to it. However, this would neither be time nor memory-efficient. Therefore, we focus on equivalence relations defined in the following way.
Let $\sign$ be a function that maps policies to objects called \emph{signatures}. Given such a function, we define an equivalence relation $\sim$ between policies by declaring $\pi_1 \sim \pi_2$ iff $\sign(\pi_1) = \sign(\pi_2)$. Note that this way the signature of a policy is the representative of its equivalence class.}

Algorithm~\ref{alg:and-star-ep} shows the pseudocode for AND$^*$ with equivalence pruning. It uses a set called $\done$ to store the signatures of the policies that were already expanded. At the removal of a new policy $\pi$, it checks if $\sign(\pi)$ is in $\done$. If it is, then $\pi$ is skipped. Otherwise, $\sign(\pi)$ is added to $\done$, and $\pi$ is normally expanded.

\begin{algorithm}[t]
    \DontPrintSemicolon
    \SetKwIF{If}{ElseIf}{Else}{if}{:}{else if}{else:}{endif}
    \SetKwFor{While}{while}{:}{endw}
    \SetKwFor{ForEach}{for each}{:}{endfch}

    $\initialPolicy \coloneqq \emptyset$,
    $\mathrm{Queue} \coloneqq \{\initialPolicy\}$ \\

    \While{$\mathrm{Queue} \neq \emptyset$}
    {
        Remove from $\mathrm{Queue}$ some policy~$\pi$ with least $f(\pi)$ \\
        \If{\upshape $\pi$ is a solution}
        {
            \textbf{return} $\pi$
        }
        \ElseIf {$\sign(\pi) \not\in \mathrm{Done}$}
        {
            Insert $\sign(\pi)$ in $\mathrm{Done}$ \\
            \ForEach{\upshape $\pi' \in \successors(\pi)$}
            {
                Insert $\pi'$ in $\mathrm{Queue}$
            }
        }
    }

    \Return{$\bot$} \tcp*[l]{unsolvable task}

    \caption{AND$^*$ with Equivalence Pruning}
    \label{alg:and-star-ep}
\end{algorithm}

Note that Algorithm~\ref{alg:and-star} is a special case of Algorithm~\ref{alg:and-star-ep} because they are equivalent when Algorithm~\ref{alg:and-star-ep}'s $\sign$ is the identity (i.e., $\sign(\pi) = \pi$)\footnote{\label{fnt:no-policy-twice}No policy $\pi$ is generated twice by AND$^*$ \citep{messa2023best}.}.

\subsection{Relation to A$^*$}

\rev{A$^*$~\citep{hart1968formal} is a best-first heuristic search algorithm that uses a priority queue to explore a search space. It starts with just the initial node in the queue. Each node is a pair $n = \langle s, n' \rangle$, with $s$ a state and $n'$ the parent node of $n$ ($\bot$ if $n$ is the initial node), which can be used to reconstruct the path from the initial state to $s$. A$^*$ repeatedly removes from the queue the node $n$ with the lowest $f(n)$ value, until it finds a goal state. When the heuristic function used to compute the $f$-values is admissible and goal-aware, A$^*$ always finds an optimal solution.}
\revb{As described in Subsection~\ref{ssec:and-star}, AND$^*$ can be seen as A$^*$ performed in a policy space instead of state space. However, without the need for the duplicate detection described next.}

\rev{In order to avoid expanding the same states multiple times without need, A$^*$ employs duplicate detection. It uses a map called $\mathtt{Closed}$ to store the states of the nodes that were already expanded. In this map, each state is mapped to the least $g$-value of any already expanded node containing it. Then, when expanding a node $n$, it checks if the state $s$ of $n$ is in $\mathtt{Closed}$. If it is, it expands $n$ only if $g(n) < \mathtt{Closed}[s]$ (in which case it updates $\mathtt{Closed}[s]$ to $g(n)$). Otherwise, it skips $n$. A$^*$ with duplicate detection also ensures optimality. Further improvements can be done in case the heuristic is determined \emph{``consistent''}~\citep{edelkamp2011heuristic}. In that case, A$^*$'s $\mathtt{Closed}$ map can be replaced by simply a set of states, allowing for a stronger pruning, while still ensuring optimality.}

\revb{The AND$^*$ algorithm is said to be a ``generalization'' of A$^*$ for FOND planning~\citep{messa2023best} because it produces the exact same search as a state-space search A$^*$ (without duplicate detection) when the input FOND task is actually also a classical planning task (when all actions have only one outcome possible, and thus are deterministic), while it is additionally able to handle tasks with non-determinism. Then, our intent with the AND$^*$ algorithm with equivalence pruning is to design a best-effort generalization of A$^*$ \emph{with} duplicate detection, for FOND planning.}

\revb{For simplicity, in this work we focus on the duplicate detection version that uses a set instead of a map. 
In this work, we do not study consistency for FOND planning. Our main goal is to evaluate different equivalence relations, while understanding their guarantees.}

Hereafter, we call the $\mathtt{Closed}$ set ``$\done$'' in the context of AND$^*$, to avoid confusion with the property of ``closedness'' of policies. Moreover, when we hereafter mention ``AND$^*$'', we will be referring to Algorithm~\ref{alg:and-star-ep}, not Algorithm~\ref{alg:and-star}, unless explicitly stated otherwise.

\subsection{Which $\sign$ to Use?}

Since using $\sign(\pi) = \pi$ is the same as not having equivalence pruning, we would like to have a better concept of equivalence than identity. \revb{The AND$^*$'s equivalence pruning analogous to looking at the state of an A$^*$ classical planning node (to decide whether to prune it or not) would be to use $\sign(\pi) = \front(\pi)$. This is the case because when a policy~$\pi$ is simply a single trajectory starting from the initial state---as it is the case for every proper policy in the context of classical planning---$\front(\pi)$~is a singleton set containing \emph{precisely} the state that it leads to.}

\revb{Then, the key question is how AND$^*$ would behave, given an arbitrary FOND planning task, when $\sign = \front$. This is the key question because when the input task has only deterministic actions, AND$^*$ with $\sign = \front$ will search exactly like a state-space search A$^*$ with duplicate detection would in classical planning, since that task would also be a classical planning task. It is, however, not clear at the first glance how it would behave when the task \emph{has} non-deterministic actions. Next, we present an example state space to analyze this possibility and show what can go wrong when AND$^*$ with $\sign = \front$ receives FOND tasks that are not classical planning tasks.}

\begin{figure}[ht]
    \centering
    \begin{minipage}{0.45\textwidth}
        \centering
        \scalebox{0.96}
        {
        \begin{tikzpicture}
            \node[black_node]            at ( 0.0,  0.0) (a) {$s_A$};
            \node[black_node]            at ( 0.0, -1.5) (b) {$s_B$};
            \node[black_node]            at (+1.5, -3.0) (c) {$s_C$};
            \node[black_node]            at ( 0.0, -4.5) (d) {$s_D$};
            \node[black_node]            at (-1.5, -3.0) (e) {$s_E$};
            \node[black_node, accepting] at ( 0.0, -6.0) (f) {$s_F$};
            \node[black_node]            at (+2.0, -0.158) (x) {$s_X$};
            \node[black_node, color=white]            at (-2.0, -0.158) (y) {};

            \draw[->, thick] ( 0.0, 0.9) -- (a) ;

            \draw[->, thick] (a) -- (b) ;
            \draw[->, thick] (b) -- (e) ;
            \draw[->, thick] (b) -- (c) ;
            \draw[->, thick] (b) -- (d) ;
            \draw[->, thick] (e) -- (f) ;
            \draw[->, thick] (c.250) to [bend left=20] (d.20) ;
            \draw[->, thick] (d) -- (e) ;
            \draw[->, thick] (d) to [bend left=20] (c) ;
            \draw[->, thick] (a.340) to [bend left=20] (b.45) ;
            \draw[->, thick] (a.340) to (x.180) ;

            \draw[thick] (0.0, -2.5) .. controls (+0.383, -2.424) .. (+0.707, -2.207) ;
            \draw[thick] (0.0, -2.5) .. controls (-0.383, -2.424) .. (-0.707, -2.207) ;

            \draw[thick] (+0.8, -0.158)  .. controls (+0.75, -0.32) and (+0.65, -0.45) .. (+0.49, -0.55) ;

            \node[] at (-0.25, -0.7) () {$a$};
            \node[] at (+1.04, -0.51) () {$a_{bad}$};
            \node[] at (0.18, -2.7) () {$b$};
            \node[] at (1.1, -4.1) () {$c$};
            \node[] at (-0.45, -3.62) () {$d_L$};
            \node[] at (0.38, -3.28) () {$d_R$};
            \node[] at (-1, -4.5) () {$e$};
        \end{tikzpicture}
        }
    \end{minipage}
    \caption{\rev{Example state space to show the incompleteness of AND$^*$ with equivalence pruning using $\sign(\pi) = \escape(\pi)$ or $\sign(\pi) = \langle \domain(\pi), \escape(\pi) \rangle$.}}
    \label{fig:state-space-2}
\end{figure}

 The example state space, depicted in Figure~\ref{fig:state-space-2}, has seven states $s_A$, $s_B$, $s_C$, $s_C$, $s_E$, $s_F$, and $s_X$; and seven actions $a$, $a_{bad}$, $b$, $c$, $d_L$, $d_R$, and $e$.. The initial state is $s_A$, the single goal state is $s_F$, and $s_X$ is a dead-end state.  The only two non-deterministic actions are $b$ (which can lead to either $s_E$, $s_D$, or $s_C$ from $s_B$) and $a_{bad}$ (which can lead to either $s_B$ or $s_X$ from $s_a$).

Let $\pi_L$ and $\pi_R$ be the two policies with domain $\{s_A, s_B, s_D\}$, but differing on which action apply to $s_D$. $\pi_L = \{s_A \mapsto a, s_B \mapsto b, s_D \mapsto d_L\}$ maps $s_D$ to the left action that leads to $s_E$, while $\pi_R = \{s_A \mapsto a, s_B \mapsto b, s_D \mapsto d_R\}$ maps~$s_D$ to the right action that leads to $s_C$. Both $\pi_L$ and $\pi_R$ have $\{s_C, s_E\}$ escape states. Therefore, if $\sign = \front$, we have $\pi_L \sim \pi_R$. But while $\pi_L$ is two mappings from becoming a solution, $\pi_R$ cannot become a solution since the only mapping possible for $s_C$ -- to the action~$c$ -- will lead to a successor policy with a deadlock with $s_D$. Thus, if we set $\sign = \front$, AND$^*$ can discard all solution-leading policies and incorrectly return $\bot$ for a task with solutions. Note that the same problem would have arisen even if we had set $\sign(\pi) = \langle \domain(\pi), \front(\pi) \rangle$ for each policy $\pi$ since $\pi_L \sim \pi_R$ would still be the case.

\subsection{The Concretizer}

We just showed that AND$^*$ using $\sign(\pi) = \front(\pi)$ or $\sign(\pi) = \langle \domain(\pi), \front(\pi) \rangle$ may fail to find a solution for solvable tasks. This happens because two policies $\pi_1$ and $\pi_2$ with the same signature do not necessarily have the same mappings -- and in FOND planning, not only do the escape states matter, but also how we will act after we happen to cycle back to each domain state.

In this subsection, we present a procedure that effectively allows a sound and complete use of $\sign(\pi) = \langle \domain(\pi), \front(\pi) \rangle$ equivalence pruning.
Given only the sets $\langle \domain(\pi), \front(\pi) \rangle$ --- i.e. without the mappings --- this procedure produces a solution policy if the sets match the ones from a solution policy, in time polynomial in $|\domain(\pi)| + |\Pi|$,\footnote{For any policy $\pi$, $|\front(\pi)| \leq (\max_{a \in \actions} |\effs(a)|) \cdot |\domain(\pi)|$, so we do not need to factor the size of $\front$ here.} where $|\Pi|$ is the size of the compact representation of the task, which is usually of the same order of $|\facts| + \sum_{a \in \actions} |\effs(a)|$.
Namely, we present a procedure (Algorithm~\ref{alg:unhollower}), called \emph{the concretizer}, that --- given a domain set $D$ and a escape set $E$ --- produces a \emph{proper} policy $\pi$ with $\domain(\pi) = D$ and $\front(\pi) \subseteq E$ iff exists any, in $O(|D|^2 \cdot |D \cup E| \cdot |\Pi|)$ time\footnote{With a more efficient implementation, the concretizer runs in $O(|D| \cdot |D \cup E| \cdot |\Pi|)$ time.}. Otherwise, it returns~$\bot$.

\begin{algorithm}[ht]
    \DontPrintSemicolon
    \SetKwIF{If}{ElseIf}{Else}{if}{:}{else if}{else:}{endif}
    \SetKwFor{While}{while}{:}{endw}
    \SetKwFor{ForEach}{for each}{:}{endfch}

    \textbf{Input:} $\langle D, E \rangle$ \tcp{a ``hollow'' policy (without mappings) with $D \cap E = \emptyset$}

    $R \coloneqq D \sqcup E$ \\

    $\pi \coloneqq \emptyset$ \\

    $D_{handled} \coloneqq \emptyset$ \\
    $D_{remaining} \coloneqq D$

    \While{$D_{remaining} \neq \emptyset$}
    {
        \If {\upshape $\exists s \in D_{remaining}$ \textbf{and} $\exists a \in \actions\ \text{applicable to}\ s$ \textbf{such that} $\succs(s, a) \cap (E \sqcup D_{handled}) \neq \emptyset$ \textbf{and} $\succs(s, a) \subseteq R$}
        {
            Choose such an $s$ and $a$. \\
            $\pi \coloneqq \pi \sqcup \{s \mapsto a\}$ \\
            $D_{handled} \coloneqq D_{handled} \sqcup \{s\}$ \\
            $D_{remaining} \coloneqq D_{remaining} \setminus \{s\}$
        }
        \Else
        {
            \Return{$\bot$} \tcp*[l]{unconcretizable hollow policy}
        }
    }
    \Return{$\pi$}

    \caption{Policy Concretizer}
    \label{alg:unhollower}
\end{algorithm}

The concretizer starts with an empty policy~$\pi$ and uses a pair of sets $D_{handled}$ and $D_{remaining}$ to keep track of which states of $D$ it has already defined a mapping for. It repeatedly finds some state-action pair $\langle s, a \rangle$ with $s \in D_{remaining}$ such that, if we add $s \mapsto a$ to $\pi$, $\pi$ remains proper, while completely disregarding mappings $s \mapsto a$ such that $\succs(s, a) \not\subseteq E \cup D$. It ensures that the policy $\pi$ remains proper, by ensuring that the mapping to be added $s \mapsto a$ has $\succs(s, a) \cap E \neq \emptyset$ or $\succs(s, a) \cap D_{handled} \neq \emptyset$. That takes into account that $E \cap \domain(\pi)$ is always $\emptyset$, and that, for each $s'$ already in $D_{handled}$, we have $\reach(\pi, s') \cap E \neq \emptyset$ inductively ensured. In the end, $\pi$~will be a proper policy with $\domain(\pi) = D$. Moreover, $\reach(\pi)$ will be a subset of~$D \cup E$ due to the filtering of mappings, meaning that $\front(\pi)$ will be a subset of $E$. We formally prove the soundness and completeness of the concretizer after exemplifying its execution.

\subsubsection{Example}

We use our running example of state space (Figure~\ref{fig:state-space-2}) to exemplify the concretizer execution. We choose $D = \{s_A, s_B, s_C, s_D, s_E\}$ and $E = \{s_F\}$ because there are some ways to go wrong when mapping states in~$D$ and we want to show that, by construction, the concretizer is hindered from including the incorrect mappings. Figure~\ref{fig:state-space-2-concretizer-example-1} illustrates every step of the execution example we will show. It displays in blue the states in $E \cup D_{handled}$, in black the states in $D_{remaining}$, and in red the states outside $E \cup D$. Mappings that were already added to $\pi$ are also shown in blue. Transitions that lead to a blue state are highlighted in green.

\begin{figure}[p]
    \centering
    \setlength\tabcolsep{-2.2pt}
    \begin{tabular}{ccc}
        \scalebox{1}
        {
            \begin{tikzpicture}
                \node[black_node]            at ( 0.0,  0.0) (a) {$s_A$};
                \node[black_node]            at ( 0.0, -1.5) (b) {$s_B$};
                \node[black_node]            at (+1.5, -3.0) (c) {$s_C$};
                \node[black_node]            at ( 0.0, -4.5) (d) {$s_D$};
                \node[black_node]            at (-1.5, -3.0) (e) {$s_E$};
                \node[black_node, accepting, color=blue, very thick] at ( 0.0, -6.0) (f) {$\mathbf{s_F}$};
                \node[black_node, thin, color=red]            at (+2.0, -0.158) (x) {$s_X$};

                \draw[->] (a) -- (b) ;
                \draw[->] (b) -- (e) ;
                \draw[->] (b) -- (c) ;
                \draw[->] (b) -- (d) ;
                \draw[->, very thick, color=green] (e) -- (f) ;
                \draw[->] (c.250) to [bend left=20] (d.20) ;
                \draw[->] (d) -- (e) ;
                \draw[->] (d) to [bend left=20] (c) ;
                \draw[->] (a.340) to [bend left=20] (b.45) ;
                \draw[->] (a.340) to (x.180) ;

                \draw[] (0.0, -2.5) .. controls (+0.383, -2.424) .. (+0.707, -2.207) ;
                \draw[] (0.0, -2.5) .. controls (-0.383, -2.424) .. (-0.707, -2.207) ;

                \draw[] (+0.8, -0.158)  .. controls (+0.75, -0.32) and (+0.65, -0.45) .. (+0.49, -0.55) ;

                \node[] at (-0.25, -0.7) () {$a$};
                \node[] at (+1.04, -0.51) () {$a_{bad}$};
                \node[] at (0.18, -2.7) () {$b$};
                \node[] at (1.1, -4.1) () {$c$};
                \node[] at (-0.45, -3.62) () {$d_L$};
                \node[] at (0.38, -3.28) () {$d_R$};
                \node[] at (-1, -4.5) () {$e$};
            \end{tikzpicture}
        } &
        \scalebox{1}
        {
            \begin{tikzpicture}
                \node[black_node]            at ( 0.0,  0.0) (a) {$s_A$};
                \node[black_node]            at ( 0.0, -1.5) (b) {$s_B$};
                \node[black_node]            at (+1.5, -3.0) (c) {$s_C$};
                \node[black_node]            at ( 0.0, -4.5) (d) {$s_D$};
                \node[black_node, color=blue, very thick]            at (-1.5, -3.0) (e) {$\mathbf{s_E}$};
                \node[black_node, accepting, color=blue, very thick] at ( 0.0, -6.0) (f) {$\mathbf{s_F}$};
                \node[black_node, thin, color=red]            at (+2.0, -0.158) (x) {$s_X$};

                \draw[->] (a) -- (b) ;
                \draw[->, very thick, color=green] (b) -- (e) ;
                \draw[->] (b) -- (c) ;
                \draw[->] (b) -- (d) ;
                \draw[->, very thick, color=blue] (e) -- (f) ;
                \draw[->] (c.250) to [bend left=20] (d.20) ;
                \draw[->, very thick, color=green] (d) -- (e) ;
                \draw[->] (d) to [bend left=20] (c) ;
                \draw[->] (a.340) to [bend left=20] (b.45) ;
                \draw[->] (a.340) to (x.180) ;

                \draw[] (0.0, -2.5) .. controls (+0.383, -2.424) .. (+0.707, -2.207) ;
                \draw[] (0.0, -2.5) .. controls (-0.383, -2.424) .. (-0.707, -2.207) ;

                \draw[] (+0.8, -0.158)  .. controls (+0.75, -0.32) and (+0.65, -0.45) .. (+0.49, -0.55) ;

                \node[] at (-0.25, -0.7) () {$a$};
                \node[] at (+1.04, -0.51) () {$a_{bad}$};
                \node[] at (0.18, -2.7) () {$b$};
                \node[] at (1.1, -4.1) () {$c$};
                \node[] at (-0.45, -3.62) () {$d_L$};
                \node[] at (0.38, -3.28) () {$d_R$};
                \node[color=blue] at (-1, -4.5) () {$\mathbf{e}$};
            \end{tikzpicture}
        } &
        \scalebox{1}
        {
            \begin{tikzpicture}
                \node[black_node]            at ( 0.0,  0.0) (a) {$s_A$};
                \node[black_node, color=blue, very thick]            at ( 0.0, -1.5) (b) {$\mathbf{s_B}$};
                \node[black_node]            at (+1.5, -3.0) (c) {$s_C$};
                \node[black_node]            at ( 0.0, -4.5) (d) {$s_D$};
                \node[black_node, color=blue, very thick]            at (-1.5, -3.0) (e) {$\mathbf{s_E}$};
                \node[black_node, accepting, color=blue, very thick] at ( 0.0, -6.0) (f) {$\mathbf{s_F}$};
                \node[black_node, thin, color=red]            at (+2.0, -0.158) (x) {$s_X$};

                \draw[->, very thick, color=green] (a) -- (b) ;
                \draw[->, very thick, color=blue] (b) -- (e) ;
                \draw[->, thin, color=blue] (b) -- (c) ;
                \draw[->, thin, color=blue] (b) -- (d) ;
                \draw[->, very thick, color=blue] (e) -- (f) ;
                \draw[->] (c.250) to [bend left=20] (d.20) ;
                \draw[->, very thick, color=green] (d) -- (e) ;
                \draw[->] (d) to [bend left=20] (c) ;
                \draw[->, very thick, color=green] (a.340) to [bend left=20] (b.45) ;
                \draw[->] (a.340) to (x.180) ;

                \draw[color=blue] (0.0, -2.5) .. controls (+0.383, -2.424) .. (+0.707, -2.207) ;
                \draw[color=blue] (0.0, -2.5) .. controls (-0.383, -2.424) .. (-0.707, -2.207) ;

                \draw[] (+0.8, -0.158)  .. controls (+0.75, -0.32) and (+0.65, -0.45) .. (+0.49, -0.55) ;

                \node[] at (-0.25, -0.7) () {$a$};
                \node[] at (+1.04, -0.51) () {$a_{bad}$};
                \node[color=blue] at (0.18, -2.7) () {$\mathbf{b}$};
                \node[] at (1.1, -4.1) () {$c$};
                \node[] at (-0.45, -3.62) () {$d_L$};
                \node[] at (0.38, -3.28) () {$d_R$};
                \node[color=blue] at (-1, -4.5) () {$\mathbf{e}$};
            \end{tikzpicture}
        } \\
        {\footnotesize (a)\hspace{5.2mm}} & {\footnotesize (b)\hspace{5.2mm}} & {\footnotesize (c)\hspace{5.2mm}} \\[10pt]
        \scalebox{1}
        {
            \begin{tikzpicture}
                \node[black_node, color=blue, very thick]            at ( 0.0,  0.0) (a) {$\mathbf{s_A}$};
                \node[black_node, color=blue, very thick]            at ( 0.0, -1.5) (b) {$\mathbf{s_B}$};
                \node[black_node]            at (+1.5, -3.0) (c) {$s_C$};
                \node[black_node]            at ( 0.0, -4.5) (d) {$s_D$};
                \node[black_node, color=blue, very thick]            at (-1.5, -3.0) (e) {$\mathbf{s_E}$};
                \node[black_node, accepting, color=blue, very thick] at ( 0.0, -6.0) (f) {$\mathbf{s_F}$};
                \node[black_node, thin, color=red]            at (+2.0, -0.158) (x) {$s_X$};

                \draw[->, very thick, color=blue] (a) -- (b) ;
                \draw[->, very thick, color=blue] (b) -- (e) ;
                \draw[->, thin, color=blue] (b) -- (c) ;
                \draw[->, thin, color=blue] (b) -- (d) ;
                \draw[->, very thick, color=blue] (e) -- (f) ;
                \draw[->] (c.250) to [bend left=20] (d.20) ;
                \draw[->, very thick, color=green] (d) -- (e) ;
                \draw[->] (d) to [bend left=20] (c) ;
                \draw[->, color=gray] (a.340) to [bend left=20] (b.45) ;
                \draw[->, color=gray] (a.340) to (x.180) ;

                \draw[color=blue] (0.0, -2.5) .. controls (+0.383, -2.424) .. (+0.707, -2.207) ;
                \draw[color=blue] (0.0, -2.5) .. controls (-0.383, -2.424) .. (-0.707, -2.207) ;

                \draw[color=gray] (+0.8, -0.158)  .. controls (+0.75, -0.32) and (+0.65, -0.45) .. (+0.49, -0.55) ;

                \node[color=blue] at (-0.25, -0.7) () {$\mathbf{a}$};
                \node[color=gray] at (+1.04, -0.51) () {$a_{bad}$};
                \node[color=blue] at (0.18, -2.7) () {$\mathbf{b}$};
                \node[] at (1.1, -4.1) () {$c$};
                \node[] at (-0.45, -3.62) () {$d_L$};
                \node[] at (0.38, -3.28) () {$d_R$};
                \node[color=blue] at (-1, -4.5) () {$\mathbf{e}$};
            \end{tikzpicture}
        } &
        \scalebox{1}
        {
            \begin{tikzpicture}
                \node[black_node, color=blue, very thick]            at ( 0.0,  0.0) (a) {$\mathbf{s_A}$};
                \node[black_node, color=blue, very thick]            at ( 0.0, -1.5) (b) {$\mathbf{s_B}$};
                \node[black_node]            at (+1.5, -3.0) (c) {$s_C$};
                \node[black_node, color=blue, very thick]            at ( 0.0, -4.5) (d) {$\mathbf{s_D}$};
                \node[black_node, color=blue, very thick]            at (-1.5, -3.0) (e) {$\mathbf{s_E}$};
                \node[black_node, accepting, color=blue, very thick] at ( 0.0, -6.0) (f) {$\mathbf{s_F}$};
                \node[black_node, thin, color=red]            at (+2.0, -0.158) (x) {$s_X$};

                \draw[->, very thick, color=blue] (a) -- (b) ;
                \draw[->, very thick, color=blue] (b) -- (e) ;
                \draw[->, thin, color=blue] (b) -- (c) ;
                \draw[->, thin, color=blue] (b) -- (d) ;
                \draw[->, very thick, color=blue] (e) -- (f) ;
                \draw[->, very thick, color=green] (c.250) to [bend left=20] (d.20) ;
                \draw[->, very thick, color=blue] (d) -- (e) ;
                \draw[->, color=gray] (d) to [bend left=20] (c) ;
                \draw[->, color=gray] (a.340) to [bend left=20] (b.45) ;
                \draw[->, color=gray] (a.340) to (x.180) ;

                \draw[color=blue] (0.0, -2.5) .. controls (+0.383, -2.424) .. (+0.707, -2.207) ;
                \draw[color=blue] (0.0, -2.5) .. controls (-0.383, -2.424) .. (-0.707, -2.207) ;

                \draw[color=gray] (+0.8, -0.158)  .. controls (+0.75, -0.32) and (+0.65, -0.45) .. (+0.49, -0.55) ;

                \node[color=blue] at (-0.25, -0.7) () {$\mathbf{a}$};
                \node[color=gray] at (+1.04, -0.51) () {$a_{bad}$};
                \node[color=blue] at (0.18, -2.7) () {$\mathbf{b}$};
                \node[] at (1.1, -4.1) () {$c$};
                \node[color=blue] at (-0.45, -3.62) () {$\mathbf{d_L}$};
                \node[color=gray] at (0.38, -3.28) () {$d_R$};
                \node[color=blue] at (-1, -4.5) () {$\mathbf{e}$};
            \end{tikzpicture}
        } &
        \scalebox{1}
        {
            \begin{tikzpicture}
                \node[black_node, color=blue, very thick]            at ( 0.0,  0.0) (a) {$\mathbf{s_A}$};
                \node[black_node, color=blue, very thick]            at ( 0.0, -1.5) (b) {$\mathbf{s_B}$};
                \node[black_node, color=blue, very thick]            at (+1.5, -3.0) (c) {$\mathbf{s_C}$};
                \node[black_node, color=blue, very thick]            at ( 0.0, -4.5) (d) {$\mathbf{s_D}$};
                \node[black_node, color=blue, very thick]            at (-1.5, -3.0) (e) {$\mathbf{s_E}$};
                \node[black_node, accepting, color=blue, very thick] at ( 0.0, -6.0) (f) {$\mathbf{s_F}$};
                \node[black_node, thin, color=red]            at (+2.0, -0.158) (x) {$s_X$};

                \draw[->, very thick, color=blue] (a) -- (b) ;
                \draw[->, very thick, color=blue] (b) -- (e) ;
                \draw[->, thin, color=blue] (b) -- (c) ;
                \draw[->, thin, color=blue] (b) -- (d) ;
                \draw[->, very thick, color=blue] (e) -- (f) ;
                \draw[->, very thick, color=blue] (c.250) to [bend left=20] (d.20) ;
                \draw[->, very thick, color=blue] (d) -- (e) ;
                \draw[->, color=gray] (d) to [bend left=20] (c) ;
                \draw[->, color=gray] (a.340) to [bend left=20] (b.45) ;
                \draw[->, color=gray] (a.340) to (x.180) ;

                \draw[color=blue] (0.0, -2.5) .. controls (+0.383, -2.424) .. (+0.707, -2.207) ;
                \draw[color=blue] (0.0, -2.5) .. controls (-0.383, -2.424) .. (-0.707, -2.207) ;

                \draw[color=gray] (+0.8, -0.158)  .. controls (+0.75, -0.32) and (+0.65, -0.45) .. (+0.49, -0.55) ;

                \node[color=blue] at (-0.25, -0.7) () {$\mathbf{a}$};
                \node[color=gray] at (+1.04, -0.51) () {$a_{bad}$};
                \node[color=blue] at (0.18, -2.7) () {$\mathbf{b}$};
                \node[color=blue] at (1.1, -4.1) () {$\mathbf{c}$};
                \node[color=blue] at (-0.45, -3.62) () {$\mathbf{d_L}$};
                \node[color=gray] at (0.38, -3.28) () {$d_R$};
                \node[color=blue] at (-1, -4.5) () {$\mathbf{e}$};
            \end{tikzpicture}
        } \\
        {\footnotesize (d)\hspace{5.2mm}} & {\footnotesize (e)\hspace{5.2mm}} & {\footnotesize (f)\hspace{5.2mm}}
    \end{tabular}
    \caption{\rev{Illustration of the concretizer execution example.}}
    \label{fig:state-space-2-concretizer-example-1}
\end{figure}

Initially, $D_{handled} = \emptyset$. Therefore, in order for a mapping $s' \mapsto a'$ to be added to $\pi$, $\succs(s', a')$ must include $s_F$, since $F = \{s_F\}$. The concretizer determines that $s_E \mapsto e$ is only the mapping that fulfills that requirement (Figure~\ref{fig:state-space-2-concretizer-example-1}a). Moreover, $\succs(s_E, e) \subseteq D \cup F$ is also true (the other requirement). Therefore, the concretizer adds $s_E \mapsto e$ to~$\pi$. Now $D_{handled} = \{s_E\}$. Therefore, the requirement of $\succs(s, a) \cap \{s_F\} \neq \emptyset$ is weakened to $\succs(s', a') \cap \{s_E, s_F\} \neq \emptyset$, allowing candidate mappings $s_B \mapsto b$ and $s_D \mapsto d_L$ to be added (Figure~\ref{fig:state-space-2-concretizer-example-1}b). Mapping $s_D$ to $d_L$ avoids the problem of deadlock that would happen if we map $s_D$ to $d_R$. In order to show that the mapping $s_D$ to $d_R$ will never be an option, we make the concretizer choose not to add $s_D \mapsto d_L$ yet, but first $s_B \mapsto b$. Now, $\pi = \{s_B \mapsto b, s_E \mapsto e\}$, and the requirement is weakened to $\succs(s', a') \cap \{s_B, s_E, s_F\} \neq \emptyset$. Three mappings meet the weakened requirement, $s_D \mapsto d_L$ (as it already met its stronger version), $s_A \mapsto a$, and $s_A \mapsto a_{bad}$ (Figure~\ref{fig:state-space-2-concretizer-example-1}c). Another way the concretizer could go wrong was if it decided to add $s_A \mapsto a_{bad}$. However, this is the only mapping that does not meet the other requirement: $\succs(s_A, a_{bad}) = \{s_B, s_X\} \not\subseteq (D \cup F) = \{s_A, s_B, s_C, s_D, s_E, s_F\}$. Therefore, $s_A \mapsto a_{bad}$ is discarded. The candidates are $s_D \mapsto d_L$ and $s_A \mapsto a$. Let's say the concretizer decides to delay adding $s_D \mapsto d_L$ once more, and adds $s_A \mapsto a$ to $\pi$. Now $\pi = \{s_A \mapsto a, s_B \mapsto b, s_E \mapsto e\}$, and the requirement is weakened to $\succs(s', a') \cap \{s_A, s_B, s_E, s_F\} \neq \emptyset$. Now, the only mapping that meets both requirements is $s_D \mapsto d_L$ (Figure~\ref{fig:state-space-2-concretizer-example-1}d). The mapping of $s_D \mapsto d_R$ that would construct an improper policy with the $s_D$ and $s_C$ deadlock was never available. Since $s_D \mapsto d_R$ leads only to $s_C$, it would only be available if we had handled $s_C$. Given that we could not, we have no option but to map $s_D$ to another action, one that leads to some handled state. $s_D \mapsto d_L$ is added to~$\pi$. Now, $\pi = \{s_A \mapsto a, s_B \mapsto b, s_D \mapsto d_L, s_E \mapsto e\}$. Finally, $s_C \mapsto c$ meets the requirements, as $s_D$ is now in $\domain(\pi) = D_{handled}$ (Figure~\ref{fig:state-space-2-concretizer-example-1}e). The concretizer adds $s_C \mapsto c$ to $\pi$ and returns because all states in $D$ are mapped (Figure~\ref{fig:state-space-2-concretizer-example-1}f). The returned policy $\pi = \{s_A \mapsto a, s_B \mapsto b, s_C \mapsto c, s_D \mapsto d_L, s_E \mapsto e\}$ is proper, has $\domain(\pi) = D$, and $\front(\pi) \subseteq F$.

\subsubsection{Theoretical Properties}

We now prove the soundness and completeness of the concretizer.

\begin{lemma} \label{lma:concretizer-termination}
    The concretizer always terminates.
\end{lemma}

\textit{Proof.} At each iteration, either the concretizer ends returning $\bot$, or it decreases the size of $D_{remaining}$ by one. Thus, since the input sets are finite, the number of iterations is finite. Then, since each instruction takes a finite time, each execution of the concretizer must eventually terminate.$~\hfill\square$

\begin{lemma} \label{lma:concretizer-no-false-positives-explicit}
    If the concretizer returns a policy $\pi$, then it is proper, $\domain(\pi) = D$, and $\front(\pi) \subseteq E$.
\end{lemma}

\textit{Proof.} The return must have occurred at Line~14. We show by induction on the iterations of the concretizer that $\pi$ had $\domain(\pi) \subseteq D$, $\reach(\pi) \subseteq D \cup E$ and $\forall s \in \domain(\pi), \reach(\pi, s) \cap E \neq \emptyset$ (implying properness) throughout the entire execution. By proving that, we will achieve the desired conclusion, taking into account that when Line~14 is reached, $D_{remaining} = D \setminus \domain(\pi) = \emptyset$, and therefore $\domain(\pi) = D$.

\begin{itemize}
    \item Before the first iteration, the hypothesis is trivially true since $\pi = \emptyset$.
    \item We show that if, at the beginning of an iteration, the hypothesis is true, then it must also be at the end of that iteration. Note that since $\bot$ was never returned, then the if-statement was satisfied at each iteration.
    \begin{itemize}
        \item Given that $\domain(\pi) \subseteq D$, then $\domain(\pi \cup \{s \mapsto a\}) \subseteq D$ by choice of $s \in D_{remaining} = D \setminus \domain(\pi)$.
        \item Given that $\reach(\pi) \subseteq D \cup E$, then $\reach(\pi \cup \{s \mapsto a\}) \subseteq D \cup E$ by choice of $s$ and $a$ with $\succs(s, a) \subseteq D \cup E$.
        \item Given that each $s' \in \domain(\pi)$ $\pi$-reaches a state in $E$, then each $s' \in \domain(\pi \cup \{s \mapsto a\})$ $(\pi \cup \{s \mapsto a\})$-reaches a state in $E$ by choice of $s$ and $a$ with $\succs(s, a) \cap (E \cup \domain(\pi)) \neq \emptyset$ (since $\domain(\pi) = D_{handled}$).$~\hfill\square$
    \end{itemize}
\end{itemize}

\begin{lemma} \label{lma:concretizer-no-false-positives}
    The concretizer is correct whenever there is no proper policy $\pi'$ with $\domain(\pi') = D$ and $\front(\pi') \subseteq E$.
\end{lemma}

\textit{Proof.} By contraposition, assume it is not correct. Then, for some input $\langle D, E \rangle$ for which there is no proper policy $\pi'$ with $\domain(\pi') = D$ and $\front(\pi') \subseteq E$, it returns something different from $\bot$ (otherwise it would be correct). Thus, it returns some policy $\pi$ at Line~14. Then, by applying Lemma~\ref{lma:concretizer-no-false-positives-explicit}, we obtain a contradiction with our assumption of the nonexistence of a proper policy $\pi'$ with $\domain(\pi') = D$ and $\front(\pi') \subseteq E$.$~\hfill\square$

\begin{lemma} \label{lma:concretizer-no-false-negatives}
    The concretizer is correct whenever there is some proper policy $\pi'$ with $\domain(\pi') = D$ and $\front(\pi') \subseteq E$.
\end{lemma}

\textit{Proof.} By contraposition, assume it is not correct. Then, for some input $\langle D, E \rangle$ for which there is some proper policy $\pi'$ with $\domain(\pi') = D$ and $\front(\pi') \subseteq E$, it returns something different from a policy with such properties. Then, by Lemma~\ref{lma:concretizer-no-false-positives-explicit}, it must have returned no policy at all. The only alternative is that it returned $\bot$ at Line~13. Then, at the beginning of some iteration it occurred that did not exist $s \in D_{remaining} = D \setminus \domain(\pi)$ and $a \in \actions$ with $a$ applicable~to~$s$, $\succs(s, a) \cap (E \cup \domain(\pi)) \neq \emptyset$, and $\succs(s, a) \subseteq D \cup E$. We show that that could not have occurred under the current assumptions. Let $\pi'' = \{s \mapsto \pi'[s] : s \in \domain(\pi') \setminus \domain(\pi)\} = \{s \mapsto \pi'[s] : s \in D \setminus \domain(\pi)\}$. Since $\pi'' \subseteq \pi'$ and $\pi'$ is proper, $\pi''$ is proper. Therefore $\open(\pi'') \neq \emptyset$ and $\exists s \in \domain(\pi'')$ such that $\succs(s, \pi[s]) \cap \open(\pi'') \neq \emptyset$. We choose such a state $s$ and action~$a = \pi[s]$, and we note that $\succs(s, a) \subseteq D \cup E$ because $\reach(\pi') \subseteq D \cup E$ and $\pi'[s] = \pi''[s] = a$ (by construction of $\pi''$). Moreover, since $\succs(s, a) \cap \open(\pi'') \neq \emptyset$ and $\open(\pi'') = \reach(\pi'') \setminus \domain(\pi'') \subseteq \reach(\pi') \setminus \domain(\pi'') \subseteq (D \cup E) \setminus (D \setminus \domain(\pi)) \subseteq E \cup \domain(\pi)$, then it must be true that $\succs(s, a) \cap (E \cup \domain(\pi)) \neq \emptyset$. Therefore, there exists $s \in D \setminus \domain(\pi)$ and $a \in \actions$ applicable~to~$s$, with $\succs(s, a) \cap (E \cup \domain(\pi)) \neq \emptyset$ and $\succs(s, a) \subseteq D \cup E$. Contradiction.$~\hfill\square$

\begin{theorem} \label{thm:concretizer-sound-and-complete}
    The concretizer is sound and complete.
\end{theorem}

\textit{Proof.} By combination of Lemmas~\ref{lma:concretizer-termination}, \ref{lma:concretizer-no-false-positives}, and~\ref{lma:concretizer-no-false-negatives}.$~\hfill\square$

\subsection{Domain-Escape Equivalence Pruning}
\label{ssec:psym-io}

\color{revbcolor}

The main consequence of the existence of the concretizer is that we do not need to find a solution $\pi_*$ during a policy-space search. It suffices to find the domain set and a superset of the escape set of some solution $\pi_*$, if the second set is composed only of goal states. We prove that if we find such sets, we can call the concretizer to obtain a solution. First, it will not return~$\bot$ due to the existence of $\pi_*$. Then, let $\pi'$ be the proper policy it returns. Since the provided~$E$ set is composed only of goal states, $\pi'$ is goal-closed. Finally, we know the returned policy handles the initial state because $\pi_*$ does so, and they share the same domain set. Therefore, $\pi'$ is a solution.

The following question is how we can find such sets. As one can expect, finding a valid pair $\langle D, E \rangle$ that satisfies the condition should be a hard task. While we could change the constructive search space to explicit search for those pairs, we decided to proceed with the same policy search space and use the $\domain$ and $\front$ sets of candidate policies as potential guesses. Importantly, we prove a key result in this section: in order to make AND$^*$ sound and complete with domain-escape equivalence pruning, we only need to run the concretizer on $\langle \domain(\pi), \front(\pi) \rangle$ for the visited policies~$\pi$ that have $\remain(\pi) = \emptyset$ (Algorithm~\ref{alg:extract-solution-unhollower}).
As a bonus, if the concretizer does not return $\bot$, it is guaranteed in such cases that the policy it returned is a solution, so we do not really need to check it is a solution.

\begin{algorithm}[t]
    \color{revbcolor}
    \DontPrintSemicolon
    \SetKwIF{If}{ElseIf}{Else}{if}{:}{else if}{else:}{endif}
    \SetKwFor{While}{while}{:}{endw}
    \SetKwFor{ForEach}{for each}{:}{endfch}

    $\initialPolicy \coloneqq \emptyset$,
    $\mathrm{Queue} \coloneqq \{\initialPolicy\}$ \\

    \While{$\mathrm{Queue} \neq \emptyset$}
    {
        Remove from $\mathrm{Queue}$ some policy~$\pi$ with least $f(\pi)$ \\
        \If {\upshape $\remain(\pi) = \emptyset$}
        {
            \If {\upshape $\pi$ is a solution}
            {
                \Return{$\pi$}
            }
            \If {\upshape $\concretizer(\domain(\pi), \front(\pi))$ is a solution}
            {
                \Return{$\concretizer(\domain(\pi), \front(\pi))$}
            }
        }
        \If {$\sign(\pi) \not\in \mathrm{Done}$}
        {
            Insert $\sign(\pi)$ in $\mathrm{Done}$ \\
            \ForEach{\upshape $\pi' \in \successors(\pi)$}
            {
                Insert $\pi'$ in $\mathrm{Queue}$
            }
        }
    }

    \Return{$\bot$} \tcp*[l]{unsolvable task}

    \caption{AND$^*$ with equivalence pruning and the concretizer}
    \label{alg:extract-solution-unhollower}
\end{algorithm}

\begin{figure}[t]
    \centering
    \begin{tikzpicture}
        \node[black_rec_node] at (0, 0) (I) {$\initialPolicy$, $\emptyset$, $\emptyset$};
        \node[black_rec_node] at (0, -1.5) (1) {$\pi_1$, $\{s_A\}$, $\{s_B\}$};
        \node[black_rec_node] at (+4, -1) (2) {$\pi_2$, $\{s_A\}$, $\{s_X\}$};
        \node[black_rec_node] at (0, -3) (3) {$\pi_3$, $\{s_A, s_B\}$, $\{s_C, s_D, s_E\}$};
        \node[black_rec_node] at (-3, -4.5) (L) {$\pi_L$, $\{s_A, s_B, s_D\}$, $\{s_C, s_E\}$};
        \node[black_rec_node] at (+3, -4.5) (R) {$\pi_R$, $\{s_A, s_B, s_D\}$, $\{s_C, s_E\}$};
        \node[black_rec_node] at (-3.5, -6.6) (??) {\begin{tabular}{c}
            ?, $\{s_A, s_B, s_D, s_E\}$, $\{s_C\}$ \\
            or \\
            ?, $\{s_A, s_B, s_C, s_D\}$, $\{s_E\}$
        \end{tabular}};
        \node[black_rec_node] at (+3.5, -6.6) (67) {\begin{tabular}{c}
            $\pi_6$, $\{s_A, s_B, s_D, s_E\}$, $\{s_C\}$ \\
            or \\
            $\pi_7$, $\{s_A, s_B, s_C, s_D\}$, $\{s_E\}$
        \end{tabular}};
        \node[black_rec_node, accepting] at (-4, -8.7) (*) {$\pi_*$, $\{s_A, s_B, s_C, s_D, s_E\}$, $\{s_F\}$};
        \node[black_rec_node] at (+4, -8.7) (8) {$\pi_8$, $\{s_A, s_B, s_C, s_D, s_E\}$, $\{s_F\}$};

        \draw[->, thick] (0, 0.9) -- (I) ;

        \draw[->, thick] (I) -- (1) ;
        \draw[->, thick] (I) -- (2) ;
        \draw[->, thick] (1) -- (3) ;
        \draw[->, thick] (3) -- (L) ;
        \draw[->, thick] (3) -- (R) ;
        \draw[->, thick] (L) -- (??) ;
        \draw[->, thick] (R) -- (67) ;
        \draw[->, thick] (??) -- (*) ;
        \draw[->, thick] (67) -- (8) ;

        \draw[-, very thick, color=red] (-6, -5) -- (0, -4) ;
        \draw[-, very thick, color=red] (-6, -4) -- (0, -5) ;

        \draw[->, ultra thick, dashed, color=Green] (8.188.9) to [bend left=50] (*.351) ;

        \node[] at (0, -4.5) () {$\sim$};
        \node[] at (0, -6.2) () {$\sim$};
        \node[] at (0, -7.0) () {$\sim$};
        \node[] at (0, -8.7) () {$\sim$};
        \node[color=Green] at (0, -10.05) () {$\concretizer$};
    \end{tikzpicture}
    \caption{\rev{The policy space from the execution example that illustrates how the use of the concretizer fixes the issues of AND$^*$ using domain-escape.}}
    \label{fig:policy-space}
\end{figure}

We revisit our running example from Figure~\ref{fig:state-space-2} to show that, by using the concretizer as described in Algorithm~\ref{alg:extract-solution-unhollower}, we find the solution even if we use $\sign(\pi) = \langle \domain(\pi), \front(\pi) \rangle$. \rev{We note the reader might find useful to follow Figure~\ref{fig:policy-space} while reading the following example.}
$\pi_\initialCondition = \emptyset$ is the initial policy. It has two successors $\pi_1 = \{s_A \mapsto a\}$ and $\pi_2 = \{s_A \mapsto a_{bad}\}$. $\pi_2$ leads to no successor because there is no applicable action for $s_X$ and $\remain(\pi_2) = \{s_X\}$. $\pi_1$ has a single successor $\pi_3 = \{s_A \mapsto a, s_B \mapsto b\}$. $\remain(\pi_3) = \front(\pi_3) = \{s_C, s_D, s_E\}$. AND$^*$ arbitrarily chooses a state in $\remain(\pi_3)$ to map. In order to reach the problematic situation we previously showed, let's say AND$^*$ chooses to map first $s_D$. $\pi_3$ generates two successors $\pi_L = \{s_A \mapsto a, s_B \mapsto b, s_D \mapsto d_L\}$ and $\pi_R = \{s_A \mapsto a, s_B \mapsto b, s_D \mapsto d_R\}$. $\pi_L$ and $\pi_R$ are the two policies we have brought attention to in previous discussions. Note that $\pi_L \sim \pi_R$, since they have the same $\domain$ and $\front$ sets. While $\pi_L$ can become a solution, $\pi_R$ cannot because it will have a deadlock on $s_D$ and $s_C$ when we eventually map $s_C \in \remain(\pi_R)$ to $c$. If we try to expand $\pi_R$ before $\pi_L$, we will prune $\pi_L$. This was a critical problem before the concretizer. Now, it causes no harm. After expanding $\pi_R$, we generate either $\pi_6 = \{s_A \mapsto a, s_B \mapsto b, s_D \mapsto d_R, s_C \mapsto c\}$ or $\pi_7 = \{s_A \mapsto a, s_B \mapsto b,\break s_D \mapsto d_R, s_E \mapsto e\}$, since $\remain(\pi_R) = \{s_C, s_E\}$. Regardless of what the successor of $\pi_R$ is, the successor of its successor will be $\pi_8 = \{s_A \mapsto a, s_B \mapsto b,\break s_C \mapsto c, s_D \mapsto d_R, s_E \mapsto e\}$. $\pi_8$ is the first policy to have an empty $\remain$ set. It has an empty $\remain$ set because it is goal-closed (all states in $\front(\pi_8) = \{s_F\}$ are goal states), and it deals with the initial state ($s_A \in \domain(\pi_8)$). The only reason it is not a solution is that it is not proper, as it has a deadlock on $c$ and $d$. Since $\remain(\pi_8) = \emptyset$, the concretizer will be called on $\langle \domain(\pi_8), \front(\pi_8) \rangle = \langle \{s_A, s_B, s_C, s_D, s_E\}, \{s_F\} \rangle$, deducing the solution $\pi_* = \{s_A \mapsto a, s_B \mapsto b, s_C \mapsto c, s_D \mapsto d_L, s_E \mapsto e\}$\footnote{The process of deducting $\{s_A \mapsto a,\; s_B \mapsto b,\; s_C \mapsto c,\; s_D \mapsto d_L,\; s_E \mapsto e\}$ from $\langle \{s_A, s_B, s_C, s_D, s_E\}, \{s_F\} \rangle$ is displayed as an example earlier on this section.}, which $\pi_L$ (the policy pruned by $\pi_R$) would lead to if not pruned. This is the main benefit of having the concretizer. If the only problem that makes a policy not a solution is improperness, and its $\domain$ and $\front$ sets are correct, the concretizer will fix the properness by finding the correct mappings. Figure~\ref{fig:policy-space} illustrates how using the concretizer fixes what previously was considered a mis-pruning. It displays the $\domain$ and $\front$ set of each policy on the generated policy space.

\subsubsection{Theoretical Properties}

We now formally prove that AND$^*$ with the concretizer is sound and complete when using domain-escape equivalence pruning. We also prove that it returns solutions with minimal domain size when using the~$\deltaNearest(\hLMCut)$ heuristic.

Arguing that AND$^*$ always either fails or returns solutions, (absence of false positives) is fairly straightforward, regardless of the equivalence relation used. The difficult part is proving that it always returns a solution when the given task is solvable (absence of false negatives), despite the introduced pruning.
One of the key ideas for proving the latter will be to show that if there exists a sequence of mappings that transforms one given policy~$\pi_1$ into a solution, then the same sequence of mappings also transforms any equivalent policy~$\pi_2 \equiv \pi_1$ into a policy where the concretizer is able to retrieve a solution. With that, we can show that whenever we prune a policy that would lead to a solution, there is another policy, with the same signature (and thus same size), that also leads to a solution and that was successfully expanded before.

\begin{lemma} \label{lma:termination-and-soundness}
    Regardless of the choice of $\sign$, AND$^*$ never returns false positives (i.e., a policy that is not a solution). Moreover, if $\sign$ can be computed in finite time, AND$^*$ always terminates.
\end{lemma}

\revb{\textit{Proof.} By design, AND$^*$ only has return statements that return solutions. Moreover, it always terminates because the number of successors of a policy is finitely bounded by $|\actions|$, the maximum size a policy may have is $|S|$, and every successor of a policy has an incremented size.$~\hfill\square$}

\begin{definition} \label{def:certificate-io}
    Let a solution policy be optimal if it has minimal domain size among all solution policies. We call a policy~$\pi$ a ``good candidate policy'' iff there exists a policy~$\pi'$ subset of an optimal solution policy~$\pi_*$ (i.e. $\pi' \subseteq \pi_*$) with $\domain(\pi') = \domain(\pi)$ and $\front(\pi') = \front(\pi)$. In such a case, we denote $P(\pi) = \top$. Otherwise, we denote $P(\pi) = \bot$.
\end{definition}

\begin{lemma} \label{lma:continuity-io}
    For any given policy~$\pi$, if $P(\pi) = \top$, then either Algorithm~\ref{alg:extract-solution-unhollower} returns a solution when evaluating $\pi$, or $\exists \pi' \in \successors(\pi)$ such that $P(\pi') = \top$.
\end{lemma}

\textit{Proof.} Assume, for the sake of contradiction, that neither Algorithm~\ref{alg:extract-solution-unhollower} returns a solution when evaluating $\pi$, nor $\pi$ has a successor $\pi'$ with $P(\pi') = \top$.
Let $\pi_B$ be the policy subset of an optimal solution policy~$\pi_*$ with $\domain(\pi_B) = \domain(\pi)$ and $\front(\pi_B) = \front(\pi)$, that exists due to $P(\pi) = \top$.
Firstly, since Algorithm~\ref{alg:extract-solution-unhollower} does not return a solution when evaluating $\pi$, then $\pi$ is not a solution and either $\remain(\pi) \neq \emptyset$ or the concretizer returns $\bot$ for $\langle \domain(\pi), \front(\pi) \rangle$. However, the latter could not have happened because of the existence of~$\pi_B$, which is proper as it is a subset of a solution, so $\remain(\pi) \neq \emptyset$ must be the case.
Then, $\exists s \in \remain(\pi)$ such that for every action $a$ applicable to $s$, $\pi \sqcup \{s \mapsto a\} \in \successors(\pi)$.
Note that such state $s$ must be in $\domain(\pi_*)$, as it is in $\remain(\pi_B) = \remain(\pi)$ and $\pi_B \subseteq \pi_*$. Otherwise $s \in \remain(\pi_*)$ and $\pi_*$ would not be a solution.
Let $a' = \pi_*[s]$ and let $\pi' = \pi \sqcup \{s \mapsto a'\}$. $\pi'$ is in $\successors(\pi)$.
Moreover, let $\pi'_B = \pi_B \sqcup \{s \mapsto a'\} \subseteq \pi_*$. $\domain(\pi'_B) = \domain(\pi')$ and $\front(\pi'_B) = \front(\pi'_B)$.
Therefore, $P(\pi') = \top$. Contradiction.$~\hfill\square$

\begin{lemma} \label{lma:invariant-io}
    If two policies $\pi_1$ and $\pi_2$ are equivalent according to domain-escape, then $P(\pi_1) = P(\pi_2)$ and they have the same domain size.
\end{lemma}

\textit{Proof.} Trivial, since $P$ only depends on the domain and escape sets of a policy.$~\hfill\square$

\begin{note} \label{note:delta-works}
    We note that for two policies $\pi_1$ and $\pi_2$ with $\domain(\pi_1) = \domain(\pi_2)$:
    \begin{enumerate}
        \item If $\front(\pi_1) = \front(\pi_2)$, then $\deltaNearest(\pi_1) = \deltaNearest(\pi_2)$.
        \item If $\front(\pi_1) \subseteq \front(\pi_2)$, then $\deltaNearest(\pi_1) \leq \deltaNearest(\pi_2)$.
    \end{enumerate}
    Trivially from the definition of $\deltaNearest$~\citep{messa2023best}.
\end{note}

\begin{theorem} \label{thm:sound-and-complete-io}
    AND$^*$ with $\sign(\pi) = \langle \domain(\pi), \front(\pi) \rangle$ using the concreti-zer is sound and complete. Moreover, it returns solutions with minimal domain size when guided by the $\deltaNearest$ heuristic using $\hLMCut$.
\end{theorem}

\textit{Proof.} By Lemma~\ref{lma:termination-and-soundness}, we know AND$^*$ without the concretizer always terminates and never returns false-positives. This does not change with the use of the concretizer, since the concretizer always terminates (Lemma~\ref{lma:concretizer-termination}) and since we still always check if a policy is a solution before returning it. Therefore, we only need to prove that it always returns a solution when the task is solvable to ensure its soundness and completeness. Assume, for the sake of contradiction, that, for some solvable task, it does not return a solution. Then, it must have returned precisely $\bot$ (Algorithm~\ref{alg:extract-solution-unhollower}'s Line~13), as it never returns false-positives. That means that every policy $\pi$ that was inserted in $\queue$ was removed from it. Consider the policy~$\pi$ with $P(\pi) = \top$ with maximal domain size that was removed from $\queue$ (there is at least one since $P(\emptyset) = \top$, as $\emptyset$ is a subset of any solution). If there is more than one such policy, consider the first one that was removed. Let's look for the moment it was removed from $\queue$. By Lemma~\ref{lma:invariant-io}, it must also not have been pruned, since that would imply that there was another policy~$\pi'$ with $P(\pi') = \top$ and same domain size that was removed from $\queue$ before~$\pi$. We also know that AND$^*$ with the concretizer did not return any solution when analyzing $\pi$, since $\bot$ was the return value. Then, each policy in $\successors(\pi)$ was inserted in $\queue$. However, at least one $\pi' \in \successors(\pi)$ must have $P(\pi') = \top$ (by Lemma~\ref{lma:continuity-io}). And it must have been removed from $\queue$ at some point. However, such a $\pi'$ has domain size greater than $\pi$'s, contradicting the choice of $\pi$. This proves that AND$^*$ must return a solution when the task is solvable, and therefore it is sound and complete.

Now, let us prove the optimality. Assume, for the sake of contradiction, that there is a solution with domain size $k$, but it returns a suboptimal solution~$\pi_*'$ with $|\domain(\pi_*')| > k$. Let $\pi'$ be the policy that AND$^*$ with the concretizer analyzed in order to return $\pi_*'$. Then, either $\pi' = \pi_*'$ or $\pi'$ is a policy with $\domain(\pi') = \domain(\pi_*)$ and $\front(\pi') \supseteq \front(\pi_*)$. Then, we note that $\deltaNearest(\pi') \geq |\domain(\pi_*')|$ due to the heuristic goal-awareness plus Note~\ref{note:delta-works}.2.
Then, all policies $\pi$ with $\deltaNearest(\pi) \leq k < |\domain(\pi_*')| \leq \deltaNearest(\pi')$ inserted in $\queue$ must have been eventually removed from it, as of Algorithm~\ref{alg:extract-solution-unhollower}'s Line~3's best-first choices. Therefore, all policies $\pi$ with $P(\pi) = \top$ inserted in $\queue$ were eventually removed from it, because they must have $\deltaNearest(\pi) \leq k$, by the heuristic admissibility, Note~\ref{note:delta-works}.1, and the fact that a policy $\pi$ only has $P(\pi) = \top$ if it has the same domain and escape as a policy that is a subset of some optimal solution (by the definition of $P$). The remainder of the proof follows equal to the previous paragraph. We consider some policy $\pi$ with maximal domain size that has $P(\pi) = \top$ and was once removed from $\queue$, and we reach a contradiction in the end.
$~\hfill\square$

\color{black}

\subsubsection{Empirical Evaluation}

\color{revbcolor}

\begin{table}[hp]\color{revbcolor}
\begin{subtable}[t]{\textwidth}
    \centering
    \setlength\tabcolsep{5pt}
    \fontsize{9}{10}\selectfont
    \begin{adjustbox}{center}
    \begin{tabular}{l|ccc|ccc}
        \toprule
        \multicolumn{1}{c}{} & \multicolumn{3}{c}{identity} & \multicolumn{3}{c}{domain-escape} \\
        \multicolumn{1}{c}{} & \multicolumn{3}{c}{\textit{(without the concretizer)}} & \multicolumn{3}{c}{} \\

        \cmidrule[\heavyrulewidth](lr){2-4} \cmidrule[\heavyrulewidth](lr){5-7}

        \textbf{Domain} & $\%\mathrm{S}$ & $\%\mathrm{T}$ & $\%\mathrm{M}$ & $\%\mathrm{S}$ & $\%\mathrm{T}$ & $\%\mathrm{M}$ \\

        \midrule
        acrobatics & 0.50 & \textcolor{gray}{0} & 0.50 & \textbf{1} & \textcolor{gray}{0} & \textcolor{gray}{0} \\
        beam-walk & 0.82 & 0.18 & \textcolor{gray}{0} & 0.82 & 0.18 & \textcolor{gray}{0} \\
        blocksworld-original & 0.47 & 0.33 & 0.20 & 0.50 & 0.33 & 0.17 \\
        blocksworld-advanced & 0.20 & 0.60 & 0.20 & 0.20 & 0.64 & 0.16 \\
        chain-of-rooms & 0.30 & \textcolor{gray}{0} & 0.70 & 0.30 & \textcolor{gray}{0} & 0.70 \\
        earth-observation & 0.20 & \textcolor{gray}{0} & 0.80 & 0.20 & \textcolor{gray}{0} & 0.80 \\
        elevators & 0.47 & \textcolor{gray}{0} & 0.53 & 0.47 & \textcolor{gray}{0} & 0.53 \\
        faults & 0.42 & \textcolor{gray}{0} & 0.58 & 0.40 & \textcolor{gray}{0} & 0.60 \\
        first-responders & 0.44 & \textcolor{gray}{0} & 0.56 & 0.48 & \textcolor{gray}{0} & 0.52 \\
        tireworld-triangle & \textbf{1} & \textcolor{gray}{0} & \textcolor{gray}{0} & \textbf{1} & \textcolor{gray}{0} & \textcolor{gray}{0} \\
        zenotravel & 0.33 & 0.67 & \textcolor{gray}{0} & 0.33 & 0.67 & \textcolor{gray}{0} \\
        \midrule
        doors & \textbf{1} & \textcolor{gray}{0} & \textcolor{gray}{0} & 0.87 & \textcolor{gray}{0} & 0.13 \\
        islands & 0.38 & \textcolor{gray}{0} & 0.62 & 0.42 & \textcolor{gray}{0} & 0.58 \\
        miner & 0.10 & 0.02 & 0.88 & 0.10 & 0.02 & 0.88 \\
        tireworld-spiky & 0.45 & \textcolor{gray}{0} & 0.55 & 0.45 & \textcolor{gray}{0} & 0.55 \\
        tireworld-truck & 0.16 & \textcolor{gray}{0} & 0.84 & 0.18 & \textcolor{gray}{0} & 0.82 \\
        \midrule
        TOTAL & 0.453 & 0.113 & 0.435 & 0.482 & 0.115 & 0.403 \\
        \bottomrule
    \end{tabular}
    \end{adjustbox}
    \caption{Ratio of successful, timeout, and memory-out runs per configuration.}
\end{subtable}

\bigskip

\begin{subtable}[t]{\textwidth}
    \centering
    \setlength\tabcolsep{5pt}
    \fontsize{9}{10}\selectfont
    \begin{adjustbox}{center}
    \begin{tabular}{lc|rrr|rrr}
        \toprule
        \multicolumn{2}{c}{} & \multicolumn{3}{c}{identity} & \multicolumn{3}{c}{domain-escape} \\
        \multicolumn{2}{c}{} & \multicolumn{3}{c}{\textit{(without the concretizer)}} & \multicolumn{3}{c}{} \\

        \cmidrule[\heavyrulewidth](lr){3-5} \cmidrule[\heavyrulewidth](lr){6-8}

        \textbf{Domain} & $\%{\cap}\mathrm{S}$ & $t(s)$ & $\#\pi$ & $|\pi_*|$ & $t(s)$ & $\#\pi$ & $|\pi_*|$ \\

        \midrule
        acrobatics & 0.50 & 0.32 & 39,409 & 14.00 & 0.04 & 197 & 14.00 \\
        beam-walk & 0.82 & 0.38 & 454 & 453.22 & 0.22 & 454 & 453.22 \\
        blocksworld-original & 0.47 & 82.59 & 408,803 & 12.79 & 81.46 & 226,549 & 12.79 \\
        blocksworld-advanced & 0.20 & 0.41 & 8,444 & 9.64 & 0.39 & 4,552 & 9.64 \\
        chain-of-rooms & 0.30 & 13.75 & 1,193,474 & 57.00 & 1.83 & 118,996 & 57.00 \\
        earth-observation & 0.20 & 7.58 & 756,538 & 13.88 & 7.30 & 693,463 & 13.88 \\
        elevators & 0.47 & 0.33 & 42,196 & 14.86 & 0.39 & 42,196 & 14.86 \\
        faults & 0.40 & 3.16 & 329,743 & 15.36 & 2.28 & 238,657 & 15.36 \\
        first-responders & 0.43 & 2.40 & 336,141 & 7.56 & 2.56 & 194,506 & 7.56 \\
        tireworld-triangle & \textbf{1} & 27.46 & 485 & 244.00 & 26.26 & 485 & 244.00 \\
        zenotravel & 0.33 & 2.33 & 1,808 & 14.80 & 2.33 & 1,808 & 14.80 \\
        \midrule
        doors & 0.87 & 8.00 & 5,048 & 5,038.62 & 9.05 & 5,048 & 5,038.62 \\
        islands & 0.38 & 1.38 & 271,729 & 4.65 & 1.50 & 271,515 & 4.65 \\
        miner & 0.10 & 211.36 & 2,271,451 & 14.20 & 212.47 & 2,209,231 & 14.20 \\
        tireworld-spiky & 0.45 & 45.99 & 1,268,072 & 25.00 & 49.41 & 1,268,072 & 25.00 \\
        tireworld-truck & 0.16 & 1.18 & 282,463 & 13.00 & 1.51 & 268,504 & 13.00 \\
        \midrule
        TOTAL & 0.442 & 4.18 & 68,857 & 29.94 & 3.20 & 37,209 & 29.94 \\
        \bottomrule
    \end{tabular}
    \end{adjustbox}
    \caption{Average runtime, number of generated policies, and size of returned solutions, per configuration, over the intersection of solved tasks.}
\end{subtable}
\caption{Empirical results for AND$^*$ with and without domain-escape equivalence pruning.}
\label{tab:results-domain-escape}
\end{table}

\begin{figure}[tb]
    \centering
    \includegraphics[width=.46\textwidth]{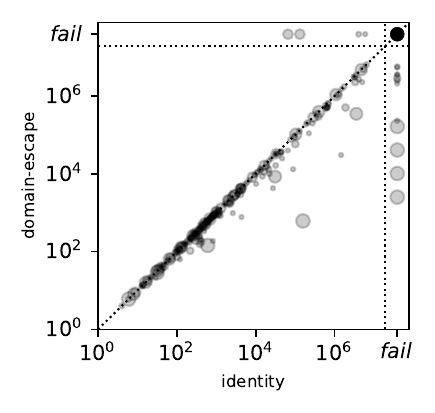}
    \caption{\revb{Per-task comparison on the number of generated policies between using domain-escape equivalence pruning or not.}}
    \label{fig:scatter-domain-escape}
\end{figure}

We empirically compare AND$^*$ using identity and domain-escape equivalence pruning (the latter with the concretizer). Table~\ref{tab:results-domain-escape} shows the results through two subtables.
The first subtable presents the ratio of tasks that were solved, i.e., the \emph{coverage} ($\%\mathrm{S}$), the ratio of tasks that failed by the time limit ($\%\mathrm{T}$), and the ratio of tasks that failed by the memory limit ($\%\mathrm{M}$), for each domain and each approach.
The second subtable presents the average execution time in seconds ($t(s)$), the average number of generated policies ($\#\pi$), and the average returned solution size ($|\pi_*|$) for each domain and each approach. The second subtable considers only the tasks solved by AND$^*$ using both identity and domain-escape equivalence pruning. The $\%{\cap}\mathrm{S}$ column shows for each domain the ratio of the tasks solved using both approaches.
In both subtables, the first block of domains is the domains from the IPC-FOND benchmark, while the second block of domains is the domains from the NEW-FOND benchmark. \rev{The final row aggregates the results from all domains.
All domains have the same weight in the aggregation, independently of the number of (solved) tasks each has.
We use arithmetic means to aggregate ratios and geometric means to aggregate other kinds of information to avoid giving more weight to hard domains.}

The number of generated policies has decreased in several domains. Major decreases have occurred in five domains: \emph{acrobatics} (99.5\% reduction), \emph{blocksworld-original} (44.6\% reduction), \emph{blocksworld-advanced} (46.1\%), \emph{chain-of-rooms} (90.0\%), \emph{faults} (27.6\%) and \emph{first-responders} (42.1\%) -- all from the IPC-FOND benchmark.

Figure~\ref{fig:scatter-domain-escape} depicts the per-task differences in the number of generated policies between identity and domain-escape pruning. It has one mark per task, with proportionally bigger marks for tasks from domains with fewer tasks.

The reduction was often not enough to increase coverage. We verify that the average ratio of solved tasks (coverage) is similar for both approaches, with small changes in most domains. A major change appears only in one domain: \emph{acrobatics} (from 0.50 to 1).

From now on, we assume that AND$^*$ always follows Algorithm~\ref{alg:extract-solution-unhollower}.

\subsection{Escape Equivalence Pruning}
\label{ssec:psym-o}

One can ask if the problem we previously showed regarding the use of $\sign = \front$ equivalence pruning still holds if we use the concretizer. This is a very reasonable question since domain-escape equivalence pruning was not complete before and became complete with the use of the concretizer. Unfortunately, escape equivalence pruning is still sound but not complete, even if we use the concretizer. In this subsection, we show how one can cope with that, and effectively use escape equivalence pruning.

Figure~\ref{fig:state-space-3} shows an example state space with five states: $s_A$ -- the initial state --, $s_B$, $s_C$, $s_D$, and $s_E$ -- the single goal state. We use it to show the incompleteness of escape equivalence pruning. The choices made by $\successors$ method regarding which unmapped state to map when expanding a policy determine the specific policy space AND$^*$ is searching on. We analyze a possible policy space for this state space. The initial search node is $\pi_\initialCondition = \emptyset$ and it has two successors: $\pi_1 = \{s_A \mapsto a_L\}$ and $\pi_2 = \{s_A \mapsto a_R\}$. We note that $\front(\pi_1) = \{s_B, s_D\}$ and $\front(\pi_2) = \{s_C, s_D\}$. Let's say AND$^*$ chooses to map $s_B$ instead of $s_D$ for $\pi_1$ and $s_C$ instead of $s_D$ for $\pi_2$. Then, the single successor of $\pi_1$ is $\pi_3 = \{s_A \mapsto a_L, s_B \mapsto b\}$ and the single successor of $\pi_2$ is $\pi_4 = \{s_A \mapsto a_R, s_C \mapsto c\}$. We note that $\front(\pi_3) = \{s_C, s_D\}$ and $\front(\pi_4) = \{s_B, s_D\}$. Both $\pi_3$ and $\pi_4$ are two steps away from becoming a solution policy. The critical problem is that both $\pi_3$ and $\pi_4$ will be pruned because $\pi_1 \sim \pi_4$ and $\pi_2 \sim \pi_3$ if AND$^*$ chooses to expand~$\pi_1$~and~$\pi_2$ before expanding either $\pi_3$ or $\pi_4$. This kind of ``crossed-mutual'' pruning, where the descendant of some policy is pruned by the ancestor of another policy and vice-versa, cannot happen when using previous equivalence pruning concepts because $\pi \sim \pi'$ implies/requires $|\domain(\pi)| = |\domain(\pi')|$ when $\sign$ is identity or domain-escape.\footnote{We note that even if we prevent ``crossed-mutual'' prunings from occurring, there are more complex examples where escape equivalence pruning is still incomplete.}

\begin{figure}[bt]
    \centering
    \begin{minipage}{0.45\textwidth}
        \centering
        \begin{tikzpicture}
            \node[black_node]            at ( 0.0,  0.0) (a) {$s_A$};
            \node[black_node]            at (-1.0, -2.0) (b) {$s_B$};
            \node[black_node]            at (+1.0, -2.0) (c) {$s_C$};
            \node[black_node]            at ( 0.0, -4.0) (d) {$s_D$};
            \node[black_node, accepting] at ( 0.0, -6.0) (e) {$s_E$};

            \draw[->, thick] ( 0.0, 0.9) -- (a) ;

            \draw[->, thick] (a.225) -- (b) ;
            \draw[->, thick] (a.315) -- (c) ;
            \draw[->, thick] (b.315) -- (c) ;
            \draw[->, thick] (b.315) -- (d) ;
            \draw[->, thick] (c.225) -- (b) ;
            \draw[->, thick] (c.225) -- (d) ;
            \draw[->, thick] (d) -- (e) ;

            \draw[->, thick] (a.225) .. controls (-2, -1.5) and (-2.5, -2.5) .. (d.160) ;
            \draw[->, thick] (a.315) .. controls (+2, -1.5) and (+2.5, -2.5) .. (d.20) ;

            \draw[thick] (-0.5, -0.8) .. controls (-0.62, -0.72) .. (-0.7, -0.6) ;
            \draw[thick] (+0.5, -0.8) .. controls (+0.62, -0.72) .. (+0.7, -0.6) ;
            \draw[thick] (-0.51, -2.7) .. controls (-0.38, -2.55) .. (-0.3, -2.25) ;
            \draw[thick] (+0.51, -2.7) .. controls (+0.38, -2.55) .. (+0.3, -2.25) ;

            \node[] at (-0.91, -0.39) () {$a_L$};
            \node[] at (0.9, -0.4) () {$a_R$};
            \node[] at (-0.2, -2.55) () {$b$};
            \node[] at (0.2, -2.6) () {$c$};
            \node[] at (-0.2, -4.88) () {$d$};

        \end{tikzpicture}
    \end{minipage}
    \caption{\rev{Example state space to show the incompleteness of AND$^*$ with escape equivalence pruning, even with the concretizer.}}
    \label{fig:state-space-3}
\end{figure}

Despite being incomplete, AND$^*$ always terminates and never returns false-positive answers when using escape equivalence pruning (Lemma~\ref{lma:termination-and-soundness}). So, AND$^*$ always returns either $\bot$ or a solution. Never a non-solution policy~$\pi$. That is the case because, by construction, it has no means of returning a policy that is not a solution, regardless of the choice of $\sign$.
We can use this fact to make AND$^*$ with escape equivalence pruning sound and complete in a straightforward way.
\rev{If we run first an incomplete but sound algorithm with termination guarantees, and then -- only if it fails -- run a sound and complete algorithm, we create a hybrid algorithm that has the desired guarantees. This approach was taken by \cite{hoffmann2001ff} to create the FF planning system.
We follow the same approach. We set AND$^*$ with escape equivalence pruning as the first phase algorithm, and with domain-escape equivalence pruning as the second phase algorithm. \emph{Optimality~is~lost~here.}}

\rev{Later in this work, we present other approaches that improve on AND$^*$ with escape equivalence pruning. These other approaches are also incomplete, so we always use them in the same way, making them complete by setting up a fallback to AND$^*$ with domain-escape equivalence pruning. Nevertheless, this second phase was never needed in any of our experiments. The hybrid algorithms either failed due to resource limits during the first phase or found a solution through the first phase.}

\begin{figure}[tb]
    \centering
    \includegraphics[width=.46\textwidth]{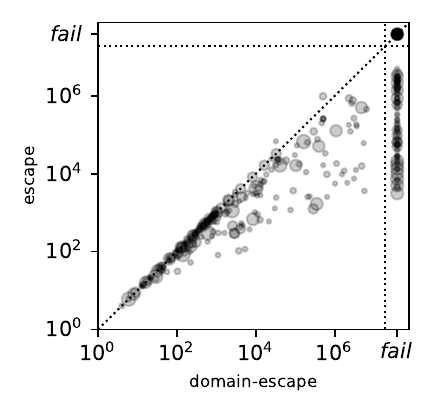}
    \caption{\rev{Per-task comparison on the number of generated policies between using domain-escape and escape equivalence pruning.}}
    \label{fig:scatter-out}
\end{figure}

\begin{table}[hp]
\begin{subtable}[t]{\textwidth}
    \centering
    \setlength\tabcolsep{5pt}
    \fontsize{9}{10}\selectfont
    \begin{adjustbox}{center}
    \begin{tabular}{l|ccc|ccc}
        \toprule
        \multicolumn{1}{c}{} & \multicolumn{3}{c}{domain-escape} & \multicolumn{3}{c}{escape} \\

        \cmidrule[\heavyrulewidth](lr){2-4} \cmidrule[\heavyrulewidth](lr){5-7}

        \textbf{Domain} & $\%\mathrm{S}$ & $\%\mathrm{T}$ & $\%\mathrm{M}$ & $\%\mathrm{S}$ & $\%\mathrm{T}$ & $\%\mathrm{M}$ \\

        \midrule
        acrobatics & \textbf{1} & \textcolor{gray}{0} & \textcolor{gray}{0} & \textbf{1} & \textcolor{gray}{0} & \textcolor{gray}{0} \\
        beam-walk & 0.82 & 0.18 & \textcolor{gray}{0} & 0.82 & 0.18 & \textcolor{gray}{0} \\
        blocksworld-original & 0.50 & 0.33 & 0.17 & 0.53 & 0.47 & \textcolor{gray}{0} \\
        blocksworld-advanced & 0.20 & 0.64 & 0.16 & 0.20 & 0.71 & 0.09 \\
        chain-of-rooms & 0.30 & \textcolor{gray}{0} & 0.70 & \textbf{1} & \textcolor{gray}{0} & \textcolor{gray}{0} \\
        earth-observation & 0.20 & \textcolor{gray}{0} & 0.80 & 0.33 & 0.15 & 0.53 \\
        elevators & 0.47 & \textcolor{gray}{0} & 0.53 & 0.80 & \textcolor{gray}{0} & 0.20 \\
        faults & 0.40 & \textcolor{gray}{0} & 0.60 & 0.56 & \textcolor{gray}{0} & 0.44 \\
        first-responders & 0.48 & \textcolor{gray}{0} & 0.52 & 0.69 & 0.25 & 0.05 \\
        tireworld-triangle & \textbf{1} & \textcolor{gray}{0} & \textcolor{gray}{0} & \textbf{1} & \textcolor{gray}{0} & \textcolor{gray}{0} \\
        zenotravel & 0.33 & 0.67 & \textcolor{gray}{0} & 0.33 & 0.67 & \textcolor{gray}{0} \\
        \midrule
        doors & 0.87 & \textcolor{gray}{0} & 0.13 & \textbf{1} & \textcolor{gray}{0} & \textcolor{gray}{0} \\
        islands & 0.42 & \textcolor{gray}{0} & 0.58 & 0.63 & \textcolor{gray}{0} & 0.37 \\
        miner & 0.10 & 0.02 & 0.88 & 0.24 & 0.76 & \textcolor{gray}{0} \\
        tireworld-spiky & 0.45 & \textcolor{gray}{0} & 0.55 & 0.82 & 0.18 & \textcolor{gray}{0} \\
        tireworld-truck & 0.18 & \textcolor{gray}{0} & 0.82 & 0.46 & \textcolor{gray}{0} & 0.54 \\
        \midrule
        TOTAL & 0.482 & 0.115 & 0.403 & 0.651 & 0.211 & 0.138 \\
        \bottomrule
    \end{tabular}
    \end{adjustbox}
    \caption{\rev{Ratio of successful, timeout, and memory-out runs per configuration.}}
\end{subtable}

\bigskip

\begin{subtable}[t]{\textwidth}
    \centering
    \setlength\tabcolsep{5pt}
    \fontsize{9}{10}\selectfont
    \begin{adjustbox}{center}
    \begin{tabular}{lc|rrr|rrr}
        \toprule
        \multicolumn{2}{c}{} & \multicolumn{3}{c}{domain-escape} & \multicolumn{3}{c}{escape} \\

        \cmidrule[\heavyrulewidth](lr){3-5} \cmidrule[\heavyrulewidth](lr){6-8}

        \textbf{Domain} & $\%{\cap}\mathrm{S}$ & $t(s)$ & $\#\pi$ & $|\pi_*|$ & $t(s)$ & $\#\pi$ & $|\pi_*|$ \\

        \midrule
        acrobatics & \textbf{1} & 3.10 & 27,206 & 126.50 & 2.08 & 11,109 & 126.50 \\
        beam-walk & 0.82 & 0.22 & 454 & 453.22 & 0.21 & 454 & 453.22 \\
        blocksworld-original & 0.50 & 121.56 & 402,523 & 13.40 & 152.25 & 182,954 & 13.47 \\
        blocksworld-advanced & 0.20 & 0.39 & 4,552 & 9.64 & 0.37 & 1,729 & 9.64 \\
        chain-of-rooms & 0.30 & 1.83 & 118,996 & 57.00 & 0.40 & 828 & 57.00 \\
        earth-observation & 0.20 & 7.30 & 693,463 & 13.88 & 0.08 & 3,725 & 14.00 \\
        elevators & 0.47 & 0.39 & 42,196 & 14.86 & 0.07 & 383 & 14.86 \\
        faults & 0.40 & 2.28 & 238,657 & 15.36 & 0.73 & 79,559 & 16.27 \\
        first-responders & 0.48 & 4.96 & 487,417 & 8.03 & 8.47 & 18,001 & 8.08 \\
        tireworld-triangle & \textbf{1} & 26.26 & 485 & 244.00 & 25.72 & 485 & 244.00 \\
        zenotravel & 0.33 & 2.33 & 1,808 & 14.80 & 2.33 & 934 & 14.80 \\
        \midrule
        doors & 0.87 & 9.05 & 5,048 & 5,038.62 & 7.86 & 5,048 & 5,038.62 \\
        islands & 0.42 & 2.96 & 575,809 & 4.84 & 0.67 & 17,057 & 4.84 \\
        miner & 0.10 & 212.47 & 2,209,231 & 14.20 & 214.63 & 166,141 & 14.20 \\
        tireworld-spiky & 0.45 & 49.41 & 1,268,072 & 25.00 & 37.97 & 138,650 & 25.00 \\
        tireworld-truck & 0.18 & 3.20 & 511,915 & 13.15 & 0.15 & 2,568 & 13.15 \\
        \midrule
        TOTAL & 0.482 & 4.89 & 60,651 & 34.70 & 2.11 & 6,305 & 34.87 \\
        \bottomrule
    \end{tabular}
    \end{adjustbox}
    \caption{\rev{Average runtime, number of generated policies, and size of returned solutions, per configuration, over the intersection of solved tasks.}}
\end{subtable}
\caption{\rev{Empirical results for AND$^*$ using escape and domain-escape equivalence pruning.}}
\label{tab:results-out}
\end{table}

We empirically compare AND$^*$ using escape equivalence pruning againt AND$^*$ using just domain-escape equivalence pruning (Table~\ref{tab:results-out}). We can check that there was an expressive increase in coverage in comparison to domain-escape equivalence pruning (from 0.482 to 0.651). Major gains have occurred in \emph{chain-of-rooms} (from 0.30 to 1), \emph{elevators} (from 0.47 to 0.80), \emph{first-responders} (from 0.48 to 0.69), \emph{islands} (from 0.42 to 0.63), \emph{tireworld-spiky} (from 0.45 to 0.82), and \emph{tireworld-truck} (from 0.18 to 0.46). There is a dramatic improvement in the number of generated policies in several domains. Decreases of at least 50\% have occurred in \emph{acrobatics}, \emph{blocksworld-original}, \emph{blocksworld-advanced}, \emph{chain-of-rooms}, \emph{earth-observation}, \emph{elevators}, \emph{faults}, \emph{first-responders}, \emph{zenotravel}, \emph{islands}, \emph{miner}, \emph{tireworld-spiky}, and \emph{tire-world-truck} -- all but three domains. Decreases of at least one order of magnitude have occurred in seven domains: \emph{chain-of-rooms}, \emph{earth-observation}, \emph{elevators}, \emph{first-responders}, \emph{islands}, \emph{miner}, and \emph{tireworld-truck}. Figure~\ref{fig:scatter-out} depicts, per task, the differences.

\section{Improving Equivalence Pruning with State Symmetries}
\label{sec:ssym}

\color{revcolor}
In this section, we focus on applying the knowledge of \emph{states} that are equivalent to each other, to broaden the definition of policies that are equivalent to each other, consequently strengthening the policy equivalence pruning.

Two states are considered symmetrically equivalent to each other if they participate on ``symmetrical" trajectories in the state space. For instance, in our example from Figure~\ref{fig:state-space-3}, $s_B$ and $s_C$ are symmetrically equivalent (or simply symmetric) to each other. We can use this knowledge to define a new signature function for policies.

Let~$\signStates$ be a function with $\domain(\signStates) = \states$ that maps symmetric states to the same signature. Each signature defines an equivalence class of states. Using this~$\signStates$ function, we can define a new escape equivalence definition as $\sign(\pi) = \{\!\!\{\signStates(s) : s \in \escape(\pi)\}\!\!\}$\footnote{The double curly brackets notation denotes a multiset: i.e., a collection of elements, that unlike a set, allows repeated elements, but like a set, does not have an order.}.
With this new definition, two policies are considered equivalent to each other iff they contain in their $\escape$ the same number of states from each equivalence class of symmetric states.
However, two questions arise. How can we determine that two states are symmetric to each other? And what happens when we use this new definition of equivalence pruning with AND$^*$?

Structural state symmetries \citep{pochter2011exploiting,shleyfman2015heuristics} is a well-defined concept of state symmetries that uses the structure of the planning task to determine which states are symmetric to each other.
Structural state symmetries capture part of all existing state symmetries, and \cite{winterer2016structural} shows we can apply the concept of structural state symmetries for FOND planning.
In this work, we use structural state symmetries to define the $\signStates$ function.

Regarding the second question, we evaluate the empirical impact of using the new definition of equivalence pruning with AND$^*$, using two different techniques to find structural state symmetries: an approximated (greedy) approach and the perfect approach.
The greedy approach comes from the literature.
\cite{pochter2011exploiting} and \cite{winterer2016structural} compute structural state symmetries using a greedy hill-climbing procedure, respectively in the context of classical and FOND planning.
The justification given for using an approximated approach is that computing all structural state symmetries (the perfect approach) is \emph{intractable}~\citep{winterer2016structural}.

We show that we can \emph{in practice} efficiently perfectly compute structural state symmetries for FOND planning tasks using group theory techniques from \cite{jefferson2019minimal}, when using a slightly more restrictive definition of structural state symmetries.
\color{black}

\subsection{Structural State Symmetries}

Let $\Pi = \langle \facts, \initialCondition, \goalCondition, \actions \rangle$ be a FOND planning task. A structural symmetry for~$\Pi$ is\footnote{We combine the definitions from \cite{shleyfman2015heuristics} and \cite{winterer2016structural}. The former defines structural symmetries for classical planning tasks in~STRIPS \citep{fikes1971strips}, while the latter defines them for FOND planning tasks in SAS+ \citep{backstrom1995complexity}. Since our work's running definition of FOND planning tasks uses~the STRIPS formalism, our formal definition of structural symmetries is an adaptation of~theirs. Finally, since we use SAS+ in our implementation, we also talk about it.} a permutation $\sigma : \facts \sqcup \actions \to \facts \sqcup \actions$, such that:

\begin{enumerate}
    \item $\sigma = \sigma_\facts \sqcup \sigma_\actions$, where $\sigma_\facts : \facts \to \facts$ is a permutation of facts and $\sigma_\actions : \actions \to \actions$ is a permutation of actions;
    \item $\sigma(f) \in \goalCondition \Leftrightarrow f \in \goalCondition$; and
    \item $\{\sigma(f) : f \in \pre(a)\} = \pre(\sigma(a))$ and $\{\{\sigma(f) : f \in e\} : e \in \effs(a)\} = \{e : e \in \effs(\sigma(a))\}$, for each $a \in \actions$.
\end{enumerate}

We overload $\sigma$, so that, for each state $s \in \states$, $\sigma(s)$ denotes the state $s' \in \states$ with $\facts[s'] = \{\sigma(f) : f \in \facts[s]\}$. We note that if $\sigma$ is a structural symmetry, $\sigma(s)$ is a goal state if and only if $s$ is a goal state. Moreover, an action $\sigma(a)$ is applicable to $\sigma(s)$ if and only if the action $a$ is applicable to $s$. Finally, the set $\{s' : s' \in \succs(\sigma(s), \sigma(a))\}$ is equal to the set $\{\sigma(s') : s' \in \succs(s, a)\}$.

The definition of structural symmetries we use in this work is more restrictive. It has two extra conditions:
\begin{enumerate}
    \setcounter{enumi}{3}
    \item $P_\facts(\sigma(f)) = P_\facts(f)$, for each $f \in \facts$, where $P_\facts : \facts \to \mathbb{Z}$ is a \emph{given} partition of the facts; and
    \item $P_\actions(\sigma(a)) = P_\actions(f)$, for each $a \in \actions$, where $P_\actions : \actions \to \mathbb{Z}$ is a \emph{given} partition of the actions.
\end{enumerate}

The partition $P_\facts$ comes from the fact we use a SAS+~\citep{backstrom1995complexity} formalism to represent the FOND planning tasks in our implementation. When using SAS+, facts are partitioned into variables such that, for each variable, exactly one fact of that variable is true at each state. This allows a more compact representation of actions and states~\citep{helmert2009concise}. With the first extra restriction, our definition of structural symmetries matches the definition from~\cite{winterer2016structural}.

The second extra restriction is an optimization. We look only for structural symmetries that permute actions that were grounded from the same lifted action. This potentially loses symmetries between distinct lifted actions but allows a more focused computation of structural symmetries.

\color{revcolor}
\subsection{Computing Structural Symmetries}

Let $Aut(\Pi)$ be the \emph{permutation group} containing all structural symmetries for $\Pi$. The ideal objective would be to have a \emph{canonical} $\signStates$ function: i.e., a $\signStates$ function such that $\signStates(s) = \signStates(s')$ iff there exists a structural symmetry~$\sigma \in Aut(\Pi)$ with $s' = \sigma(s)$\footnote{Note that if $\sigma$ is a structural symmetry for a task~$\Pi$, then $\sigma^{-1}$ also is. So if exists a structural symmetry $\sigma \in Aut(\Pi)$ with $s' = \sigma(s)$, then exists a structural symmetry $\sigma' = \sigma^{-1} \in Aut(\Pi)$ with $s = \sigma'(s')$.}.

Let $Orb(s) = \{\sigma(s) : \sigma \in Aut(\Pi)\}$ be the set of states symmetric to a given state $s$.
We could set $\signStates(s) = Orb(s)$ for each state $s$, and obtain a canonical $\signStates$ function.
Moreover, if we want the $\signStates$ function to have smaller signature objects, and map states to states instead of mapping states to sets of states, we could choose an arbitrary relation of total order~$\prec$ over states, and then we could instead set $\signStates(s) = \min Orb(s)$ w.r.t.\! $\prec$.
The issue is that $Aut(\Pi)$ is usually too large even to be enumerated. In the worst case, the asymptotic size of $Aut(\Pi)$ is factorial to the size of facts and actions.
Therefore, this direct approach cannot be used in practice.
Nevertheless, the idea of finding the minimal symmetric state is still used in the techniques we present next.

\subsubsection{Approximating a Canonical Function}
\color{black}

\cite{pochter2011exploiting} and \cite{shleyfman2015heuristics} show we can compute a \emph{generator} for $Aut(\Pi)$ using a \emph{Problem Description Graph} (PDG) of the task. A generator of a permutation group~$\mathcal{G}$ is a set of permutations~$g$, such that for every permutation $\sigma \in \mathcal{G}$, there exists some sequence $\langle \sigma_1, \sigma_2, \dots, \sigma_n \rangle$~of permutations in $g$, possibly with repetitions, such that the result of their \rev{composition} is~$\sigma$. In other words, a generator of a permutation~group~$\mathcal{G}$ is a compact representation of~$\mathcal{G}$.

\cite{winterer2016structural} \emph{approximate} the desired canonical function using a greedy hill-climbing procedure \rev{w.r.t.\! $\prec$} on the local state neighborhood~$\mathcal{N}$ provided by the direct application of generator permutations -- i.e. $\mathcal{N}(s) = \{\sigma(s) : \sigma \in g\}$, where $g$ is the computed generator of $Aut(\Pi)$. Following their approach, we could define $\signStates(s)$ to be the \rev{final state} resulting from the hill-climbing procedure starting from $s$. If $s \sim s'$ for the $\signStates$ defined from this greedy hill-climbing procedure, then $s \sim s'$ also holds for a canonical $\signStates$, but the opposite is not always true.

\color{revcolor}
We note that after we compute the generator $g$ of $Aut(\Pi)$, we could, for the purposes of this work, ignore the actions part of each $\sigma \in g$.
This is because the symmetries between actions are not necessary for the computation of $\signStates$. Knowledge about the actions is only necessary at the first step, to ensure that $Aut(\Pi)$ contains only correct symmetry relations for the task, which requires associating the facts with the actions.
In other words, once we have already computed the generator $g$ of $Aut(\Pi)$, we can create a new generator $g' = \{\sigma_\facts : \sigma \in g\}$, which induces a new $Aut(\Pi)$, simplified, that acts only on the facts, but retains the same symmetries between states.
We make this remark because the definitions we use for presenting the next technique mostly assume that the permutation group acts only on the~facts.

\subsubsection{Perfectly Computing a Canonical Function}

In this work, we apply a technique presented by \cite{jefferson2019minimal} that allows the computation of the canonical signature of a state by using a (compactly represented) \emph{stabilizer chain} of $Aut(\Pi)$.

A fact~$f$ is said to be \emph{point-wise stable} in a permutation group~$\mathcal{G} \leq Aut(\Pi)$, where $\leq$ denotes that $\mathcal{G}$ is a \emph{subgroup} of $Aut(\Pi)$, iff, for every structural symmetry $\sigma \in \mathcal{G}$, $\sigma(f) = f$.

A stabilizer chain of $Aut(\Pi)$ is a sequence $\langle \mathcal{G}_0, \mathcal{G}_1, \dots, \mathcal{G}_n \rangle$ of subgroups of $Aut(\Pi)$, such that:

\begin{enumerate}
    \item $\mathcal{G}_0 = Aut(\Pi)$;
    \item $\mathcal{G}_n$ is a group that contains only the identity structural symmetry (i.e., the structural symmetry $\sigma$ that maps every fact to itself and every action to itself); and
    \item for each $i \in \{1, 2, \dots, n\}$:
    \begin{enumerate}
        \item $\mathcal{G}_i \leq \mathcal{G}_{i-1}$;
        \item the set of facts that are point-wise stable in $\mathcal{G}_i$ is a strict superset of the set of facts that are point-wise stable in $\mathcal{G}_{i-1}$; and
        \item there is no subgroup $\mathcal{G}'$ of $Aut(\Pi)$ such that $\mathcal{G}_i \leq \mathcal{G}' \leq \mathcal{G}_{i-1}$ that still stabilizes more facts than $\mathcal{G}_{i-1}$.
    \end{enumerate}
\end{enumerate}

For every possible stabilizer chain of $Aut(\Pi)$, there is a total ordering of facts that compactly represents it unambiguously. Conversely, every total ordering of facts compactly represents some stabilizer chain of $Aut(\Pi)$. These total orderings of facts are called \emph{bases}. Let $\mathcal{B} = \langle f_1, f_2, \dots, f_{|\facts|} \rangle$ be a base. The stabilizer chain that $\mathcal{B}$ represents is the $\langle \mathcal{G}_0, \mathcal{G}_1, \dots, \mathcal{G}_{n} \rangle$ stabilizer chain where, for each $i \in \{1, 2, \dots, n\}$, $\mathcal{G}_i$ is the maximal subgroup of $\mathcal{G}_{i-1}$ that stabilizes the first fact of $\mathcal{B}$ that is not stabilized by~$\mathcal{G}_{i-1}$.

Let $\mathcal{G}$ and $\mathcal{G}'$ be two permutation groups that appear consecutively in a stabilizer chain of $Aut(\Pi)$ and let $f$ be the first fact in the base of the stabilizer chain that is stabilized by $\mathcal{G}'$ but not by $\mathcal{G}$. The \emph{cosets} of $\mathcal{G}'$ in $\mathcal{G}$ are the result of the partition of the permutation group in $\mathcal{G}$ into the unique \emph{minimal set} of permutation groups $\{\mathcal{G}_1, \mathcal{G}_2, \dots, \mathcal{G}_m\}$, such that, for each coset $\mathcal{G}_i$, every pair of permutations $\sigma_1, \sigma_2 \in \mathcal{G}_i$ satisfies $\sigma_1(f) = \sigma_2(f)$. A \emph{transversal} is a set of permutations $\{\sigma_1, \sigma_2, \dots, \sigma_m\}$, with exactly one permutation from each coset. An \emph{inverse transversal}, on the other hand, is a set of permutations $\{\sigma_1, \sigma_2, \dots, \sigma_m\}$, such that $\{\sigma_1^{-1}, \sigma_2^{-1}, \dots, \sigma_m^{-1}\}$ is a transversal.

\cite{jefferson2019minimal} show that if we have a generator $g$ of a permutation group $\mathcal{G}$, a base $\mathcal{B}$ that represents a stabilizer chain $\langle \mathcal{G}_0, \mathcal{G}_1, \dots, \mathcal{G}_n \rangle$ of~$\mathcal{G}$, a set of elements $\langle e_1, e_2, \dots, e_n \rangle$ such that each $e_i$ is the first element of $\mathcal{B}$ that~$\mathcal{G}_i$ stabilizes but $\mathcal{G}_{i-1}$ does not, and a set of inverse transversals $\langle \mathcal{I}_1, \mathcal{I}_2, \dots, \mathcal{I}_n \rangle$ such that each $\mathcal{I}_i$ is the inverse transversal of $\mathcal{G}_i$ in $\mathcal{G}_{i-1}$, then we can use that to compute a \emph{canonical image}\footnote{A canonical image of a subset of elements is a representative chosen so that two subsets have the same canonical image if and only if one can be transformed into the other by permutations from the group.} of any subset of elements w.r.t.~$\mathcal{G}$. Applied to our context, if $\mathcal{G} = Aut(\Pi)$, we can compute the canonical signature of any given state, by looking at the canonical image of the set of facts that are true in that state.

\begin{algorithm}[p]\color{revcolor}
    \DontPrintSemicolon
    \SetKwIF{If}{ElseIf}{Else}{if}{:}{else if}{else:}{endif}
    \SetKwFor{While}{while}{:}{endw}
    \SetKwFor{ForEach}{for each}{:}{endfch}

    \revb{\textbf{Task Constants:} $g$, $\mathcal{B}$, $\langle f_1, f_2, \dots, f_n \rangle$, $\langle \mathcal{I}_1, \mathcal{I}_2, \dots, \mathcal{I}_n \rangle$} \\
    \textbf{Input:} $s$ \tcp{a state}

    $\mathrm{SymmetricStates} \coloneqq \{s\}$ \\
    $M \coloneqq \emptyset$ \\

    \ForEach{\upshape $f \in \mathcal{B}$ \textbf{while} $|M| < |\facts[s]|$}
    {
        \If{\upshape $\exists f_i \in \langle f_1, f_2, \dots, f_n \rangle$ \textbf{such that} $f_i = f$}
        {
            $\mathrm{NewSymmetricStates} \coloneqq \emptyset$ \\
            \ForEach {$\sigma \in \mathcal{I}_i$}
            {
                \ForEach {$s' \in \mathrm{SymmetricStates}$}
                {
                    $X \coloneqq \{\sigma(f') : f' \in \facts[s']\}$ \\
                    Let $s''$ be the state with $\facts[s''] = X$. \\
                    $\mathrm{NewSymmetricStates} \coloneqq \mathrm{NewSymmetricStates} \cup \{s''\}$
                }
            }
            $\mathrm{SymmetricStates} \coloneqq \mathrm{NewSymmetricStates}$
        }
        $\mathrm{SSWithF} \coloneqq \{s' : s' \in \mathrm{SymmetricStates} \land f \in \facts[s']\}$ \\
        $\mathrm{SSWithoutF} \coloneqq \{s' : s' \in \mathrm{SymmetricStates} \land f \not\in \facts[s']\}$ \\
        \If{\upshape $\mathrm{SSWithF} \neq \emptyset$ \textbf{and} $\mathrm{SSWithoutF} \neq \emptyset$}
        {
            $\mathrm{SymmetricStates} \coloneqq \mathrm{SSWithF}$ \tcp*[l]{\upshape This was an insight from \cite{jefferson2019minimal}. We can this way safely prune part of the symmetry search space, without losing the canonical image of the input, in case we are following the inverse transversals of a stabilizer chain.}
        }
        \If{\upshape $\mathrm{SSWithF} \neq \emptyset$}
        {
            $M \coloneqq M \cup \{f\}$
        }
    }

    Let $s'$ be the state with $\facts[s'] = M$. \\
    \textbf{return} $s'$ \tcp*[l]{\cite{jefferson2019minimal} prove that $M$ is the minimal symmetric subset (minimal image) of the input subset (in this case, $\facts[s]$) w.r.t the group represented by $g$ and the ordering induced by $\mathcal{B}$, when the algorithm terminates. These minimal images become canonical representatives if $\mathcal{B}$ is fixed in all runs.}

    \caption{Adapatation from one of \cite{jefferson2019minimal}'s algorithms to the context of FOND planning.}
    \label{alg:jefferson}
\end{algorithm}

Algorithm~\ref{alg:jefferson} shows the pseudocode of \cite{jefferson2019minimal}'s technique\footnote{Throughout their work, they present more than one technique to compute the canonical image of a subset of elements w.r.t.\! to a permutation group that acts on those elements. We follow their simplest presented technique (\texttt{FixedOrbit}). The impact of using the other presented techniques, which vary $\mathcal{B}$ throughout different runs, is an interesting question for future work.} applied to our context, where elements are facts, and subsets of elements are states.
As far as we know, this is the first time their work is being applied to planning.

We show next that with their~technique, we are able to compute a canonical signature of states almost as fast as we can compute a greedy signature using the hill-climbing procedure.

\subsection{Empirical Evaluation}
\color{black}

\begin{table}[ht]
    \centering
    \fontsize{9}{10.5}\selectfont
    \begin{tabular}{|c|lc}
        \toprule

        \multicolumn{1}{c}{} &
        \textbf{Domain} &
        $\%$ tasks w/ sym. \\

        \midrule
        \multirow{11}{*}{\rotatebox[origin=c]{90}{IPC-FOND}}
         & acrobatics & none \\
         & beam-walk & none \\
         & chain-of-rooms & none \\
         & tireworld-triangle & none \\
         & earth-observation & just one \\
         & blocksworld-original & $50\%$ \\
         & blocksworld-advanced & $55\%$ \\
         & elevators & $80\%$ \\
         & faults & all but one \\
         & first-responders & all but one \\
         & zenotravel & all \\
        \midrule
        \multirow{5}{*}{\rotatebox[origin=c]{90}{NEW\hspace{-0.3mm}-\hspace{-0.3mm}FOND}}
         & doors & none \\
         & tireworld-truck & $89\%$ \\
         & tireworld-spiky & all but one \\
         & islands & all but one \\
         & miner & all \\
        \bottomrule
    \end{tabular}
    \caption{\rev{Ratio of tasks with non-trivial symmetries for each domain in the benchmark.}}
    \label{tab:benchmarks-symmetries}
\end{table}

Table~\ref{tab:benchmarks-symmetries} shows, for each domain, the ratio of tasks with structural symmetries (besides the identity symmetry). For eight of the 16 domains, at least 80\% of tasks in each domain present structural symmetries. And for ten, at least half the tasks present structural symmetries. We use Nauty~\citep{mckay2014practical} to compute automorphism group generators of PDGs, from which we can extract $Aut(\Pi)$ generators. We are able to compute a generator of $Aut(\Pi)$ in at most 1.75 seconds for all tasks that have symmetries. We set an overcautious time limit of five seconds for this computation. If Nauty fails to conclude its execution within this time limit, the use of structural symmetries is disabled. It only fails for tasks \emph{p7} and \emph{p8} of \emph{acrobatics}, tasks \emph{p6} to \emph{p11} of \emph{beam-walk}, and tasks \emph{p11} to \emph{p40} of \emph{tireworld-triangle} -- all of which likely do not have structural symmetries because, in these domains, the computation does not find symmetries for all tasks where it concludes. \rev{We use GAP~(\emph{Groups, Algorithms, and Programming} software) to compute the inverse transversals for a permutation group, given an arbtirary base and a generator. We also use GAP to order the facts by their \emph{orbit sizes}\footnote{\rev{I.e., the number of different facts they can be mapped to by a symmetry $\sigma \in Aut(\Pi)$. This ordering was one of the suggestions made by~\cite{jefferson2019minimal}.}} in $Aut(\Pi)$, descendingly, to obtain the base.}
For every task where computing $Aut(\Pi)$ within time limits was successful, we are able to compute the stabilizer chain of $Aut(\Pi)$ in at most five extra seconds.

\begin{table}[hp]
\begin{subtable}[t]{\textwidth}
    \centering
    \setlength\tabcolsep{5pt}
    \fontsize{9}{10}\selectfont
    \begin{adjustbox}{center}
    \begin{tabular}{l|ccc|ccc|ccc}
        \toprule
        \multicolumn{1}{c}{} & \multicolumn{3}{c}{none} & \multicolumn{3}{c}{greedy} & \multicolumn{3}{c}{canonical} \\

        \cmidrule[\heavyrulewidth](lr){2-4} \cmidrule[\heavyrulewidth](lr){5-7} \cmidrule[\heavyrulewidth](lr){8-10}

        \textbf{Domain} & $\%\mathrm{S}$ & $\%\mathrm{T}$ & $\%\mathrm{M}$ & $\%\mathrm{S}$ & $\%\mathrm{T}$ & $\%\mathrm{M}$ & $\%\mathrm{S}$ & $\%\mathrm{T}$ & $\%\mathrm{M}$ \\

        \midrule
        acrobatics & \textbf{1} & \textcolor{gray}{0} & \textcolor{gray}{0} & \textbf{1} & \textcolor{gray}{0} & \textcolor{gray}{0} & \textbf{1} & \textcolor{gray}{0} & \textcolor{gray}{0} \\
        beam-walk & 0.82 & 0.18 & \textcolor{gray}{0} & 0.82 & 0.18 & \textcolor{gray}{0} & 0.82 & 0.18 & \textcolor{gray}{0} \\
        blocksworld-original & 0.53 & 0.47 & \textcolor{gray}{0} & 0.53 & 0.47 & \textcolor{gray}{0} & 0.53 & 0.47 & \textcolor{gray}{0} \\
        blocksworld-advanced & 0.20 & 0.71 & 0.09 & 0.20 & 0.71 & 0.09 & 0.20 & 0.71 & 0.09 \\
        chain-of-rooms & \textbf{1} & \textcolor{gray}{0} & \textcolor{gray}{0} & \textbf{1} & \textcolor{gray}{0} & \textcolor{gray}{0} & \textbf{1} & \textcolor{gray}{0} & \textcolor{gray}{0} \\
        earth-observation & 0.33 & 0.15 & 0.53 & 0.33 & 0.15 & 0.53 & 0.33 & 0.15 & 0.53 \\
        elevators & 0.80 & \textcolor{gray}{0} & 0.20 & 0.80 & \textcolor{gray}{0} & 0.20 & 0.80 & \textcolor{gray}{0} & 0.20 \\
        faults & 0.56 & \textcolor{gray}{0} & 0.44 & \textbf{1} & \textcolor{gray}{0} & \textcolor{gray}{0} & \textbf{1} & \textcolor{gray}{0} & \textcolor{gray}{0} \\
        first-responders & 0.69 & 0.25 & 0.05 & 0.73 & 0.25 & 0.01 & 0.75 & 0.24 & 0.01 \\
        tireworld-triangle & \textbf{1} & \textcolor{gray}{0} & \textcolor{gray}{0} & \textbf{1} & \textcolor{gray}{0} & \textcolor{gray}{0} & \textbf{1} & \textcolor{gray}{0} & \textcolor{gray}{0} \\
        zenotravel & 0.33 & 0.67 & \textcolor{gray}{0} & 0.33 & 0.67 & \textcolor{gray}{0} & 0.33 & 0.67 & \textcolor{gray}{0} \\
        \midrule
        doors & \textbf{1} & \textcolor{gray}{0} & \textcolor{gray}{0} & \textbf{1} & \textcolor{gray}{0} & \textcolor{gray}{0} & \textbf{1} & \textcolor{gray}{0} & \textcolor{gray}{0} \\
        islands & 0.63 & \textcolor{gray}{0} & 0.37 & 0.80 & 0.13 & 0.07 & 0.80 & 0.13 & 0.07 \\
        miner & 0.24 & 0.76 & \textcolor{gray}{0} & 0.55 & 0.45 & \textcolor{gray}{0} & 0.53 & 0.47 & \textcolor{gray}{0} \\
        tireworld-spiky & 0.82 & 0.18 & \textcolor{gray}{0} & \textbf{1} & \textcolor{gray}{0} & \textcolor{gray}{0} & \textbf{1} & \textcolor{gray}{0} & \textcolor{gray}{0} \\
        tireworld-truck & 0.46 & \textcolor{gray}{0} & 0.54 & 0.57 & \textcolor{gray}{0} & 0.43 & 0.57 & \textcolor{gray}{0} & 0.43 \\
        \midrule
        TOTAL & 0.651 & 0.211 & 0.138 & 0.729 & 0.188 & 0.083 & 0.728 & 0.189 & 0.083 \\
        \bottomrule
    \end{tabular}
    \end{adjustbox}
    \caption{\rev{Ratio of successful, timeout, and memory-out runs per configuration.}}
\end{subtable}

\bigskip

\begin{subtable}[t]{\textwidth}
    \centering
    \setlength\tabcolsep{5pt}
    \fontsize{9}{10}\selectfont
    \begin{adjustbox}{center}
    \begin{tabular}{lc|rrr|rrr|rrr}
        \toprule
        \multicolumn{2}{c}{} & \multicolumn{3}{c}{none} & \multicolumn{3}{c}{greedy} & \multicolumn{3}{c}{canonical} \\

        \cmidrule[\heavyrulewidth](lr){3-5} \cmidrule[\heavyrulewidth](lr){6-8} \cmidrule[\heavyrulewidth](lr){9-11}

        \textbf{Domain} & $\%{\cap}\mathrm{S}$ & $t(s)$ & $\#\pi$ & $|\pi_*|$ & $t(s)$ & $\#\pi$ & $|\pi_*|$ & $t(s)$ & $\#\pi$ & $|\pi_*|$ \\

        \midrule
        acrobatics & \textbf{1} & 2.08 & 11,109 & 126.50 & 3.36 & 11,109 & 126.50 & 3.36 & 11,109 & 126.50 \\
        beam-walk & 0.82 & 0.21 & 454 & 453.22 & 2.45 & 454 & 453.22 & 2.45 & 454 & 453.22 \\
        blocksworld-original & 0.53 & 229.56 & 270,882 & 14.12 & 229.84 & 270,709 & 14.12 & 229.86 & 270,708 & 14.12 \\
        blocksworld-advanced & 0.20 & 0.37 & 1,729 & 9.64 & 0.37 & 1,730 & 9.64 & 0.49 & 1,730 & 9.64 \\
        chain-of-rooms & \textbf{1} & 33.01 & 7,968 & 162.00 & 33.75 & 7,968 & 162.00 & 33.29 & 7,968 & 162.00 \\
        earth-observation & 0.32 & 2.72 & 219,401 & 26.08 & 2.83 & 219,178 & 26.08 & 2.74 & 219,172 & 26.08 \\
        elevators & 0.80 & 8.87 & 374,049 & 25.33 & 5.49 & 161,881 & 25.33 & 6.56 & 161,881 & 25.33 \\
        faults & 0.56 & 4.80 & 534,692 & 19.13 & 0.06 & 273 & 19.13 & 0.73 & 273 & 19.13 \\
        first-responders & 0.69 & 84.72 & 244,376 & 10.63 & 59.93 & 172,288 & 10.71 & 60.78 & 171,978 & 10.71 \\
        tireworld-triangle & \textbf{1} & 25.72 & 485 & 244.00 & 29.46 & 485 & 244.00 & 30.31 & 485 & 244.00 \\
        zenotravel & 0.33 & 2.33 & 934 & 14.80 & 1.93 & 689 & 14.80 & 2.72 & 689 & 14.80 \\
        \midrule
        doors & \textbf{1} & 118.06 & 17,484 & 17,473.73 & 114.54 & 17,484 & 17,473.73 & 117.27 & 17,484 & 17,473.73 \\
        islands & 0.63 & 22.38 & 373,141 & 5.74 & 3.23 & 21,403 & 5.74 & 8.66 & 19,214 & 5.74 \\
        miner & 0.24 & 623.20 & 978,423 & 15.33 & 138.27 & 153,427 & 15.33 & 147.98 & 153,427 & 15.33 \\
        tireworld-spiky & 0.82 & 380.18 & 918,175 & 25.00 & 1.70 & 3,798 & 25.00 & 2.41 & 3,798 & 25.00 \\
        tireworld-truck & 0.46 & 27.17 & 488,319 & 13.94 & 0.27 & 4,454 & 13.94 & 0.80 & 4,454 & 13.94 \\
        \midrule
        TOTAL & 0.651 & 15.60 & 44,784 & 45.43 & 5.81 & 10,012 & 45.45 & 8.38 & 9,944 & 45.45 \\
        \bottomrule
    \end{tabular}
    \end{adjustbox}
    \caption{\rev{Average runtime, number of generated policies, and size of returned solutions, per configuration, over the intersection of solved tasks.}}
\end{subtable}
\caption{\rev{Empirical results for AND$^*$ using canonical and greedy structural symmetries, and not using them.}}
\label{tab:results-structural-symmetries}
\end{table}

\begin{figure}[tb]
    \centering
    \includegraphics[width=.44\textwidth]{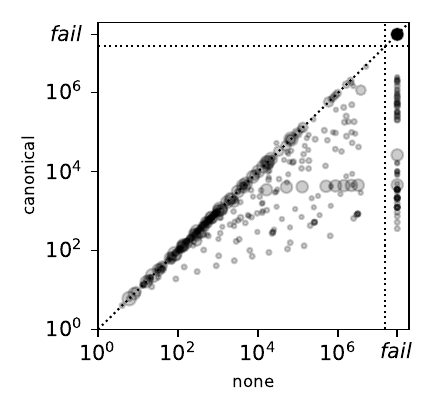}
    \caption{\rev{Per-task comparison on the number of generated policies between using canonical structural state symmetries and not using state symmetries.}}
    \label{fig:scatter-structural-symmetries}
\end{figure}

We compare using greedy and canonical structural symmetries against not using state symmetries (Table~\ref{tab:results-structural-symmetries}). The results demonstrate that \cite{jefferson2019minimal}'s contributions allow a fast computation of canonical structural symmetries. Computing the canonical structural symmetries takes only $45\%$ more (aggregated) time per generated policy than computing the structural symmetries through hill-climbing for the solved tasks. On the other hand, the greedy structural symmetries provided by the hill-climbing procedure prove effective, as only $0.69\%$ more policy generations are required to solve the solved tasks on average, a difference that comes almost exclusively from the domain of \emph{islands}.

Both approaches improve AND$^*$'s performance to a coverage near of $0.73$, much superior to the coverage of $0.651$ we had before using state symmetries. The number of generated policies reduces by at least $25\%$ when using state symmetries on the domains of \emph{elevators}, \emph{faults}, \emph{zenotravel}, \emph{islands}, \emph{miner}, \emph{tireworld-spiky}, and \emph{tireworld-truck}. The reduction is astonishing in \emph{faults} ($-99.95\%$), \emph{tireworld-spiky} ($-99.59\%$), and \emph{tireworld-truck} ($-99.09\%$). Figure~\ref{fig:scatter-structural-symmetries} depicts the per-task differences in the number of generated policies between using canonical symmetries and not using structural symmetries. The gains in coverage occur in \emph{faults}, \emph{first-responders}, \emph{islands}, \emph{miner}, \emph{tireworld-spiky}, and \emph{tireworld-truck}. The major gains are in \emph{faults} (from $0.56$ to $1$) and \emph{miner} (from $0.24$ to $0.55$ and $0.53$). No change occurs in domains without structural symmetries besides the increase in execution time due to the execution of Nauty. The returned solution size remains, in general, mostly unchanged. We decide to stick with canonical structural state symmetries for the remainder of this work.

\section{Solution Compression}
\label{sec:comp}

In this section, we introduce a procedure, called \emph{the compressor}, that allows the compression of solution policies. The compressor receives as input a policy~$\pi$ defined over states, and returns a policy~$\tau$ defined over partial states that is \emph{optimal} (has the minimum number of partial states), such that~$\tau$ maps each state in the $\domain(\pi)$ to the same action it is mapped in~$\pi$, and~$\tau$ maps no state in the $\front(\pi)$. The central idea of the compressor is to use an integer program iteratively, which is fast in practice, to determine the partial states part of policy~$\tau$. The compressor is a general procedure and can compress solutions of any planner that finds policies defined over states.

\subsection{Partial States and Partial Policies}

A \emph{partial state} is a \emph{partial} function $p : \facts \rightharpoonup \{\true, \false\}$. Partial states represent simple properties regarding the truth valuation of facts, allowing the concise representation of the set of states with such a property. We note that we can view a state~$s$ as a total function $s : \facts \to \{\true, \false\}$ with $s[f] = \true$ iff $f \in \facts[s]$. We say a state $s$ \emph{holds} a property (partial state) $p$ -- denoted $s \models p$ -- iff, for each $f \in \facts$, $p[f]$ is undefined or $s[f] = p[f]$. The set of states represented by a partial state $p$ is $\states[p] = \{s : s \in \states \;\land\; s \models p\}$. We can perceive the task goal condition and the action preconditions as partial states.
A~\emph{partial policy} $\tau$ is a partial function mapping partial states to~actions. For a given state~$s$, we define $\tau[s] = \{\tau[p] : p \in \domain(\tau) \;\land\; s \models p\}$. Note that $\tau[s]$ is a set of actions. We define $\decompress(\tau) = \{s \mapsto a : s \in \states \;\land\; \tau[s] = \{a\}\}$.\break $\decompress(\tau)$ is a partial function mapping states to actions, and it maps a state $s$ iff $\tau[s]$ is a singleton. Therefore, there are two distinct ways for a state~$s$ to become unmapped on $\decompress(\tau)$. Namely, if either $\tau[s] = \emptyset$ or $|\tau[s]| \geq 2$. We use $\buggy(\tau) = \{s : s \in \states \;\land\; |\tau[s]| \geq 2\}$ to contain each state that becomes unmapped on $\decompress(\tau)$ because there are multiple partial states in~$\tau$ representing it, and they do not agree on the action. Let $\pruned(\pi) = \pi|_{\reach(\pi, \initialState)}$. A partial policy~$\tau_*$ is a \emph{solution partial policy} for $\Pi$ iff $\pruned(\decompress(\tau_*))$ is a solution policy for $\Pi$ and $\reach(\pruned(\decompress(\tau_*))) \cap \buggy(\tau_*) = \emptyset$.\footnote{We ensure no state~$s$ with $|\tau_*[s]| \geq 2$ is reachable because there are multiple ways to deal with such states. Setting $\decompress(\tau_*)[s] = \bot$ is just one possibility.
Therefore, if we were to let $\reach(\pruned(\decompress(\tau_*))) \cap \buggy(\tau_*) \neq \emptyset$, one would need to be informed about how to proceed at $\buggy(\tau_*)$ states in order to use~$\tau_*$ properly.
Some works set an external rule that defines which action should be used at~$s$ if $|\tau_*[s]| \geq 2$.
For instance, by presenting a total order for the mappings in $\tau_*$.}

\subsection{The Compressor}

Algorithm~\ref{alg:compressor} displays the pseudocode for the compressor.
The compressor allows the compression of any given $\pi$ into a partial policy $\tau$ such that $\decompress(\tau)|_{\reach(\pi)} = \pi$ and $\reach(\pi) \cap \buggy(\tau) = \emptyset$, with minimal $|\domain(\tau)|$.
In other words, it produces a partial policy $\tau$ such that $\decompress(\tau)$ agrees with $\pi$ for each $s \in \domain(\pi)$ and does not map any state $s \in \front(\pi)$. Additionally, it ensures $\tau[s] = \emptyset$ for each state $s \in \front(\pi)$.

\begin{algorithm}[h]
    \DontPrintSemicolon
    \SetKwIF{If}{ElseIf}{Else}{if}{:}{else if}{else:}{endif}
    \SetKwFor{While}{while}{:}{endw}
    \SetKwFor{ForEach}{for each}{:}{endfch}

    \textbf{Input:} $\pi$

    $\tau \coloneqq \emptyset$

    \ForEach{$a \in \actions$}
    {
        $X \coloneqq \{s : s \in \domain(\pi) \;\land\; \pi[s] = a\}$ \\
        $Y \coloneqq \{s : s \in \domain(\pi) \;\land\; \pi[s] \neq a\}$ $\sqcup\;\open(\pi)$ \\
        Find some minimal-sized set $Z$ of partial states such that $\forall s \in X: \exists p \in Z, \; (s \models p)$ and $\forall s \in Y: \lnot\exists p \in Z, \; (s \models p)$ \\
        Set $\tau[p] \coloneqq a$ for each $p \in Z$
    }
    \Return{$\tau$}

    \caption{Policy Compressor}
    \label{alg:compressor}
\end{algorithm}

In order for $\pi$ to agree with $\decompress(\tau)$ in the mapping of some state~$s \in \domain(\pi)$, $\tau[s]$ must be $\{\pi[s]\}$. Therefore, there must exist at least one partial state $p$ with $s \models p$ mapped to $\pi[s]$ in $\tau$ (otherwise $\pi[s] \not\in \tau[s]$), and there must exist no other partial state $p'$ with $s \models p'$ mapped to other action besides $\pi[s]$ in $\tau$ (otherwise $\tau[s] \not\subseteq \{\pi[s]\}$). That is enough to ensure an agreement of $\pi$ and $\decompress(\tau)$ regarding states in $\domain(\pi)$. We additionally enforce that no partial state in $\tau$ represents any state in $\front(\pi)$.

The compressor subdivides the original problem of deciding all the mappings of~$\tau$ into independent subproblems of deciding all the mappings to each action. For each action, it finds the minimal set of partial states that covers all states that need to be mapped to that action without covering any other domain or escape state from $\pi$. The resulting partial policy $\tau$ has $\decompress(\tau)|_{\reach(\pi)} = \pi$ since $\decompress(\tau)$ agrees with $\pi$ for all states in $\domain(\pi)$ and has $\decompress(\tau) = \bot$ for all states in $\front(\pi)$. Since we ensure $\tau[s] = \emptyset$ for each $s \in \front(\pi)$, we have $\reach(\pi) \cap \buggy(\tau) = \emptyset$. Therefore, the resulting partial policy $\tau$ is a solution partial policy when the input policy $\pi$ is a solution policy.

\subsection{Finding the Minimal-Sized Set of Partial States $Z$}

To find the minimal set of partial states~$Z$ that fulfills the required conditions of representing each $s \in X$ but no $s \in Y$, we repeatedly try to solve an integer programming problem with an increasing number of partial states, starting from one.

The following integer program models the task of determining whether there is a set of $k$ partial states representing each $s \in X$ but no $s \in Y$, or not.
$$
    \arraycolsep=1.25pt
    \begin{array}{lcrclclc}
        \multicolumn{5}{c}{\text{minimize} \hfill \sum\limits_{i = 1}^{k} \sum\limits_{f \in \facts} \sum\limits_{b \in \mathbb{B}} p_i[f \mapsto b] \hspace{12mm}} \\\\
        \multicolumn{3}{c}{\text{subject to} \hfill \sum\limits_{i = 1}^{k} p_i[s]} &\geq& 1 && \forall s \in X & (1) \\
        && \sum\limits_{i = 1}^{k} p_i[s] &=& 0 && \forall s \in Y & (2) \\
        && p_i[f \mapsto \neg s[f]] &\leq& 1 - p_i[s] && \forall i \in [1, k], \forall s \in X \sqcup Y, \forall f \in \facts & (3) \\
        && \sum\limits_{f \in \facts} p_i[f \mapsto \neg s[f]] &\geq& 1 - p_i[s] && \forall i \in [1, k], \forall s \in X \sqcup Y & (4) \\
        && p_i[f \mapsto \true] + p_i[f \mapsto \false] &\leq& 1 && \forall i \in [1, k], \forall f \in \facts & (5) \\
        && p_i[s] &\in& \mathbb{B} && \forall i \in [1, k], \forall s \in X \sqcup Y \\
        && p_i[f \mapsto b] &\in& \mathbb{B} && \forall i \in [1, k], \forall f \in \facts, \forall b \in \mathbb{B}. \\
    \end{array}
$$

For each partial state $p_i$, with $i \in \{1, 2, \dots, k\}$, we have a boolean variable $p_i[f \mapsto b]$ for each $f \in \facts$ and $b \in \mathbb{B}$,\footnote{Since our implementation uses SAS+ FOND planning tasks instead of STRIPS FOND planning tasks the actual IPs we produce has variables $p_i[x \mapsto v]$, where $x$ is a variable and $v \in \domain(x)$. Despite we present the IPs for STRIPS FOND planning tasks, every aspect described here can be directly translated to an analogous aspect we would have in IPs for SAS+ FOND planning tasks.} and a boolean variable $p_i[s]$ for each $s \in X \sqcup Y$. $p_i[f \mapsto b] = 1$ means $p_i[f] = b$ and $p_i[s] = 1$ means $s \models p_i$.

The first set of constraints ensures each $s \in X$ is represented by some partial state $p_i$. The second set of constraints ensures each $s \in Y$ is \emph{not}~represented by any partial state $p_i$. The third and fourth sets of constraints ensure the semantics of $\models$. The third enforces $p_i$ \emph{not} to map any fact $f$ differently from $s$ if $s \models p_i$. While the fourth enforces $p_i$ to map some fact $f$ differently from $s$ if $s \not\models p_i$. The fifth set of constraints ensures $p_i$ maps each fact to at most one truth valuation.

Since we try to solve the problem first with $k = 1$, then with $k = 2$, and so on, until it is solvable (and certainly it will be if $k$ reaches $|X|$), we do not need to model the optimization of $k$ in the integer program. The optimization function minimizes instead the total number of facts mapped through the fixed $k$ partial states. While we could instead minimize zero, in order to have an easier integer program, this provides a much more human-readable output and also provides heuristic guidance for the integer programming solver we use.

We now simplify this integer program.
Firstly, we note that since every $p_i[s]$ must be zero if $s \in Y$, then we can fully remove the second set of constraints by replacing these variables with a constant zero in every other constraint they appear. This makes the third set of constraints trivial for $s \in Y$, so we make the third set of constraints exist only for $s \in X$.

Secondly, we note that the fourth set of constraints is irrelevant for $s \in X$. They hinder $\sum_{f \in \facts} p_i[f \mapsto \neg s[f]] = 0$ from occurring simultaneously with $p_i[s] = 0$ for some $s \in X$ and some $p_i$. However, allowing that causes no problem. Whenever that situation is possible, considering the other constraints, changing $p_i[s]$ to $1$ is also possible. Thus no extra solution is ever permitted by removing the constraints for $s \in X$ in the fourth set of constraints. Moreover, we do not use the values of variables $p_i[s]$ when extracting the set of partial states $p_i$, so they do not need to be precise. Therefore, we make the fourth set of constraints exist only for $s \in Y$.

Thirdly, the fifth set of constraints is irrelevant because it can only hurt to set both $p_i[f \mapsto \true]$ and $p_i[f \mapsto \false]$ for some fact $f$, since this makes $p_i$ useless as it will not be able to represent any state in $X$. And since we only try solving with $k$ partial states after failing to solve with $k-1$ (or if $k-1 = 0$), no solution has some useless $p_i$. So we remove these constraints.

The integer program resulting from these simplifications is shown below.
$$
    \arraycolsep=1.4pt
    \begin{array}{lcrclcl}
        \text{minimize} && \multicolumn{3}{c}{\sum\limits_{i = 1}^{k} \sum\limits_{f \in \facts} \sum\limits_{b \in \mathbb{B}} p_i[f \mapsto b]} \\\\
        \multicolumn{3}{c}{\text{subject to} \hfill \sum\limits_{i = 1}^{k} p_i[s]} &\geq& 1 && \forall s \in X \\
        && p_i[f \mapsto \neg s[f]] &\leq& 1 - p_i[s] && \forall i \in [1, k], \forall s \in X, \forall f \in \facts \\
        && \sum\limits_{f \in \facts} p_i[f \mapsto \neg s[f]] &\geq& 1 && \forall i \in [1, k], \forall s \in Y \\
        && p_i[s] &\in& \mathbb{B} && \forall i \in [1, k], \forall s \in X \\
        && p_i[f \mapsto b] &\in& \mathbb{B} && \forall i \in [1, k], \forall f \in \facts, \forall b \in \mathbb{B}. \\
    \end{array}
$$

Additionally, we note that we just need to create a variable $p_i[f \mapsto b]$ only if there is at least one state $s \in X$ with $s[f] = b$. Otherwise, this variable is useless, as it will always be zero in some solution.

The approach of first subdividing the original compression problem into a subproblem for each action and then trying to solve it with an increasing number~$k$ of partial states using this optimized IP formulation was critical to have a sufficiently fast and memory-efficient procedure to compress the solutions returned by AND$^*$.

\subsection{Empirical Evaluation}

\color{revbcolor}
We toggle on some satisficing features to AND$^*$, to further improve its coverage, and we test the compressor on the solution policies returned by the resulting new version of AND$^*$.

\subsubsection{Enabling a Few Satisficing Features to AND$^*$}

We toggle on the \emph{deadlock detection} presented by \cite{messa2023best}, which prunes policies with deadlocks upon generation. This procedure potentially makes AND$^*$ miss more solutions, as it may prune a policy with deadlocks that would otherwise be fixed by the concretizer\footnote{AND$^*$ with deadlock detection and domain-escape pruning is incomplete even if we enable the concretizer.}, but helps to focus the search on the parts of the policy space that are more likely to lead to a solution. We also change the evaluation of policies from $f(\pi) = \Delta_\downarrow(\pi)$ to $f(\pi) = 2\Delta_\downarrow(\pi) - |\domain(\pi)|$. This is equivalent to making AND$^*$ a ``2-Weighted AND$^*$'', inspired by Weighted~A$^*$ (WA$^*$) \citep{pohl1970heuristic}. Finally, we change the classical planning heuristic used by $\Delta_\downarrow(\pi)$ from $\hLMCut$~\citep{helmert2009landmarks} to $\hFF$~\citep{hoffmann2001ff}. Each of these three changes would alone make original AND$^*$ lose optimality, which is one of the reasons we did not apply them from the start (up to Section~\ref{ssec:psym-io} we were still aiming for optimality). We also did not apply these changes earlier to make a clean comparison between the different pruning methods introduced, over a baseline policy-space search algorithm.

\begin{figure}[bt]
    \centering
    \includegraphics[width=.46\textwidth]{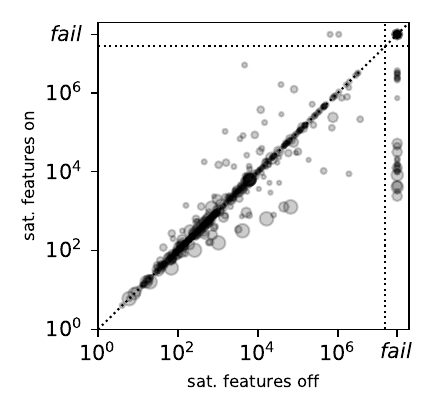}
    \caption{\revb{Per-task comparison on the number of generated policies before and after togging on the satisficing features.}}
    \label{fig:scatter-satisficing}
\end{figure}

\begin{table}[hp]\color{revbcolor}
\begin{subtable}[t]{\textwidth}
    \centering
    \setlength\tabcolsep{5pt}
    \fontsize{9}{10}\selectfont
    \begin{adjustbox}{center}
    \begin{tabular}{l|ccc|ccc}
        \toprule
        \multicolumn{1}{c}{} & \multicolumn{3}{c}{sat.\! features off} & \multicolumn{3}{c}{sat.\! features on} \\

        \cmidrule[\heavyrulewidth](lr){2-4} \cmidrule[\heavyrulewidth](lr){5-7}

        \textbf{Domain} & $\%\mathrm{S}$ & $\%\mathrm{T}$ & $\%\mathrm{M}$ & $\%\mathrm{S}$ & $\%\mathrm{T}$ & $\%\mathrm{M}$ \\

        \midrule
        acrobatics & \textbf{1} & \textcolor{gray}{0} & \textcolor{gray}{0} & \textbf{1} & \textcolor{gray}{0} & \textcolor{gray}{0} \\
        beam-walk & 0.82 & 0.18 & \textcolor{gray}{0} & \textbf{1} & \textcolor{gray}{0} & \textcolor{gray}{0} \\
        blocksworld-original & 0.53 & 0.47 & \textcolor{gray}{0} & \textbf{1} & \textcolor{gray}{0} & \textcolor{gray}{0} \\
        blocksworld-advanced & 0.20 & 0.71 & 0.09 & 0.82 & 0.18 & \textcolor{gray}{0} \\
        chain-of-rooms & \textbf{1} & \textcolor{gray}{0} & \textcolor{gray}{0} & \textbf{1} & \textcolor{gray}{0} & \textcolor{gray}{0} \\
        earth-observation & 0.33 & 0.15 & 0.53 & 0.45 & \textcolor{gray}{0} & 0.55 \\
        elevators & 0.80 & \textcolor{gray}{0} & 0.20 & 0.93 & \textcolor{gray}{0} & 0.07 \\
        faults & \textbf{1} & \textcolor{gray}{0} & \textcolor{gray}{0} & \textbf{1} & \textcolor{gray}{0} & \textcolor{gray}{0} \\
        first-responders & 0.75 & 0.24 & 0.01 & 0.99 & \textcolor{gray}{0} & 0.01 \\
        tireworld-triangle & \textbf{1} & \textcolor{gray}{0} & \textcolor{gray}{0} & \textbf{1} & \textcolor{gray}{0} & \textcolor{gray}{0} \\
        zenotravel & 0.33 & 0.67 & \textcolor{gray}{0} & 0.87 & 0.13 & \textcolor{gray}{0} \\
        \midrule
        doors & \textbf{1} & \textcolor{gray}{0} & \textcolor{gray}{0} & \textbf{1} & \textcolor{gray}{0} & \textcolor{gray}{0} \\
        islands & 0.80 & 0.13 & 0.07 & 0.75 & \textcolor{gray}{0} & 0.25 \\
        miner & 0.53 & 0.47 & \textcolor{gray}{0} & 0.59 & 0.18 & 0.24 \\
        tireworld-spiky & \textbf{1} & \textcolor{gray}{0} & \textcolor{gray}{0} & \textbf{1} & \textcolor{gray}{0} & \textcolor{gray}{0} \\
        tireworld-truck & 0.57 & \textcolor{gray}{0} & 0.43 & \textbf{1} & \textcolor{gray}{0} & \textcolor{gray}{0} \\
        \midrule
        TOTAL & 0.728 & 0.189 & 0.083 & 0.900 & 0.031 & 0.070 \\
        \bottomrule
    \end{tabular}
    \end{adjustbox}
    \caption{Ratio of successful, timeout, and memory-out runs per configuration.}
\end{subtable}

\bigskip

\begin{subtable}[t]{\textwidth}
    \centering
    \setlength\tabcolsep{5pt}
    \fontsize{9}{10}\selectfont
    \begin{adjustbox}{center}
    \begin{tabular}{lc|rrr|rrr}
        \toprule
        \multicolumn{2}{c}{} & \multicolumn{3}{c}{sat.\! features off} & \multicolumn{3}{c}{sat.\! features on} \\

        \cmidrule[\heavyrulewidth](lr){3-5} \cmidrule[\heavyrulewidth](lr){6-8}

        \textbf{Domain} & $\%{\cap}\mathrm{S}$ & $t(s)$ & $\#\pi$ & $|\pi_*|$ & $t(s)$ & $\#\pi$ & $|\pi_*|$ \\

        \midrule
        acrobatics & \textbf{1} & 3.36 & 11,109 & 126.50 & 1.44 & 318 & 126.50 \\
        beam-walk & 0.82 & 2.45 & 454 & 453.22 & 3.86 & 454 & 453.22 \\
        blocksworld-original & 0.53 & 229.86 & 270,708 & 14.12 & 0.91 & 482 & 15.44 \\
        blocksworld-advanced & 0.20 & 0.49 & 1,730 & 9.64 & 0.20 & 220 & 9.91 \\
        chain-of-rooms & \textbf{1} & 33.29 & 7,968 & 162.00 & 0.39 & 458 & 162.00 \\
        earth-observation & 0.32 & 2.74 & 219,172 & 26.08 & 0.93 & 69,272 & 26.69 \\
        elevators & 0.80 & 6.56 & 161,881 & 25.33 & 0.56 & 1,790 & 28.75 \\
        faults & \textbf{1} & 0.87 & 1,050 & 24.31 & 0.70 & 123 & 23.00 \\
        first-responders & 0.75 & 127.49 & 297,969 & 11.66 & 0.94 & 2,114 & 13.18 \\
        tireworld-triangle & \textbf{1} & 30.31 & 485 & 244.00 & 5.99 & 485 & 244.00 \\
        zenotravel & 0.33 & 2.72 & 689 & 14.80 & 0.89 & 174 & 15.60 \\
        \midrule
        doors & \textbf{1} & 117.27 & 17,484 & 17,473.73 & 136.59 & 17,484 & 17,473.73 \\
        islands & 0.75 & 31.10 & 109,284 & 6.11 & 54.24 & 352,310 & 6.11 \\
        miner & 0.49 & 513.65 & 422,493 & 16.28 & 141.26 & 662,302 & 16.64 \\
        tireworld-spiky & \textbf{1} & 4.75 & 5,913 & 25.73 & 1.37 & 6,712 & 25.73 \\
        tireworld-truck & 0.57 & 1.25 & 5,618 & 14.64 & 0.79 & 3,000 & 14.55 \\
        \midrule
        TOTAL & 0.723 & 11.13 & 13,869 & 46.96 & 2.34 & 2,439 & 48.16 \\
        \bottomrule
    \end{tabular}
    \end{adjustbox}
    \caption{Average runtime, number of generated policies, and size of returned solutions, per configuration, over the intersection of solved tasks.}
\end{subtable}
\caption{Empirical results for AND$^*$ with and without the satisficing features.}
\label{tab:results-satisficing}
\end{table}

Table~\ref{tab:results-satisficing} shows the results of AND$^*$ with and without the satisficing features. We verify an increase in coverage from 0.728 to 0.900 after enabling the satisficing features. The highest increases occur in \emph{blocksworld-original} (from 0.53 to 1), \emph{blocksworld-advanced} (0.2 to 0.82), and \emph{zenotravel} (0.33 to 0.87). Only in \emph{islands} we observe a decrease in coverage (from 0.80 to 0.75).

Regarding the number of generated policies, the results are mixed. The average number of generated policies in the intersection of solved tasks substantially decreases in all but two IPC-FOND domains (all but \emph{beam-walk} and \emph{tireworld-triangle}), but increases in three NEW-FOND domains (\emph{islands}, \emph{miner} and \emph{tireworld-spiky}). Figure~\ref{fig:scatter-satisficing} shows these mixed results, but also shows that for most of the tasks only AND$^*$ with the satisficing features solves within resource limits; it only needs to generate around 10,000 policies to solve them.

\color{black}

\subsubsection{Compression Results}

 We use IBM's ILOG CPLEX to solve integer programming problems.

\begin{table}[htb]
    \centering
    \setlength\tabcolsep{5pt}
    \fontsize{9}{10}\selectfont
    \begin{adjustbox}{center}
    \begin{tabular}{lc|rr}
        \toprule

        \textbf{Domain} & $\%{\cap}\mathrm{S}$ & $|\pi_*|$ & $|\tau_*|$ \\

        \midrule
        acrobatics & \textbf{1} & 126.50 & 126.50 \\
        beam-walk & \textbf{1} & 1,487.73 & 1,487.73 \\
        blocksworld-original & \textbf{1} & 25.20 & 24.77 \\
        blocksworld-advanced & 0.82 & 41.56 & 37.76 \\
        chain-of-rooms & \textbf{1} & 162.00 & 162.00 \\
        earth-observation & 0.45 & 34.17 & 28.78 \\
        elevators & 0.93 & 33.79 & 29.71 \\
        faults & \textbf{1} & 23.00 & 14.18 \\
        first-responders & 0.99 & 16.84 & 15.42 \\
        tireworld-triangle & \textbf{1} & 244.00 & 163.00 \\
        zenotravel & 0.87 & 37.38 & 36.92 \\
        \midrule
        doors & \textbf{1} & 17,473.73 & 18.00 \\
        islands & 0.75 & 6.11 & 6.11 \\
        miner & 0.59 & 16.97 & 16.97 \\
        tireworld-spiky & \textbf{1} & 25.73 & 22.36 \\
        tireworld-truck & \textbf{1} & 18.64 & 14.69 \\
        \midrule
        TOTAL & 0.900 & 65.44 & 38.09 \\
        \bottomrule
    \end{tabular}
    \end{adjustbox}
\caption{\rev{Comparison on the size of the returned solutions before and after compression.}}
\label{tab:results-compression}
\end{table}
 
 For each solved task, the compressor was able to compress the returned solution within the memory and time limits that remained after the execution performed by AND$^*$. For the majority of the tasks, we are able to compress the solutions returned in less than 20 ms. Besides \emph{p14} and \emph{p15} of \emph{doors}, and \emph{p11} of \emph{beam-walk}, all other solved tasks had their returned solutions compressed in at most 12 seconds. The most time-consuming compression (on \emph{p15} of \emph{doors}) took 55 seconds to complete. Only four domains have some task for which the returned solution took more than 250 ms to be compressed: \emph{beam-walk}, \emph{chain-of-rooms}, \emph{doors}, and \emph{tireworld-triangle}. The domain of \emph{doors} is especially hard because, while the largest solution returned for some other domain has size 8,191 (from \emph{p11} of \emph{beam-walk}), the optimal solution size of \emph{p15} of \emph{doors} is 131,070 before compression.

Table~\ref{tab:results-compression} shows that the compressor reduces the size of the returned solution in 11 of the 16 domains. Decreases of at least 9$\%$ occur in eight domains: \emph{blocksworld-advanced}, \emph{earth-observation}, \emph{elevators}, \emph{faults}, \emph{tireworld-triangle}, \emph{doors}, \emph{tireworld-spiky}, and \emph{tireworld-truck}. In \emph{doors}, the average size of returned solutions decreases by three orders of magnitude.

\rev{We note that if we were to run the compressor on the solution policies returned by a configuration of AND$^*$ that returns solution policies with minimal domain size, the resulting compressed solutions would \emph{not} necessarily be the most compact possible solution partial policies for the tasks. The compressor only ensures that the resulting compressed solution is the most compact solution partial policy that agrees with the input solution policy.}

\revb{We also note that even though in some contexts performing data compression implies waiving human-readability, most of our compression gains came together with making solutions more human-readable. A clear example is \emph{doors}, not only due to the dramatic decreases in the number of partial states, but also due to decreases in facts per mapping. Compressed solutions for \emph{doors} had always either one or two facts per partial state, while complete states for \emph{p15}, for example, have 34 facts each. We argue these decreases make  the solution much easier to understand.}

In the next section, we compare the after-compression size of AND$^*$ returned solutions with the size of solutions returned by two state-of-the-art FOND planners that natively reason over partial states.

\section{Comparison with the State of the Art}
\label{sec:sota}

In this section, we compare the best version of AND$^*$ (with escape equivalence pruning, structural symmetries, solution compression, and the satisficing features enabled in the previous section) with some FOND planners from the literature. We choose to compare AND$^*$ with PRP \citep{muise2012improved}, FOND-SAT \citep{geffner2018compact}, IDFSP \citep{pereira2022iterative}\rev{, CFOND-ASP \citep{yadav2023declarative}, and PR2 \citep{muise2024prp}}.

\subsection{State-of-the-Art Planners}

PRP \citep{muise2012improved} is a long-standing state-of-the-art satisficing FOND planner. PRP uses a classical planner to find trajectories to the~goal. Starting by finding a trajectory from the initial state to the goal, it constructs a partial policy by repeatedly generalizing each new trajectory found using partial states. This process repeats until all the states reachable by the decompressed version of the current partial policy are covered, or until it concludes that one of them is a dead end. In the case of the latter, it learns a constraint that prevents the same error from occurring again after it restarts the search.

FOND-SAT \citep{geffner2018compact} compiles FOND planning tasks into SAT instances. When trying to solve a FOND planning task, it constructs a series of SAT instances that encode requirements a partial policy (with an increasing number of partial states) would need to have in order to be a solution for the given task, until one of them is solved.
From the solution of the solved SAT instance, it extracts a compact solution partial policy for the FOND planning task.
FOND-SAT aims to have good scalability in tasks with high degrees of non-determinism due to the fact that different branching futures can be dealt with in conjunction with the SAT formulation.

IDFSP \citep{pereira2022iterative} is a state-of-the-art satisficing FOND planner. IDFSP solves FOND planning tasks by performing an iterative depth-first \emph{implicit} policy-space search. Bounded by an increasing search depth limit, it first tries to find a trajectory from the initial state to the goal. By doing so, it implicitly maps each state in the sequence to an action. Then, in a depth-first manner, it tries to find a goal trajectory for each other state reachable by this implicit policy. In case of a dead end or failure, it backtracks, removing mappings from the implicit policy. It uses different rules to cut short the exploration in some directions, taking into account the current search depth limit, the heuristic value of the states, and the previous explorations of those states. In case of ultimate failure, it restarts the search with an increased search depth limit.

\color{revcolor}

CFOND-ASP \citep{yadav2023declarative}, like FOND-SAT, compiles\break FOND planning tasks into another formalism. Specifically, it translates FOND planning tasks into Answer Set Programming (ASP)~\citep{lifschitz2002answer} instances. Similar to FOND-SAT and PRP, CFOND-ASP generates a compact solution representation using partial states. A key distinguishing feature of CFOND-ASP is its novel ability to produce compact solutions that retain the best possible action for each complete state, unlike FOND-SAT, whose SAT-based formulation inherently trades state-specific decision precision for increased compactness.

PRP Rebooted (or PR2 for short)~\citep{muise2024prp} is contemporary to this work. It makes several advancements over the original PRP, establishing itself as an unrivaled new state-of-the-art satisficing FOND planner. A key innovation in PR2 lies in its internal representation of the search states. It uses a pair of interconnected structures that efficiently link the explored state space with the incumbent partial policy. This dual-structure design accurately captures most of the inherent non-determinism required to handle dead ends and other critical aspects of FOND planning during the search process. Additionally, PR2 incorporates several clever engineering optimizations that exploit the structure of common FOND planning tasks to further enhance its performance.

\color{black}

We ran each of the FOND planners in their recommended settings. We compare them with the best version of AND$^*$.

\begin{table}[hp]\color{revcolor}
\begin{subtable}[t]{\textwidth}
    \centering
    \setlength\tabcolsep{5pt}
    \fontsize{9}{10}\selectfont
    \begin{adjustbox}{center}
    \begin{tabularx}{1.2\textwidth}{l| *{6}{Y} }
        \toprule
        \multicolumn{1}{c}{} & \multicolumn{1}{c}{FOND-SAT} & \multicolumn{1}{c}{CFOND-ASP} & \multicolumn{1}{c}{IDFSP} & \multicolumn{1}{c}{PRP} & \multicolumn{1}{c}{AND$^*$} & \multicolumn{1}{c}{PR2} \\

        \cmidrule[\heavyrulewidth](lr){2-2} \cmidrule[\heavyrulewidth](lr){3-3} \cmidrule[\heavyrulewidth](lr){4-4} \cmidrule[\heavyrulewidth](lr){5-5} \cmidrule[\heavyrulewidth](lr){6-6} \cmidrule[\heavyrulewidth](lr){7-7}

        \textbf{Domain} & $\%\mathrm{S}$ & $\%\mathrm{S}$ & $\%\mathrm{S}$ & $\%\mathrm{S}$ & $\%\mathrm{S}$ & $\%\mathrm{S}$ \\

        \midrule
        acrobatics & 0.38 & 0.50 & \textbf{1} & \textbf{1} & \textbf{1} & \textbf{1} \\
        beam-walk & 0.18 & 0.27 & \textbf{1} & \textbf{1} & \textbf{1} & \textbf{1} \\
        blocksworld-original & 0.33 & 0.33 & 0.97 & \textbf{1} & \textbf{1} & \textbf{1} \\
        blocksworld-advanced & 0.24 & 0.20 & 0.75 & \textbf{1} & 0.82 & \textbf{1} \\
        chain-of-rooms & \textcolor{gray}{0} & 0.10 & \textbf{1} & \textbf{1} & \textbf{1} & \textbf{1} \\
        earth-observation & 0.12 & 0.17 & 0.65 & \textbf{1} & 0.45 & \textbf{1} \\
        elevators & 0.47 & 0.47 & 0.53 & \textbf{1} & 0.93 & \textbf{1} \\
        faults & 0.51 & 0.67 & \textbf{1} & \textbf{1} & \textbf{1} & \textbf{1} \\
        first-responders & 0.59 & 0.55 & 0.69 & \textbf{1} & 0.99 & \textbf{1} \\
        tireworld-triangle & 0.05 & 0.07 & 0.23 & \textbf{1} & \textbf{1} & \textbf{1} \\
        zenotravel & 0.33 & 0.33 & 0.60 & \textbf{1} & 0.87 & \textbf{1} \\
        \midrule
        doors & 0.67 & \textbf{1} & 0.93 & 0.80 & \textbf{1} & \textbf{1} \\
        islands & 0.98 & \textbf{1} & \textbf{1} & 0.52 & 0.75 & \textbf{1} \\
        miner & 0.94 & 0.98 & \textbf{1} & 0.27 & 0.59 & \textbf{1} \\
        tireworld-spiky & 0.18 & 0.91 & 0.91 & 0.09 & \textbf{1} & \textbf{1} \\
        tireworld-truck & 0.85 & \textbf{1} & 0.68 & 0.31 & \textbf{1} & \textbf{1} \\
        \midrule
        TOTAL & 0.426 & 0.535 & 0.808 & 0.812 & 0.900 & \textbf{1} \\
        \bottomrule
    \end{tabularx}
    \end{adjustbox}
    \caption{\rev{Coverages per FOND planner.}}
\end{subtable}

\bigskip

\begin{subtable}[t]{\textwidth}
    \centering
    \setlength\tabcolsep{5pt}
    \fontsize{9}{10}\selectfont
    \begin{adjustbox}{center}
    \begin{tabular}{lc|rr|rr|r|rr|rr|rr}
        \toprule
        \multicolumn{2}{c}{} & \multicolumn{2}{c}{FOND-SAT} & \multicolumn{2}{c}{CFOND-ASP} & \multicolumn{1}{c}{IDFSP} & \multicolumn{2}{c}{PRP} & \multicolumn{2}{c}{AND$^*$} & \multicolumn{2}{c}{PR2} \\

        \cmidrule[\heavyrulewidth](lr){3-4} \cmidrule[\heavyrulewidth](lr){5-6} \cmidrule[\heavyrulewidth](lr){7-7} \cmidrule[\heavyrulewidth](lr){8-9} \cmidrule[\heavyrulewidth](lr){10-11} \cmidrule[\heavyrulewidth](lr){12-13}

        \textbf{Domain} & $\%{\cap}\mathrm{S}$ & $t(s)$ & $|\tau_*|$ & $t(s)$ & $|\tau_*|$ & $t(s)$ & $t(s)$ & $|\tau_*|$ & $t(s)$ & $|\tau_*|$ & $t(s)$ & $|\tau_*|$ \\

        \midrule
        acrobatics & 0.38 & 8.09 & 8.33 & 1.07 & 8.33 & 0.02 & 2.55 & 8.33 & 0.04 & 8.33 & 0.03 & 8.33 \\
        beam-walk & 0.18 & 2.85 & 11.00 & 0.31 & 11.00 & 0.01 & 0.14 & 11.00 & 0.04 & 11.00 & 0.04 & 11.00 \\
        blocksworld-original & 0.33 & 27.76 & 10.10 & 9.85 & 10.50 & 0.06 & 0.01 & 10.70 & 0.41 & 10.70 & 0.01 & 10.70 \\
        blocksworld-advanced & 0.20 & 27.04 & 8.82 & 8.57 & 8.82 & 0.06 & 0.01 & 9.82 & 0.21 & 9.55 & 0.01 & 9.82 \\
        chain-of-rooms & \textcolor{gray}{0} & - & - & - & - & - & - & - & - & - & - & - \\
        earth-observation & 0.12 & 6.08 & 8.80 & 0.76 & 8.80 & 0.02 & 0.01 & 9.60 & 0.04 & 8.60 & 0.00 & 9.60 \\
        elevators & 0.47 & 98.65 & 14.86 & 5.46 & 14.86 & 0.04 & 0.00 & 16.71 & 0.70 & 15.86 & 0.01 & 16.29 \\
        faults & 0.51 & 256.62 & 10.25 & 13.25 & 11.25 & 0.07 & 0.01 & 10.25 & 0.67 & 10.25 & 0.01 & 10.25 \\
        first-responders & 0.49 & 79.13 & 8.30 & 94.82 & 8.30 & 58.12 & 0.18 & 8.86 & 0.75 & 8.59 & 0.01 & 8.84 \\
        tireworld-triangle & 0.05 & 18.90 & 11.00 & 2.34 & 11.00 & 0.03 & 0.01 & 16.00 & 0.05 & 11.00 & 0.00 & 11.00 \\
        zenotravel & 0.33 & 205.84 & 14.80 & 639.85 & 14.80 & 0.51 & 0.02 & 21.60 & 0.90 & 15.40 & 0.01 & 21.60 \\
        \midrule
        doors & 0.47 & 306.89 & 16.00 & 18.52 & 16.00 & 0.40 & 0.04 & 16.00 & 0.46 & 16.00 & 0.01 & 16.00 \\
        islands & 0.52 & 11.25 & 5.84 & 1.20 & 5.84 & 0.06 & 49.16 & 5.84 & 1.67 & 5.84 & 0.00 & 5.84 \\
        miner & 0.24 & 129.23 & 15.58 & 32.79 & 15.58 & 0.20 & 197.02 & 17.75 & 58.37 & 16.17 & 0.01 & 15.83 \\
        tireworld-spiky & 0.09 & 419.76 & 22.00 & 34.75 & 22.00 & 0.07 & 20.04 & 22.00 & 0.07 & 22.00 & 0.01 & 22.00 \\
        tireworld-truck & 0.26 & 123.07 & 12.11 & 2.43 & 12.11 & 32.15 & 170.05 & 23.05 & 0.48 & 12.32 & 1.34 & 15.84 \\
        \midrule
        TOTAL & 0.290 & 48.43 & 11.23 & 7.50 & 11.33 & 0.15 & 0.21 & 12.81 & 0.35 & 11.47 & 0.01 & 12.06 \\
        \bottomrule
    \end{tabular}
    \end{adjustbox}
    \caption{\rev{Average runtime and size of returned partial-state solutions, per FOND planner, over the intersection of solved tasks.}}
\end{subtable}
\caption{\rev{Empirical results for the state-of-the-art FOND planners.}}
\label{tab:results-state-of-the-art}
\end{table}

\begin{figure}[tbp]
    \centering
    \includegraphics[width=.75\textwidth]{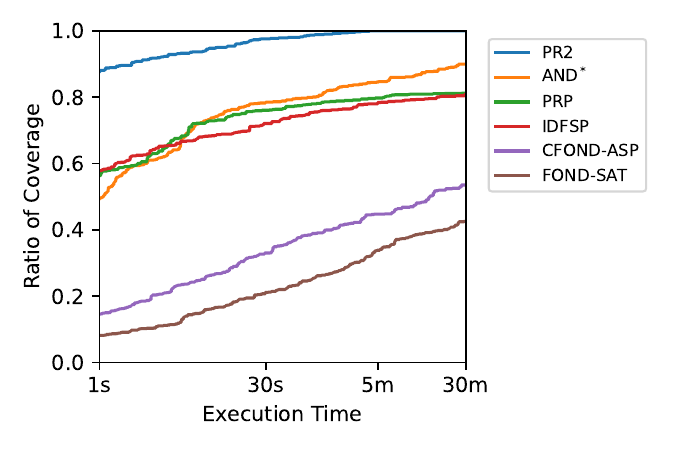}
    \caption{\rev{Aggregated ratio of solved tasks over time, for the different FOND planners.}}
    \label{fig:evolution-state-of-the-art}
\end{figure}

\begin{table}[t]\color{revbcolor}
    \centering
    \setlength\tabcolsep{5pt}
    \fontsize{9}{10}\selectfont
    \begin{adjustbox}{center}
\begin{tabularx}{1.2\textwidth}{l| *{6}{Y} }
    \toprule
    \multicolumn{1}{c}{} & \multicolumn{1}{c}{FOND-SAT} & \multicolumn{1}{c}{CFOND-ASP} & \multicolumn{1}{c}{IDFSP} & \multicolumn{1}{c}{PRP} & \multicolumn{1}{c}{AND$^*$} & \multicolumn{1}{c}{PR2} \\
    \cmidrule[\heavyrulewidth](lr){2-2} \cmidrule[\heavyrulewidth](lr){3-3} \cmidrule[\heavyrulewidth](lr){4-4} \cmidrule[\heavyrulewidth](lr){5-5} \cmidrule[\heavyrulewidth](lr){6-6} \cmidrule[\heavyrulewidth](lr){7-7}
    \textbf{Benchmark} & $\%\mathrm{S}$ & $\%\mathrm{S}$ & $\%\mathrm{S}$ & $\%\mathrm{S}$ & $\%\mathrm{S}$ & $\%\mathrm{S}$ \\
    \midrule
    IPC-FOND & 0.291 & 0.333 & 0.765 & \textbf{1} & 0.915 & \textbf{1} \\
    NEW-FOND & 0.724 & 0.978 & 0.904 & 0.398 & 0.868 & \textbf{1} \\
    \bottomrule
\end{tabularx}
    \end{adjustbox}
\caption{\revb{Aggregated coverage per benchmark for the state-of-the-art FOND planners.}}
\label{tab:results-state-of-the-art-2}
\end{table}

\subsection{Coverage Results}

\color{revcolor}

Table~\ref{tab:results-state-of-the-art} presents part of the results. From it, we can verify that AND$^*$ has the second best ratio of solved tasks (0.900), followed by PRP and IDFSP, which have similar ratios of solved tasks (0.812 and 0.808, respectively), then CFOND-ASP (0.535), and finally FOND-SAT (0.426). PR2 is the only planner that solves all tasks within the time and memory limits.
Figure~\ref{fig:evolution-state-of-the-art} shows that AND$^*$ has a slower start in comparison to PRP and IDFSP, and a better scalability with more resources.

We also verified that if we were to create a hybrid planner that can run any of the planners for a pre-determined amount of time, the combination that would be able to solve all the tasks allocating the least summed amount of pre-determined time (the \emph{virtual best solver}) would be the one that runs AND$^*$ for 14.8s and PR2 for 27.1s, totaling 41.9s.

If we were to remove the option of using AND$^*$, the virtual best solver would be to run \emph{only} PR2 for 250.9s (that is the time it takes to solve the hardest task from \emph{tireworld-triangle}). In other words, among the tested planners, AND$^*$ is, in some sense, the only one that still aggregates empirical value for solving this specific benchmark, after the introduction of PR2.

On the other hand, if we were to remove only the option of using PR2, the virtual best solver would be to run AND$^*$ for 14.8s, PRP for 17.7s, and IDFSP for 21.6s, totaling 54.1s, lower than the allocated time required when only the option of using AND$^*$ is removed.
This indicates that AND$^*$ might introduce a distinguished set of strengths that better complement the capabilities of previous FOND planners.

PR2, AND$^*$, and IDFSP are the only tested planners that achieve high coverage in both the NEW-FOND and the IPC-FOND benchmarks.

\subsection{Compactness Results}

\begin{table}[ht]\color{revcolor}
\setlength\tabcolsep{5pt}
\centering
\makebox[\textwidth][c]{
\begin{tabular}{cc}

\begin{subtable}[t]{.51\textwidth}
    \centering
    \setlength\tabcolsep{3pt}
    \fontsize{9}{10}\selectfont
    \begin{tabular}{lc|rcr}
        \toprule

        \textbf{Domain} & $\%{\cap}\mathrm{S}$ & $|\tau_*|$ && $|\tau_*|$ \\

        \midrule
        blocksworld-original & \textbf{1} & 26.63 & $>$ & 24.77 \\
        blocksworld-advanced & 0.82 & 36.91 & $<$ & 37.76 \\
        earth-observation & 0.45 & 37.89 & $>$ & 28.78 \\
        elevators & 0.93 & 38.43 & $>$ & 29.71 \\
        first-responders & 0.99 & 16.81 & $>$ & 15.42 \\
        tireworld-triangle & \textbf{1} & 244.00 & $>$ & 163.00 \\
        zenotravel & 0.87 & 49.38 & $>$ & 36.92 \\
        \midrule
        miner & 0.24 & 17.75 & $>$ & 16.17 \\
        tireworld-truck & 0.31 & 22.39 & $>$ & 12.83 \\
        \bottomrule
    \end{tabular}
    \caption{\rev{PRP vs. AND$^*$}}
\end{subtable}
&
\begin{subtable}[t]{.48\textwidth}
    \centering
    \setlength\tabcolsep{3pt}
    \fontsize{9}{10}\selectfont
    \begin{tabular}{lc|rcr}
        \toprule

        \textbf{Domain} & $\%{\cap}\mathrm{S}$ & $|\tau_*|$ && $|\tau_*|$ \\

        \midrule
        blocksworld-original & \textbf{1} & 26.63 & $>$ & 24.77 \\
        blocksworld-advanced & 0.82 & 36.91 & $<$ & 37.76 \\
        earth-observation & 0.45 & 37.17 & $>$ & 28.78 \\
        elevators & 0.93 & 37.64 & $>$ & 29.71 \\
        first-responders & 0.99 & 16.30 & $>$ & 15.42 \\
        zenotravel & 0.87 & 49.38 & $>$ & 36.92 \\
        \midrule
        miner & 0.59 & 16.87 & $<$ & 16.97 \\
        tireworld-truck & \textbf{1} & 19.31 & $>$ & 14.69 \\
        \bottomrule
    \end{tabular}
    \caption{\rev{PR2 vs. AND$^*$}}
\end{subtable} \\ \\

\begin{subtable}[t]{.5\textwidth}
    \centering
    \setlength\tabcolsep{3pt}
    \fontsize{9}{10}\selectfont
    \begin{tabular}{lc|rcr}
        \toprule

        \textbf{Domain} & $\%{\cap}\mathrm{S}$ & $|\tau_*|$ && $|\tau_*|$ \\

        \midrule
        blocksworld-original & 0.33 & 10.10 & $<$ & 10.70 \\
        blocksworld-advanced & 0.24 & 9.92 & $<$ & 10.77 \\
        earth-observation & 0.12 & 8.80 & $>$ & 8.60 \\
        elevators & 0.47 & 14.86 & $<$ & 15.86 \\
        first-responders & 0.59 & 8.98 & $<$ & 9.39 \\
        zenotravel & 0.33 & 14.80 & $<$ & 15.40 \\
        \midrule
        miner & 0.59 & 16.60 & $<$ & 16.97 \\
        tireworld-truck & 0.85 & 13.46 & $<$ & 13.90 \\
        \bottomrule
    \end{tabular}
    \caption{\rev{FOND-SAT vs. AND$^*$}}
\end{subtable}
&
\begin{subtable}[t]{.5\textwidth}
    \centering
    \setlength\tabcolsep{3pt}
    \fontsize{9}{10}\selectfont
    \begin{tabular}{lc|rcr}
        \toprule

        \textbf{Domain} & $\%{\cap}\mathrm{S}$ & $|\tau_*|$ && $|\tau_*|$ \\

        \midrule
        blocksworld-original & 0.33 & 10.50 & $<$ & 10.70 \\
        blocksworld-advanced & 0.20 & 8.82 & $<$ & 9.55 \\
        earth-observation & 0.18 & 14.14 & $>$ & 12.29 \\
        elevators & 0.47 & 14.86 & $<$ & 15.86 \\
        faults & 0.67 & 12.76 & $>$ & 11.76 \\
        first-responders & 0.55 & 8.63 & $<$ & 8.90 \\
        zenotravel & 0.33 & 14.80 & $<$ & 15.40 \\
        \midrule
        miner & 0.59 & 16.60 & $<$ & 16.97 \\
        tireworld-truck & \textbf{1} & 14.26 & $<$ & 14.69 \\
        \bottomrule
    \end{tabular}
    \caption{\rev{CFOND-ASP vs. AND$^*$}}
\end{subtable}
\end{tabular}
}
\caption{\rev{Comparison on the size of returned solutions, on the intersection of solved tasks, excluding the domains where both compared planners return solutions of the same size.}}
\label{tab:results-size-comparison}
\end{table}

We compare the (compressed) size of solutions returned by AND$^*$ with the size of solutions returned by PRP, FOND-SAT, CFOND-ASP, and PR2. We do not compare with IDFSP because it reasons over states and returns solutions over states, while the other four planners reason over partial states and return solutions over partial states.
Table~\ref{tab:results-size-comparison}a presents the comparison with PRP. The nine domains in the table are the ones with at least one task --- among the ones solved by both AND$^*$ and PRP --- with different returned solution sizes. Each row presents the average returned solution size of AND$^*$ and PRP for the entire intersection of tasks solved by both of them. Table~\ref{tab:results-size-comparison}b, \ref{tab:results-size-comparison}c, and \ref{tab:results-size-comparison}d do the same, but respectively for PR2, FOND-SAT and CFOND-ASP. We note that we disregard the mapping of the goal condition to the no-op action, which some of these planners explicitly include in their solution partial policies.

We verify that there is only one domain where AND$^*$ returns on average bigger solutions than PRP: \emph{blocksworld-advanced} (2.3\% bigger). AND$^*$ returns solution partial policies on average at least 10\% smaller for five domains: \emph{earth-observation} (24.0\% smaller), \emph{elevators} (22.7\%), \emph{tireworld-triangle} (33.2\%), \emph{zenotravel} (20.0\%), and \emph{tireworld-truck} (42.7\%). There are three other domains where AND$^*$ also returns smaller solutions on average: \emph{blocksworld-original}, \emph{first-responders}, and \emph{miner}. For the remaining seven domains, AND$^*$ and PRP have tied in this metric.

The comparison with PR2 shows similar results to the comparison with PRP. There are only two domains where AND$^*$ returns on average bigger solutions than PR2: \emph{blocksworld-advanced} (2.3\% bigger) and \emph{miner} (0.6\%). There are six domains where AND$^*$ returns smaller solutions on average: \emph{blocksworld-original}, \emph{earth-observation}, \emph{elevators}, \emph{first-responders}, \emph{zenotravel}, and \emph{tireworld-truck} --- all but \emph{blocksworld-original} and \emph{first-responders} with at least 20\% smaller solutions. For the remaining eight domains, AND$^*$ and PR2 have tied in this metric.

In comparison to FOND-SAT, the sizes of solutions returned by AND$^*$ are very comparable. The difference in the average returned solution size was never more than 10\% for a domain. Nevertheless, for seven domains, FOND-SAT returned smaller solutions, while for only one, \emph{earth-observation}, the opposite was observed.
This is possible because, while FOND-SAT is expected, by design, to return compact solution partial policies, the SAT formulation it uses has some restrictions that may make it miss the minimal-sized solution partial policy.
For the remaining eight domains, AND$^*$ and FOND-SAT have tied in this metric.

The comparison with CFOND-ASP shows similar results to the comparison with FOND-SAT. The difference in the average returned solution size was usually less than 10\%, except for \emph{earth-observation}, where AND$^*$ returned on average 13.1\% smaller solutions. AND$^*$ also returned smaller solutions on average for \emph{faults}. On the other hand, it returned bigger solutions on average for seven domains. For the remaining seven domains, AND$^*$ and CFOND-ASP have tied in this metric.

\section{Conclusion and Future Work}
\label{sec:c-fw}

\color{revbcolor}

This work contributes to the field of fully-observable non-deterministic (FOND) planning by presenting new techniques and insights into policy-space search and solution representation. Central to our contributions is the study of policy equivalences and their impact on making policy-space search a more effective approach for solving FOND planning tasks.

Through an analysis of policy-space structures, we studied two distinct policy features -- domain-escape and escape -- that enable the derivation of policy equivalences with different guarantees and effectiveness. Using escape equivalence pruning substantially improved the performance of AND$^*$~\citep{messa2023best}, the policy-space explicit search algorithm that we focus on in this work.

An additional contribution for the FOND planning community is the \emph{concretizer}, a polynomial-time procedure that, given the states of a solution policy, deduces their correct mappings. In this work, the concretizer enables the sound and complete application of domain-escape pruning. Importantly, the concretizer illustrates that knowing the set of states that need to be mapped is sufficient to retrieve a solution policy, making it a potentially transformative tool for other FOND planners and probabilistic planning approaches.

We also showed that accounting for structural symmetries among states enables stronger policy pruning, and that using group theory techniques effectively allows a more rigorous computation of these symmetries. The efficient computation of canonical symmetries for FOND planning highlights the potential for using canonical symmetries in other planning paradigms.

We also introduce the \emph{compressor}, a technique for reducing solution policies to their minimal unambiguous representation using partial states. On our benchmark tasks, the compressor was computationally efficient and allowed AND$^*$ to achieve state-of-the-art solution compactness in most domains. Importantly, the compressor can compress decompressed solutions from any FOND planner, making it a useful tool for improving the compactness of solutions in the field.

In summary, this work presents a set of distinct new contributions for the FOND planning community. The new techniques make AND$^*$ competitive and provide a basis for designing more efficient algorithms and heuristics for FOND planning.

Future directions include developing heuristics tailored to policy-space search with equivalence pruning and exploring the use of partial states to compress information during search, possibly incorporating dead-end generalization techniques from PRP and PR2. Further research on the \emph{concretizer} could unlock new possibilities, including its use in designing entirely new FOND algorithms. Specifically, we can leverage the ability to deduce solutions from domain sets~$D$ alone via concretizer calls on $\langle D, \states_* \rangle$, noting that we can efficiently check whether a state is in $\states_*$, despite its size.

\section*{Acknowledgments}

We thank UFRGS, CNPq, CAPES, and FAPERGS for partially funding this research.
The present work was carried out with the support of CNPq, Conselho Nacional de Desenvolvimento Científico e Tecnológico - Brazil.
We acknowledge support from FAPERGS with project 21/2551-0000741-9.
This study was financed in part by the Coordenação de Aperfeiçoamento de Pessoal
de Nível Superior - Brasil (CAPES) - Finance Code 001.
We thank the authors of \cite{winterer2016structural} for making their code available.

\bibliographystyle{elsarticle-harv}
\bibliography{bibliography.bib}






\pagebreak
\appendix

\section{Glossary of Symbols and Functions}
\label{a-sec:list-of-symbols}

\newenvironment{listofsymbols}[1]{
        \begin{list}{\textbf{??}}{
                \settowidth{\labelwidth}{#1}
                \setlength{\labelsep}{1em}
                \setlength{\itemindent}{0mm}
                \setlength{\leftmargin}{\labelwidth}
                \addtolength{\leftmargin}{\labelsep}
                \setlength{\rightmargin}{0mm}
                \setlength{\itemsep}{.1\baselineskip}
                \renewcommand{\makelabel}[1]{\makebox[\labelwidth][l]{##1}}
        }
}{
        \end{list}
}

\newcommand{\rbracket}{]}
\begin{listofsymbols}{$\escape(\pi, s)$}
      \item[$\facts$] The set of facts, which might be either true or false in a state.
      \item[$\facts[s\rbracket$] The set of facts that are true in the state~$s$.
      \item[$\states$] The set of states.
      \item[$\states_*$] The set of \emph{goal} states.
      \item[$\actions$] The set of actions.
      \item[$\succs(s, a)$] The set of all states that might be achieved if we apply the action~$a$ at the state~$s$. Also called the successor states of $s$ through $a$.
      \item[$\pi$] A partial function that maps non-goal states to actions. It is called a \emph{policy}.
      \item[$\pi[s\rbracket$] The action that $s$ is mapped to in $\pi$. $\pi[s] = \bot$ iff $s$ is not mapped in $\pi$.
      \item[$\domain(\pi)$] The set of states mapped in $\pi$ --- i.e., the domain of $\pi$.
      \item[$\reach(\pi)$] The set of states reachable from some state in $\domain(\pi)$ through the use of actions in $\pi$. This set includes the domain states themselves.
      \item[$\front(\pi)$] Equal to $\reach(\pi) \setminus \domain(\pi)$. A state in $\front(\pi)$ is called an escape state of $\pi$.
      \item[$\remain(\pi)$] Equal to $\front(\pi) \setminus \states_*$ plus the initial state if it is non-goal and is not in the domain of the policy. A proper policy~$\pi$ is a solution iff $\remain(\pi)$ is empty.
      \item[$\reach(\pi, s)$] The set of states reachable from $s$ through the use of actions in $\pi$. This set includes $s$ itself.
      \item[$\escape(\pi, s)$] Equal to $\reach(\pi, s) \setminus \domain(\pi)$. Also equal to $\reach(\pi, s) \cap \front(\pi)$. A~policy~$\pi$ is proper iff, for each domain state~$s$, this set is \emph{not} empty.
      \item[\revb{$\pi|_X$}] \revb{Operation that returns the policy $\pi' = \{s \mapsto \pi[s] : s \in \domain(\pi) \cap X\}$ for $X \subseteq \states$. In other words, it returns a copy of $\pi$ pruned of states not in $X$.}
\end{listofsymbols}

\end{document}